\documentclass[10pt,twocolumn,letterpaper]{article}

\usepackage[top=0.9in,bottom=0.95in,left=0.75in,right=0.75in,columnsep=0.3in]{geometry}
\ifdefined\XeTeXrevision
  \usepackage{fontspec}
\else
  \usepackage[T1]{fontenc}
  \usepackage{times}
\fi

\usepackage{microtype}
\usepackage{graphicx}
\usepackage[font=small,labelfont=bf]{caption}
\usepackage{subcaption}
\usepackage{booktabs}
\usepackage{multirow}
\usepackage[numbers,sort&compress]{natbib}
\usepackage[colorlinks=true,allcolors=blue!55!black]{hyperref}
\usepackage{xurl}

\usepackage{amsmath}
\usepackage{amssymb}
\usepackage{mathtools}
\usepackage{amsthm}
\usepackage{xcolor}
\usepackage{colortbl}
\usepackage{stfloats}
\usepackage[capitalize,noabbrev]{cleveref}
\usepackage{algorithm}
\usepackage{algorithmic}

\definecolor{tierfour}{HTML}{009E73}
\definecolor{tiereight}{HTML}{0072B2}
\definecolor{tierwarn}{HTML}{D55E00}
\definecolor{tierband}{HTML}{E69F00}

\theoremstyle{plain}
\newtheorem{theorem}{Theorem}
\newtheorem{proposition}{Proposition}
\newtheorem{lemma}{Lemma}

\newcommand{\ours}{ZenMargin}
\newcommand{\oursfwht}{ZenMargin-FWHT}
\newcommand{\oursband}{ZenMargin-Band}
\newcommand{\ourscache}{ZenMargin-Cache}
\newcommand{\sagetwo}{SageAttention2++}
\newcommand{\sagethree}{SageAttention3}
\newcommand{\thrift}{ThriftAttention}
\newcommand{\deco}{DeCo-Attention}
\newcommand{\kwide}{\texttt{K8V4-wide}}
\newcommand{\safefig}[2]{\IfFileExists{#2}{\includegraphics[width=#1]{#2}}{\fbox{\parbox[c][3cm][c]{#1}{\centering\small figure pending: \texttt{\detokenize{#2}}}}}}

\title{\textbf{\ours{}: Margin-Refined Two-Limb FP4 Attention\\for Long-Context Language Models}}

\author{Yingwei Song$^{1,3}$ \qquad Tony Zhu$^{2}$ \qquad Kelly Peng$^{3}$ \qquad Yuan Gao \qquad Huanrui Yang$^{1,3}$\\[5pt]
{\normalsize $^{1}$University of Arizona \qquad $^{2}$Tsinghua University \qquad $^{3}$First Intuition}\\[3pt]
{\footnotesize\texttt{\{yingwei,huanruiyang\}@arizona.edu} \quad \texttt{me@tonyzoo.com} \quad \texttt{kelly@firstintuition.ai} \quad \texttt{williamgaoharvard@gmail.com}}\\[4pt]
{\small Code: \url{https://github.com/wrencanfly/ZenMargin}}}
\date{}

\begin{document}

\twocolumn[{%
  \renewcommand\twocolumn[1][]{#1}%
  \maketitle
  \vspace{-1.2em}
  \begin{center}
  \safefig{\textwidth}{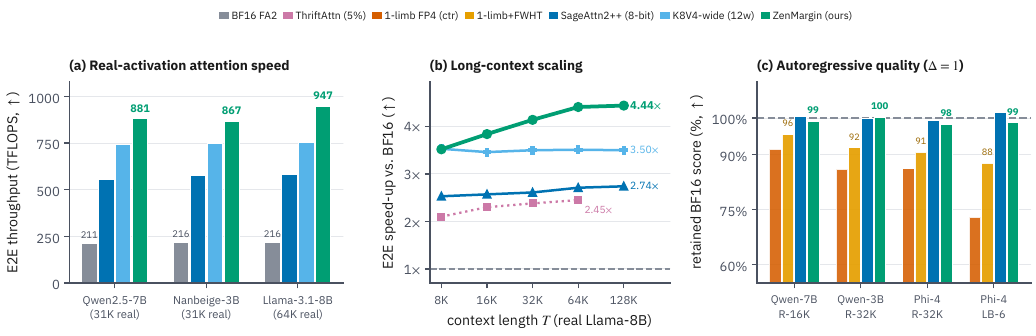}
  \vspace{-3pt}
  \captionof{figure}{\textbf{Speed and standardized autoregressive quality of \ours{}} (RTX~5090, SM120; higher is better). \emph{(a)}~E2E attention throughput on \emph{real} activations (Qwen2.5-7B 31K, Nanbeige-3B 31K, Llama-3.1-8B 64K 8-layer mean): \oursfwht{} ($\Delta{=}1$) reaches $867$--$947$\,TFLOPS ($4.01$--$4.38\times$ FA2), outperforming \sagetwo{}, \thrift{}~5\%, and 12-warp \kwide{} with $44\%$ less KV memory than BF16. \emph{(b)}~Real 8K--128K Llama-3.1-8B scaling: \oursfwht{} reaches $4.44\times$ ($967$\,TFLOPS), while \thrift{}~5\% caps at $2.45\times$ (below \sagetwo{} at $2.74\times$). \emph{(c)}~Autoregressive RULER and official LongBench scores normalized to BF16 ($100\%$): clean causal 1-limb FP4 degrades even with FWHT ($88$--$96\%$), whereas \oursfwht{} ($\Delta{=}1$) recovers $98$--$100\%$ of BF16 accuracy across all LLMs (\cref{tab:gen_main}).}
  \label{fig:hero}
  \end{center}
  \vspace{0.5em}
}]

\begin{abstract}
Long-context LLM serving faces a frustrating trade-off on 4-bit tensor cores: uniform 4-bit attention degrades long-range retrieval, while selective FP16 fallback forces systems to keep a full-precision KV cache and leaves decoding unaccelerated.
We show that no FP16 backup is needed.
Analyzing why 4-bit retrieval fails reveals two root causes: quantization error concentrates along a few dominant key directions---creating strongly correlated score errors---and retrieval heads operate with narrow score margins where modest 4-bit noise flips the winning key.
\textbf{\ours{}} resolves both directly in 4-bit precision.
It smooths keys with causal centering and a lightweight Hadamard transform, then splits every KV pair into a \emph{primary} 4-bit tensor and a \emph{residual} 4-bit tensor under a shared scale (two-limb NVFP4).
During prefill, an online margin gate reads only the primary 4-bit limb for $95$--$99\%$ of blocks and refines the remaining $1$--$5\%$ of near-peak blocks natively on FP4 tensor cores; during decoding, \textbf{\ourscache{}} reuses the exact same split layout with zero repacking.

Across six open LLMs on a single RTX~5090 (up to 128K context), \ours{} is simultaneously \emph{fast}, \emph{compact}, and \emph{faithful}.
It reaches $867$--$967$\,TFLOPS on real activations ($4.0$--$4.4\times$ FlashAttention-2, $1.5$--$1.6\times$ \sagetwo{}, and faster than FlashInfer NVFP4), holds flat throughput across diffuse and code workloads with zero degradation cliff, and speeds up 64K time-to-first-token by up to $1.9\times$.
By cutting KV-cache memory $44\%$ below BF16 while reading mostly the primary 4-bit limb, \ourscache{} lifts 32-layer single-GPU serving throughput $2.1$--$2.9\times$ over FlashInfer BF16---and outperforms FlashInfer FP8---across both long-input and incremental-flush Long-CoT workloads.
Without retraining, it matches BF16 accuracy across full 13-task RULER (up to 67.6K) and LongBench.
\end{abstract}

\section{Introduction}
\label{sec:intro}

Serving long-context large language models (LLMs) runs into two coupled bottlenecks: compute-bound self-attention during prefill over tens of thousands of tokens \citep{dao2024flashattention2}, and memory-capacity and bandwidth limits on the resulting key--value (KV) cache during batched decoding.
On NVIDIA Blackwell GPUs (SM120 / RTX~5090), block-scaled 4-bit floating-point (NVFP4) tensor cores run roughly $2\times$ faster than INT8 and $>4.5\times$ faster than BF16 \citep{alvarez2025nvfp4,zhang2025sageattention3}.
Turning that hardware speed into a practical long-context engine requires clearing three bars at once: outperforming dense 8-bit attention end-to-end in prefill, preserving BF16 retrieval fidelity at 64K+ tokens, and compressing KV memory and read bandwidth across \emph{both} prefill and decoding without keeping a full-precision fallback copy.

\textbf{Why existing 4-bit approaches stall.}
Current 4-bit attention designs miss at least one of these three bars (\cref{tab:options}).
(1)~\emph{Uniform 4-bit quantization} leaves unconditioned 1-limb NVFP4 with over $1$\,logit of score noise on retrieval heads, causing large accuracy drops on long-context multi-key retrieval even under exact causal masking (\cref{sec:exp}).
(2)~\emph{Uniform channel compensation} (\deco{}~\citep{he2026decoattention}; ResQ~\citep{saxena2024resq}) protects a fixed low-rank subspace across all keys, yet residual 4-bit noise still disrupts thin-margin retrieval keys (\cref{sec:spectral,app:tf_eval}).
(3)~\emph{Proxy-based selective FP16} (\thrift{}~\citep{sharratt2026thriftattention}) selects a fixed top-$f$ fraction of 64-key blocks via block-mean inner products $\bar q\cdot\bar k$ and refines them in FP16.
Averaging over 64 keys dilutes an isolated needle by $64\times$; worse, because refinement runs in FP16 and future queries are unknown upfront, the system must retain a full FP16 KV cache---forfeiting KV compression and capping real 64K prefill throughput \emph{below} dense 8-bit \sagetwo{}.

\begin{table}[t]
\caption{\textbf{Low-bit causal attention on Blackwell (SM120).} Storage: KV B/key/head ($d{=}128$). Speed: E2E $\times$FA2 on \emph{real} Llama-3.1-8B 64K activations (8-layer mean, RTX~5090; \cref{tab:sota_speed}). Quality: Llama-3.1-8B 10-task LongBench ($n{=}300$; \cref{tab:gen_main}).}
\label{tab:options}
\centering
\small
\setlength{\tabcolsep}{2.3pt}
\resizebox{\columnwidth}{!}{%
\begin{tabular}{@{}llllrr@{}}
\toprule
Method & Selection & Refined path & B/key & $\times$FA2 & LB-10 \\
\midrule
FlashAttn-2 (BF16) & --- & --- & 512 & $1.00\times$ & $49.14$ \\
\sagetwo{} (GQA) & --- & INT8\,/\,FP8 & 512$^\dagger$ & $2.69\times$ & $49.22$ \\
\kwide{} (12w) & --- & INT8\,/\,NVFP4 & 204 & $3.50\times$ & $49.05$ \\
\ours{} (1-limb+FWHT) & --- & NVFP4 & 144 & $4.82\times$ & $48.10$ \\
\thrift{} ($5\%$) & $\bar q\!\cdot\!\bar k$ top-$f$ & FP16 copy & 512+$^\dagger$ & $2.45\times$ & $49.50$ \\
\rowcolor{tierfour!10}
\textbf{\oursfwht{}} ($\Delta{=}1$) & Online margin & 2-limb NVFP4 & 288 & $\mathbf{4.38\times}$ & $\mathbf{49.80}$ \\
\rowcolor{tierband!10}
\textbf{\oursband{}} ($\Delta{=}1$, b4) & Margin + band & 2-limb NVFP4 & 308 & $4.04\times$ & $48.95$ \\
\bottomrule
\end{tabular}}
\\[1pt]{\footnotesize $^\dagger$Retains BF16/FP16 KV inputs (\thrift{} also stores FP4). Unified prefill+decode \ourscache{} reaches $49.74$ LB-10 and outperforms production FlashInfer XQA FP8 ($256$\,B/key) by $1.13$--$1.15\times$ in 32-layer serving.}
\end{table}

\textbf{Insight 1: Condition the bulk, gate by exact-score margin.}
Our design starts from diagnosing why 4-bit attention breaks and what an online selector must actually guarantee (\cref{sec:analysis}).
Analyzing how quantized score errors propagate through softmax (\cref{prop:firstorder}) reveals two complementary failure modes.
First, unrotated 4-bit key errors are strongly correlated across tokens because key energy concentrates in a few dominant directions; removing per-head key DC bias and applying a lightweight Walsh--Hadamard rotation largely breaks this correlation at negligible cost.
Second, $75$--$88\%$ of long-context retrieval heads have gold-vs-distractor score margins below $1$\,logit.
Rather than relying on lossy block-mean proxies, we refine any 64-key block whose cheap 4-bit maximum falls within a small logit margin $\Delta$ of the running maximum.
Crucially, a cheap-score online monotonicity lemma (\cref{lem:online_superset}) bridges to a stronger \emph{exact-score} guarantee (\cref{thm:exact_retention}): once key conditioning pushes 1-limb score noise below the margin threshold, single-sweep online gating retains exact-important tokens and bounds unrefined squared attention mass independently of context length $N$.
In practice, this refines only $1$--$5\%$ of blocks at long context while capturing $>99.9\%$ of squared attention mass.

\textbf{Insight 2: Split 4-bit refinement removes the FP16 backup and unifies prefill with decoding.}
Why did prior selective methods need an FP16 KV cache in the first place?
Because coarse scoring ran in 4-bit while refinement jumped to a separate FP16 tensor, forcing two incompatible layouts in memory.
\ours{} eliminates that split by encoding every KV tensor into two 4-bit halves under one shared per-head scale (\cref{fig:architecture}, \cref{sec:method}): a \emph{primary} NVFP4 limb ($144$\,B/key) and a \emph{residual} NVFP4 limb ($144$\,B/key), totaling $288$\,B/key---$44\%$ smaller than BF16.
During prefill, $95$--$99\%$ of long-context blocks read only the primary limb, while the $1$--$5\%$ of margin-flagged blocks fetch the residual limb on demand and complete two-limb refinement natively on FP4 tensor cores.
Because prefill writes every block directly into this split layout, decoding (\ourscache{}) consumes the exact same cache with zero repacking---reading only the primary limb plus rare residual fetches ($3.5\times$ less KV traffic than BF16 and $1.75\times$ less than FP8), and pairing naturally with static scales for bounded-tail incremental flushing during long chain-of-thought (Long-CoT) generation.

\textbf{Contributions.}
(i)~\emph{Theory and diagnosis (\cref{sec:analysis}):} we show how correlated 4-bit score errors and sub-logit retrieval margins break uniform and proxy-gated FP4 (\cref{prop:firstorder}), and prove that conditioned online margin gating guarantees exact-important token retention and context-length-free squared-mass control (\cref{lem:online_superset,thm:exact_retention}), validated across four LLM families.
(ii)~\emph{Unified algorithm and SM120 system (\cref{sec:method}):} we introduce \ours{} and \ourscache{}, unifying zero-calibration key conditioning, shared-scale two-limb NVFP4 with no FP16 backup, a single-sweep asynchronous Blackwell prefill kernel, split-KV FlashDecoding, and bit-exact 64-token incremental flushing for long-output serving.
(iii)~\emph{End-to-end validation (\cref{sec:exp}):} across six open LLMs on RTX~5090 (up to 128K), \ours{} achieves $4.0$--$4.4\times$ prefill speedup over FlashAttention-2 ($1.5$--$1.6\times$ over \sagetwo{}, faster than FlashInfer NVFP4) with flat throughput on diffuse and code workloads, $1.6$--$1.9\times$ 64K TTFT speedup, BF16-equivalent accuracy on full 13-task RULER and LongBench, and $2.1$--$2.9\times$ higher 32-layer serving throughput than FlashInfer BF16 (outperforming FlashInfer FP8) across both long-input and Long-CoT workloads.

\textbf{Positioning.}
Decode-only FP4 KV caches \citep{du2026bitdecodingfp4,ni2026nvfp4mla,gupta2026scalesearch} leave prefill unaccelerated, prefill-only kernels \citep{zhang2025sageattention3,sharratt2026thriftattention,ding2026dma} omit KV compression, and concurrent TriAttention \citep{zhang2026triattention} pairs block skipping with 4/8-bit precision; \ours{} unifies dense prefill and decoding in one two-limb NVFP4 layout without dropping tokens or keeping FP16 backups (\cref{sec:related}).

\section{Preliminaries}
\label{sec:prelim}

\textbf{Causal GQA attention.}
Consider a grouped-query attention (GQA) layer with $H$ KV heads, group size $G$ ($H_q{=}HG$), sequence length $T$, and head dimension $d{=}128$.
For query row $i$ and causal keys $j \le i$, exact attention computes logits $s_{ij} = \langle q_i, k_j\rangle / \sqrt{d}$, weights $p_{ij} = \exp(s_{ij} - m_i) / \sum_{u \le i}\exp(s_{iu} - m_i)$ with $m_i = \max_{u \le i} s_{iu}$, and output $o_i = \sum_{j \le i} p_{ij} v_j$.
FlashAttention \citep{dao2022flashattention,dao2024flashattention2} tiles keys into 64-token stages and maintains $(m_i, \ell_i, o_i)$ via online softmax rescaling without materializing $T\times T$ scores.

\textbf{Block-scaled NVFP4 arithmetic.}
NVIDIA Blackwell (SM120) supports hardware block-scaled 4-bit floating-point (NVFP4) matrix multiplication \citep{alvarez2025nvfp4}.
An NVFP4 tensor $X$ stores a per-head FP32 global scale $g_X = \max |X| / (6\cdot 448)$, one unsigned 8-bit E4M3 microscale $s_b \in [0, 448]$ per 16 elements along the inner reduction axis, and 4-bit E2M1 codes $c \in \{\pm 0, \pm 0.5, \pm 1, \pm 1.5, \pm 2, \pm 3, \pm 4, \pm 6\}$, decoding as $\hat x_{b,u} = g_X \cdot \mathrm{dec}_{\mathrm{E4M3}}(s_b) \cdot \mathrm{dec}_{\mathrm{E2M1}}(c_{b,u})$.
$Q, K$ are microscaled along $d{=}128$ ($8$ blocks of $16$ channels), while $P, V$ are microscaled along each 64-key stage ($4$ blocks of $16$ keys).
On RTX~5090, NVFP4 tensor cores deliver roughly $2\times$ the throughput of INT8 and $>4.5\times$ BF16 (hardware PTX details in \cref{app:kernel_details}).

\textbf{Long-context evaluation metrics.}
We evaluate fidelity via (i)~\emph{standardized autoregressive generation} on RULER \citep{hsieh2024ruler} (8K--67.6K, exact recall) and official LongBench \citep{bai2024longbench} (300 prompts/model) with paired bootstrap $95\%$ confidence intervals, and (ii)~\emph{teacher-forced diagnostics} on LongBench and 32K--64K 2-hop JSON chains (\cref{app:tf_eval}), measuring paired $\Delta\text{logp}$ (nats vs.\ BF16), greedy exact match (\texttt{tf-ok}), and tail failures (\texttt{bad}, $<-0.3$\,nats).

\section{Why FP4 Attention Fails on Long-Context LLMs}
\label{sec:analysis}

To understand why uniform FP4 and heuristic selective or compensated FP4 schemes break down on long-context LLMs, we analyze how quantized score errors propagate through softmax, prove exact-score retention and context-length-free squared-mass guarantees for our online cheap-score margin gate, and diagnose the geometric origin of score noise.

\begin{figure*}[t]
  \centering
  \safefig{0.92\textwidth}{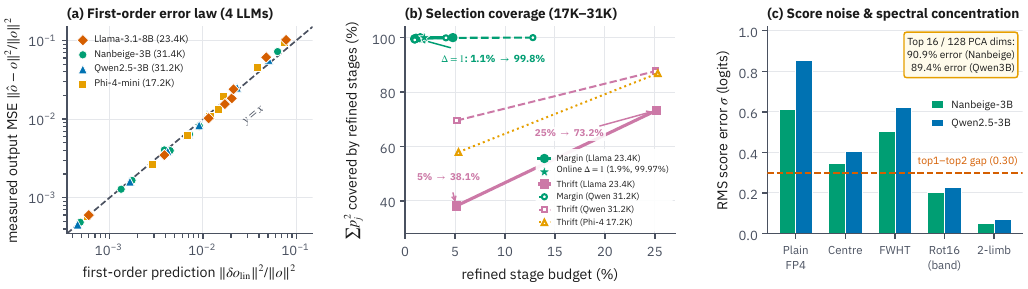}
  \vspace{-4pt}
  \caption{\textbf{Diagnosis and theory on 17K--31K prompts across four LLMs} (same \oursfwht{} suite as \cref{tab:app_theory,tab:app_online_forensics}). \emph{(a)}~Exact raw-error linearization (\cref{prop:firstorder}) matches measured MSE within $2.9$--$12.9\%$ across margin gating ($\Delta\!\in\![0.5,4]$). \emph{(b)}~Coverage of $\sum p_j^2$: on Llama-3.1-8B (23.4K), margin gating ($\Delta{=}1$) covers $99.8\%$ at $1.1\%$ global budget ($99.97\%$ at $1.9\%$ actual online budget) vs.\ $38.1\%$ ($5\%$) and $73.2\%$ ($25\%$) for \thrift{}. \emph{(c)}~All-head RMS score noise vs.\ Nanbeige's $0.30$-logit gap; $89.4$--$90.9\%$ of 1-limb error lies in the top-16 key-PCA directions.}
  \label{fig:theory}
\end{figure*}

\subsection{Raw-Error Perturbation Decomposition and Online Margin Theory}
\label{sec:theory}

Fix a query row with exact logits $s \in \mathbb{R}^N$, weights $p = \mathrm{softmax}(s)$, exact values $\{v_j\}_{j=1}^N$, and output $o = \sum_j p_j v_j$.
To isolate score-quantization error, suppose a scheme refines keys $\mathcal{R} \subseteq \{1,\dots,N\}$ and leaves $\mathcal{U} = \mathcal{R}^c$ at 1-limb FP4 with raw score errors $e_j = \hat s_j - s_j$ (end-to-end kernel error additionally includes $\hat P$ FP4 quantization and 1-limb $V$ error; \cref{sec:twolimb}).

\begin{proposition}[Raw-error first-order covariance decomposition]
\label{prop:firstorder}
Let $A = [\,p_1(v_1 - o), \dots, p_N(v_N - o)\,] \in \mathbb{R}^{d \times N}$. Because $\sum_{j=1}^N p_j(v_j - o) = 0$, the first-order softmax perturbation $\delta p_j = p_j(e_j - \sum_i p_i e_i)$ maps to the exact \emph{raw-error} linear perturbation $\delta o_{\mathrm{lin}} = \sum_{j=1}^N p_j e_j (v_j - o) = A e$. For raw score errors with mean $\mu_e = \mathbb{E}[e]$ and covariance $\Sigma_e = \mathrm{Cov}(e)$,
\begin{equation}
\begin{aligned}
  \mathbb{E}\|\delta o_{\mathrm{lin}}\|^2 &= \operatorname{tr}(A \Sigma_e A^\top) + \|A \mu_e\|^2 \\
  &= \underbrace{\sum_{j=1}^N p_j^2 \mathrm{Var}(e_j)\|v_j - o\|^2}_{\text{independent variance } (D_{\mathrm{raw}})}
  + \underbrace{\|A\mu_e\|^2}_{\text{score bias}} \\
  &\quad + \underbrace{\sum_{j \neq k} p_j p_k \mathrm{Cov}(e_j, e_k)\langle v_j - o, v_k - o\rangle}_{\text{cross-key covariance}}.
\end{aligned}
\label{eq:p2law}
\end{equation}
On any realization $e$, the exact independent-sign control $\mathbb{E}_{\xi \in \{\pm 1\}^N}\|A(\xi \odot e)\|^2 = \sum_{j=1}^N p_j^2 e_j^2 \|v_j - o\|^2 = D_{\mathrm{raw}}(e)$ isolates the raw diagonal without centering artifacts; when $e_j$ are zero-mean uncorrelated ($\sigma_c$ on $\mathcal{U}$, $\sigma_r \ll \sigma_c$ on $\mathcal{R}$), the unrefined term is bounded by $\sigma_c^2 V_{\max}^2 \sum_{j\in\mathcal{U}} p_j^2$ ($V_{\max} = \max_j \|v_j - o\|$).
\end{proposition}

\textbf{Why raw-error controls matter.}
Writing the diagonal in terms of centered errors $\tilde e_j = e_j - \sum_i p_i e_i$ introduces spurious cross-key correlation even for i.i.d.\ $e_j$ (e.g., for $p_1{=}p_2{=}\tfrac{1}{2}$ and $v_1{=}1, v_2{=}-1$, $\tilde e_2 = -\tilde e_1$ doubles the total-to-diagonal ratio despite independent $e_1, e_2$).
Evaluating \cref{eq:p2law} directly on \emph{raw} uncentered $e_j$ against both matched-variance Rademacher sign-flips ($\xi_j \in \{\pm 1\}$, matching $D_{\mathrm{raw}}$ to 3 digits) and within-set random permutations (\texttt{shuf}, preserving marginal error distributions and set means on $\mathcal{U}, \mathcal{R}$ separately) across four LLMs (\cref{tab:app_theory}) cleanly separates three regimes:
(i)~Without rotation (\texttt{ctr-only}, unrefined), true cross-key correlation amplifies linearized MSE by $3.53$--$4.10\times$ over within-set shuffled errors; applying orthogonal FWHT slashes this ratio to $1.12$--$2.18\times$ and cuts total MSE $2.0$--$3.2\times$.
(ii)~In peak-dominated unrefined or \thrift{} rows under FWHT, $\|\delta o_{\mathrm{lin}}\|^2 / D_{\mathrm{raw}}$ is only $0.75$--$1.05\times$ (confirming that centered diagonals overestimate correlation on peak-dominated rows).
(iii)~Once high-weight peaks are refined under margin gating ($\Delta\!\in\![0.5,4]$), residual error sits on the diffuse unrefined tail ($10$--$30\%$ linear mass), where raw cross-key covariance ($31$--$46\%$ of linearized MSE) plus refined-vs-unrefined set bias ($11$--$30\%$) amplify $\|\delta o_{\mathrm{lin}}\|^2$ by $\mathbf{1.79}$--$\mathbf{3.07\times}$ over $D_{\mathrm{raw}}$ ($\mathbf{1.77}$--$\mathbf{2.36\times}$ over \texttt{shuf})---explaining why conditioning and margin gating are complementary.

\begin{lemma}[Online monotonicity and superset chain]
\label{lem:online_superset}
Fix any query row $r$, quantized cheap scores $\{\tilde s_{r,j}\}_{j=1}^N$ partitioned into stages $\{B_s\}_{s=1}^S$ with stage maxima $\tilde m_{r,s} = \max_{j \in B_s} \tilde s_{r,j}$, and any stage traversal permutation $\pi$. At step $t$ visiting stage $s = \pi(t)$, the running cheap-only maximum satisfies $\tilde m_r^{\mathrm{ref}}(t) = \max_{u \le t} \tilde m_{r,\pi(u)} \le \max_{1 \le u \le S} \tilde m_{r,u} = \tilde s_{r,\max}$. Consequently, the key-level global, stage-level global, single-row online, and 16-row warp-union refinement sets obey the strict inclusion chain
\begin{equation}
  \tilde{\mathcal{R}}_{\Delta,r}^{\mathrm{key\text{-}global}}
  \;\subseteq\; \tilde{\mathcal{R}}_{\Delta,r}^{\mathrm{stage\text{-}global}}
  \;\subseteq\; \tilde{\mathcal{R}}_{\Delta,r}^{\mathrm{online}}
  \;\subseteq\; \tilde{\mathcal{R}}_{\Delta}^{\mathrm{warp\text{-}online}}.
\label{eq:superset}
\end{equation}
\end{lemma}

\Cref{lem:online_superset} acts as a bridge: because any key omitted by the single-sweep online warp kernel is also omitted by $\tilde{\mathcal{R}}_{\Delta,r}^{\mathrm{key\text{-}global}}$, we can directly transfer \emph{exact-score} retention and squared-mass guarantees onto the actual noisy online gate---without paying the context-length-dependent $\max_{1\le k\le N}|\eta_k| \asymp \sigma\sqrt{2\ln N}$ penalty of naive uniform bounds.

\begin{theorem}[Exact-score retention and context-length-free squared-mass guarantee]
\label{thm:exact_retention}
Fix a query row $r$ with exact logits $s_1,\dots,s_N$, exact softmax weights $p_j = e^{s_j}/\sum_k e^{s_k}$, $s_{\max} = \max_k s_k$, $p_{\max} = \max_k p_k$, and cheap scores $\tilde s_j = s_j + \eta_j$. For any $\Delta > 0$, let $\tilde{\mathcal{R}}_\Delta^{\mathrm{warp}}$ denote $\tilde{\mathcal{R}}_\Delta^{\mathrm{warp\text{-}online}}$ from \cref{lem:online_superset}.

\smallskip\noindent\textbf{(a)}~\emph{Deterministic gap-damped exact-score retention.} Define the gap-damped overshoot $\varepsilon_+ := \tilde s_{\max} - s_{\max} = \max_{1\le k\le N}[\,\eta_k - (s_{\max} - s_k)\,]$ and prefix overshoot $\varepsilon_+(t) := \tilde m_r^{\mathrm{ref}}(t) - s_{\max} \le \varepsilon_+$. For any $\gamma \in [0,\Delta]$, every exact-important token $j \in \mathcal{R}_\gamma^{\mathrm{exact}} := \{j : s_j \ge s_{\max} - \gamma\} = \{j : p_j \ge e^{-\gamma}p_{\max}\}$ is guaranteed to be retained ($j \in \tilde{\mathcal{R}}_\Delta^{\mathrm{warp}}$) whenever $\varepsilon_+(t_{s(j)}) - \eta_j \le \Delta - \gamma$ (sufficiently, $\varepsilon_+ - \eta_j \le \Delta - \gamma$). In particular, the exact argmax key $j^* = \arg\max_k s_k$ is omitted only if an earlier-visited distractor $k$ satisfies $\eta_k - \eta_{j^*} > \Delta + (s_{j^*} - s_k)$.

\smallskip\noindent\textbf{(b)}~\emph{Context-length-free exact-score omission and missed-mass bounds.} Suppose pairwise error differences $\eta_k - \eta_j$ ($k \ne j$) are $\tau^2$-sub-Gaussian, i.e., $\mathbb{E}[e^{\lambda(\eta_k - \eta_j)}] \le e^{\frac{1}{2}\lambda^2\tau^2}$ for all $\lambda > 0$ (canceling any row-wide bias; for zero-mean $\sigma^2$-sub-Gaussian $\eta_k$, $\tau^2 = 2\sigma^2$ under independence and $\tau^2 = 4\sigma^2$ under \emph{arbitrary} cross-key dependence via Cauchy--Schwarz). For any $\lambda \ge 1$ and any exact-important token $j \in \mathcal{R}_\gamma^{\mathrm{exact}}$ with $\gamma_j = s_{\max} - s_j < \Delta$,
\begin{equation}
  \Pr\!\big[j \notin \tilde{\mathcal{R}}_\Delta^{\mathrm{warp}}\big]
  \le N_{\mathrm{eff}}(\lambda)\exp\!\Big(\!-\!\lambda(\Delta - \gamma_j) + \tfrac{1}{2}\lambda^2\tau^2\Big),
\label{eq:exact_miss_prob}
\end{equation}
where $N_{\mathrm{eff}}(\lambda) := \sum_{k=1}^N (p_k/p_{\max})^\lambda \le 1/p_{\max}$ is independent of $N$. Whenever $\Delta - \gamma_j \ge \tau^2$, setting $\lambda^* = (\Delta - \gamma_j)/\tau^2$ gives sub-Gaussian decay $\exp\!\big(-(\Delta - \gamma_j)^2/(2\tau^2)\big)$; for the exact top-1 needle $j^*$ with margin $g = s_{(1)} - s_{(2)} \ge 0$, every distractor satisfies $p_k/p_{\max} \le e^{-g}$, tightening \cref{eq:exact_miss_prob} to $\frac{1-p_{\max}}{p_{\max}}\exp\!\big(g - (\Delta+g)^2/(2\tau^2)\big)$ whenever $\Delta + g \ge \tau^2$. Summing over $\mathcal{R}_\gamma^{\mathrm{exact}}$ simultaneously bounds expected missed exact weight ($\forall \lambda \ge 1$):
\begin{equation}
\begin{aligned}
  &\mathbb{E}\!\left[\sum_{j \in \mathcal{R}_\gamma^{\mathrm{exact}} \setminus \tilde{\mathcal{R}}_\Delta^{\mathrm{warp}}} p_j\right] \\
  &\quad \le \left(\sum_{j \in \mathcal{R}_\gamma^{\mathrm{exact}}} p_j\right)\frac{1}{p_{\max}} \exp\!\Big(\!-\!\lambda(\Delta - \gamma) + \tfrac{1}{2}\lambda^2\tau^2\Big).
\end{aligned}
\label{eq:exact_missed_mass}
\end{equation}

\smallskip\noindent\textbf{(c)}~\emph{Context-length-free unrefined squared-mass bound.} Across all $N$ keys,
\begin{equation}
\begin{aligned}
  &\mathbb{E}\!\left[\sum_{j \notin \tilde{\mathcal{R}}_\Delta^{\mathrm{warp}}} p_j^2\right] \\
  &\quad \le e^{-\Delta + \frac{1}{2}\tau^2}\!\left(1 - \sum_{j=1}^N p_j^2\right)
  \le e^{-\Delta + \frac{1}{2}\tau^2},
\end{aligned}
\label{eq:marginbound}
\end{equation}
which reduces to $e^{-\Delta}(1 - \sum_{j=1}^N p_j^2) \le e^{-\Delta}$ as $\tau \to 0$ (matching the exact noiseless bound $\sum_{j\notin\mathcal{R}_\Delta} p_j^2 < e^{-\Delta}(1 - \sum_{j\in\mathcal{R}_\Delta} p_j) \le e^{-\Delta}$).
\end{theorem}

\textbf{Explanatory power, scope, and online-vs-oracle forensics.}
\Cref{thm:exact_retention} resolves three core questions of 4-bit long-context gating:
(i)~\emph{Why context length $N \to 128\text{K}$ does not break the gate:} a naive uniform worst-case bound $e^{-(\Delta - 2\varepsilon_{\max})}$ is vacuous on $17\text{K}$--$31\text{K}$ prompts ($5.35$--$10.20$ at $\Delta{=}1$, since $\max_j |\eta_j|$ has median $1.34$--$1.66$\,logits), because extreme quantization noise on $O(N)$ low-score background keys is damped by $-(s_{\max}-s_k)$ in $\varepsilon_+$ and cannot hide important peaks. By contrast, \cref{eq:marginbound} is independent of $N$: at $\Delta{=}1$ on our four \oursfwht{} LLMs ($\sigma{=}0.316$--$0.401$), it drops from vacuous $5.35$--$10.20$ to a non-vacuous $\mathbf{0.407}$--$\mathbf{0.432}$ (independent $\tau^2{=}2\sigma^2$) and $\mathbf{0.449}$--$\mathbf{0.507}$ even under \emph{arbitrary} cross-key correlation ($\tau^2{=}4\sigma^2$; $0.020$--$0.025$ at $\Delta{=}4$), with the peakiness factor $(1 - \sum_j p_j^2)$ tightening protection on needle rows.
(ii)~\emph{Why conditioning is theoretically necessary for the gate itself:} \cref{thm:exact_retention} exposes the sharp noise threshold $\frac{1}{2}\tau^2 < \Delta$. On unconditioned 1-limb FP4 retrieval heads ($\sigma \approx 1.04$--$1.23 \implies \frac{1}{2}\tau^2 > 1$), the gate's own exact-score guarantee breaks down; centering + FWHT/Band compresses $\sigma$ to $0.20$--$0.40$ ($\frac{1}{2}\tau^2 \approx 0.04$--$0.16 \ll \Delta$), shifting the top-1 needle failure barrier from sub-logit $g$ to $\Delta + g$ and holding the noisy-gate penalty $e^{\frac{1}{2}\tau^2}$ within $1.04$--$1.17\times$ of the noiseless oracle.
(iii)~\emph{Scope and online-vs-oracle forensics:} two boundaries remain explicit---on uniform diffuse rows ($p_{\max} \approx 1/N$), $1/p_{\max}$ scales with $N$ because no single key dominates, yet $\sum p_j^2 \le 1/N$ is already negligible (and real 32K/64K diffuse workloads show zero accuracy/latency cliff; \cref{tab:app_diffuse_profile,tab:app_ruler13}); moreover, small $\sum_{j\notin\mathcal{R}} p_j^2$ still leaves $13.6$--$21.9\%$ diffuse linear mass unrefined at $\Delta{=}1$ (confirming why conditioned 1-limb accuracy remains necessary on the unrefined tail).
Empirical forensics across all four LLMs (\cref{tab:app_online_forensics}) validate \cref{lem:online_superset,thm:exact_retention}:
(1)~$\mathrm{FN} = 0.0000$ vs.\ $\tilde{\mathcal{R}}_\Delta^{\mathrm{global}}$ everywhere, with recent-first + BOS-first traversal cutting online excess refinement $29$--$56\%$ over left-to-right scanning on Nanbeige, Qwen2.5-3B, and Phi-4-mini;
(2)~on high-weight keys ($p_j \ge 0.01$), median $\max |e_j|$ is only $0.50$--$0.69$\,logits, so at $\Delta{=}1$ our online warp gate misses just $2.2$--$5.8\%$ of exact offline-oracle stages, carrying only $0.00001$--$0.00016$ squared mass and $0.05$--$0.55\%$ linear mass;
(3)~thanks to modest excess refinement ($1.5$--$3.0\%$ total stage budget vs.\ $0.7$--$1.3\%$ oracle), the actual online gate achieves $\mathbf{3.0}$--$\mathbf{5.1\times}$ \emph{lower} output MSE than the exact offline oracle ($2.27$--$3.39{\times}10^{-3}$ vs.\ $6.81$--$17.0{\times}10^{-3}$).

\subsection{Why Block-Mean Proxies Miss Retrieval Needles}
\label{sec:proxy_fail}

\thrift{} \citep{sharratt2026thriftattention} promotes a fixed fraction $f$ of 64-key stages per 64-query block ranked by block-mean scores $\bar q \cdot \bar k$.
This fails on long-context LLMs via (i)~\emph{needle cancellation}---one target key among 63 distractors is attenuated $64\times$ in $\bar k$ (and likewise across 64 queries in $\bar q$)---and (ii)~\emph{fixed-budget misallocation} between diffuse and multi-needle rows.
On Llama-3.1-8B (23.4K; \cref{fig:theory}b, \cref{tab:app_theory,tab:app_online_forensics}), \oursfwht{} margin gating ($\Delta{=}1$, 16-row tiles) covers $99.8\%$ of $\sum p_j^2$ at $1.1\%$ global budget ($99.97\%$ at $1.9\%$ actual online budget), whereas \thrift{} covers only $38.1\%$ at $5\%$ budget and $73.2\%$ even at $25\%$ budget; across Llama-3.1-8B (23.4K), Qwen2.5-3B (31.2K), and Phi-4-mini (17.2K), $\Delta{=}1$ ($1.1$--$1.6\%$ global / $1.9$--$3.0\%$ online budget) achieves $3.9$--$4.2\times$ lower MSE than \thrift{} at $5\%$ (whereas on Nanbeige4.2-3B at 31.4K, where block-mean captures $97.2\%$ of $\sum p_j^2$, \thrift{} 5\% reaches lower MSE than $\Delta{=}1$ by spending $6.6\times$ more global stage budget to trim diffuse tail error).

\textbf{Sub-logit retrieval margins.}
Across retrieval heads placing $>20\%$ mass on the gold span during 8K--31K 2-hop lookup (\cref{app:retrieval}), $80$--$83\%$ on Nanbeige4.2-3B (medians $-0.89$ to $-0.41$\,logits) and $75$--$88\%$ on Qwen2.5-3B (medians $-0.63$ to $+0.16$\,logits) have gold-vs-distractor margins below $1$\,logit.
Plain 1-limb NVFP4 injects $1.04$--$1.23$\,logits RMS noise on these retrieval rows---exceeding the margin and eroding $10$--$15\%$ of gold mass at 31K---whereas two-limb NVFP4 cuts retrieval-row noise $11$--$12\times$ to $0.09$--$0.11$\,logits ($0.05$--$0.07$\,logits averaged over all heads) and retains $100\%$ of gold mass.

\subsection{Key Bias and Spectral Error Concentration}
\label{sec:spectral}

Measuring 31K score errors (\cref{fig:theory}c; \cref{app:sigma}) reveals two key-geometry pathologies under NVFP4:
(i)~\emph{Per-head key DC bias}: large constant key offsets saturate $g_K$; subtracting a prefix mean $\mu_k$ is exact under full-precision softmax and drops Qwen2.5-3B 1-limb noise from $\sigma = 0.853$ to $0.405$\,logits.
(ii)~\emph{Spectral concentration}: in the per-head key-PCA eigenbasis, the top 16 of 128 coordinates (one NVFP4 block) carry $75.1$--$79.0\%$ of key energy and $89.4$--$90.9\%$ of all 1-limb score-error energy ($90.9\%$ on Nanbeige, $89.4\%$ on Qwen2.5-3B).
Giving those 16 coordinates a second limb cuts 1-limb noise $3$--$3.7\times$ ($0.612 \to 0.204$ on Nanbeige; $0.853 \to 0.228$ on Qwen2.5-3B), below Nanbeige's $0.30$-logit top1--top2 gap; alternatively, a randomized 128-point Walsh--Hadamard transform (FWHT) disperses outliers with zero calibration and cuts unrefined cross-key correlation $2$--$3.7\times$ (\cref{tab:app_theory}).
Yet conditioning alone leaves $0.20$--$0.38$\,logits of noise that still degrades long-context generation (\cref{sec:accuracy}); combining conditioning with margin-gated two-limb refinement solves both halves.

\section{\ours{}}
\label{sec:method}

\begin{figure*}[t]
  \centering
  \safefig{0.93\textwidth}{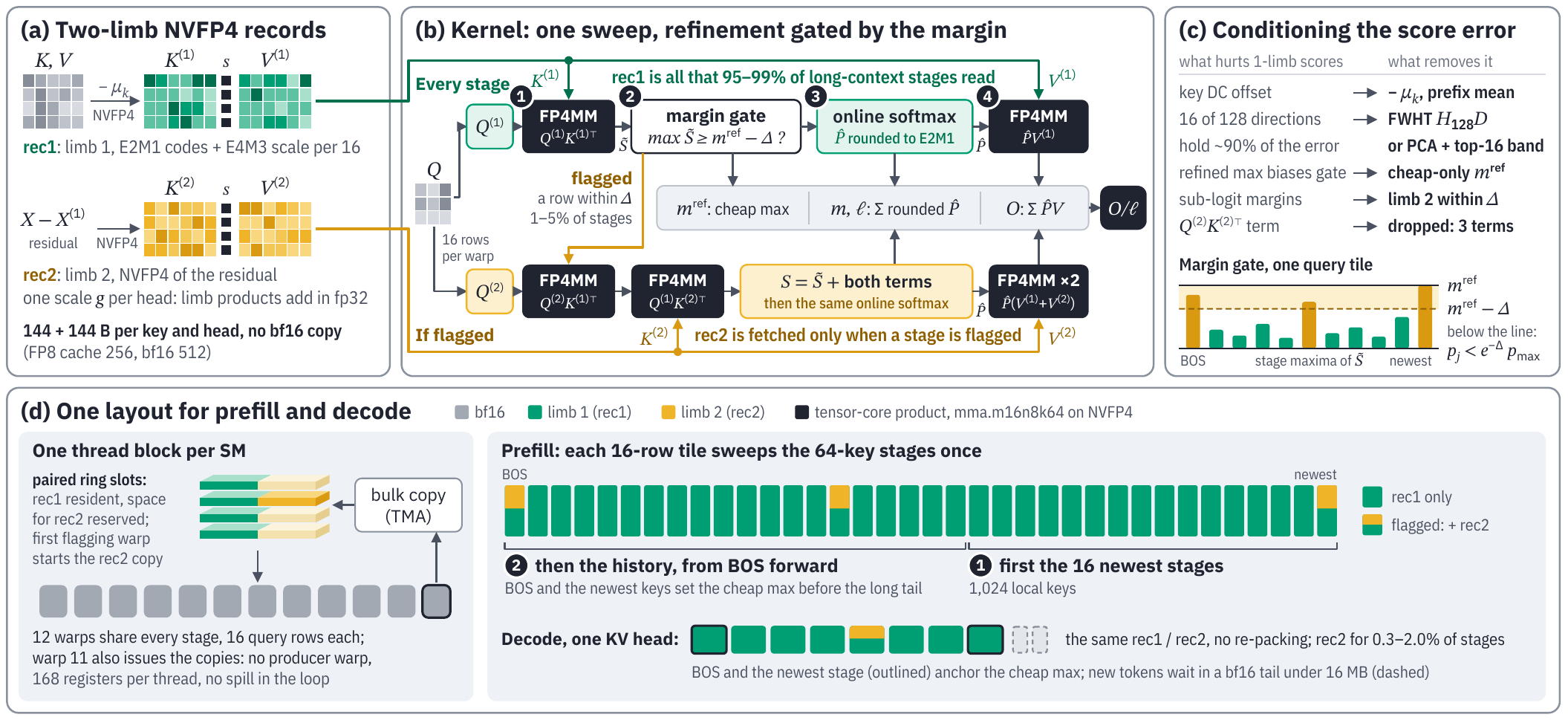}
  \vspace{-4pt}
  \caption{\textbf{Overview of \ours{} and unified prefill--decode execution.} \emph{(a)}~Shared-scale two-limb NVFP4 (\texttt{rec1}, \texttt{rec2}: $144$\,B/key each, zero FP16 backup). \emph{(b)}~Single-sweep online margin gate: stages within $\Delta$ of the cheap-only $m_r^{\mathrm{ref}}$ add the two limb-2 score terms and $\hat P V^{(2)}$ in place. \emph{(c)}~Conditioning and per-tile gate. \emph{(d)}~SM120 12-warp ring layout, recent-first then BOS-first scan, and zero-repack FlashDecoding (\ourscache{}).}
  \label{fig:architecture}
\end{figure*}

\ours{} turns \cref{sec:analysis} into a unified prefill--decode algorithm (\cref{fig:architecture}; \cref{alg:zenmargin}): shared-scale two-limb NVFP4 (\cref{sec:twolimb}), online margin gating (\cref{sec:gating}), lightweight conditioning (\cref{sec:conditioning}), and SM120 prefill and FlashDecoding kernels (\cref{sec:kernel}).

\subsection{Shared-Scale Two-Limb NVFP4: Eliminating the FP16 Backup}
\label{sec:twolimb}

Selective mixed-precision methods such as \thrift{} \citep{sharratt2026thriftattention} retain a full FP16 copy of $K,V$ alongside 4-bit tensors, forfeiting KV compression and forcing format switches.
We instead represent $X \in \{Q, K, V\}$ as two nested NVFP4 limbs sharing one per-head scale $g_X = \max |X| / (6\cdot 448)$:
\begin{equation}
  X^{(1)} = \mathcal{Q}_{\mathrm{FP4}}(X;\, g_X), \qquad
  X^{(2)} = \mathcal{Q}_{\mathrm{FP4}}\!\big(X - X^{(1)};\, g_X\big).
\label{eq:twolimb}
\end{equation}
Because both limbs share $g_X$, every cross-limb product $Q^{(a)}(K^{(b)})^{\top}$ shares the same FP32 output scale $g_Q g_K \log_2(e) / \sqrt{d}$ and accumulates directly on NVFP4 tensor cores with zero rescaling (likewise for $\hat P V^{(1)} + \hat P V^{(2)}$).

\textbf{Unified prefill--decode memory and bandwidth.}
Each 64-key stage packs into two contiguous $144$\,B/key records: \texttt{rec1} ($K^{(1)}, V^{(1)}$ and scales) and \texttt{rec2} ($K^{(2)}, V^{(2)}$)---$288$\,B/key total ($44\%$ below BF16, zero BF16 backup).
During prefill, $95$--$99\%$ of long-context stages read only \texttt{rec1}; during decoding, \ourscache{} reuses the exact $(\texttt{rec1}, \texttt{rec2})$ records with zero re-packing, reading $\approx 146$\,B/key per step ($3.5\times$ below BF16, $1.75\times$ below $256$\,B/key FP8 KV caches).
Across all heads, two-limb refinement cuts score noise $>12\times$ ($0.61$--$0.85 \to 0.05$--$0.07$\,logits; \cref{fig:theory}c), while FP4 $\hat P$ and 1-limb $V$ on unrefined stages set an all-refined error floor of $0.044$--$0.047$ (vs.\ $0.025$--$0.034$ for FP8-$V$ \sagetwo{}).

\textbf{3-term score refinement.}
Omitting the $O(\epsilon^2)$ residual $Q^{(2)}K^{(2)\top}$ changes relative output error by only $0.0004$--$0.0006$ while cutting refined score work by $25\%$ and keeping register pressure below hardware spill limits:
\begin{equation}
  S^{\mathrm{ref}} = Q^{(1)}K^{(1)\top} + Q^{(2)}K^{(1)\top} + Q^{(1)}K^{(2)\top},
\label{eq:3term}
\end{equation}
paired with $\hat P V^{(2)}$ (\oursband{} keeps all four terms on its top-16 spectral block).

\subsection{Single-Sweep Online Margin Gating}
\label{sec:gating}

Each 16-row query tile scans visible 64-key stages in a single online-softmax pass (\cref{alg:zenmargin}) using two mechanisms:
(i)~\emph{Recent-first, then BOS-first traversal:} visiting the 16 newest stages ($1{,}024$ local keys) first, then scanning earlier history from stage~$0$ (the BOS sink~\citep{xiao2024streamingllm}), seeds tight running maxima immediately and prevents spurious early refinement.
(ii)~\emph{Decoupled cheap-only reference maximum $m_r^{\mathrm{ref}}$:} comparing cheap stage maxima $\tilde m_{r,s} = \max_{j \in s} \tilde S_{r,j}$ against the refined softmax maximum $m_r$ biases the gate ($-0.024$\,nats on Qwen2.5-3B; \cref{sec:ablation,tab:app_cal}); each tile therefore tracks a dedicated cheap-only running maximum $m_r^{\mathrm{ref}} \leftarrow \max(m_r^{\mathrm{ref}}, \tilde m_{r,s})$ and flags stage~$s$ iff $\exists r:\; \tilde m_{r,s} \ge m_r^{\mathrm{ref}} - \Delta \log_2 e$.
Unflagged stages accumulate $\hat P V^{(1)}$ immediately; flagged stages refine $S$ in place via \cref{eq:3term} without recomputing $Q^{(1)}K^{(1)\top}$ and fold $\hat P(V^{(1)}+V^{(2)})$ into the same online softmax state.

\subsection{Conditioning Against Correlated Score Error}
\label{sec:conditioning}

To eliminate key DC bias and suppress the cross-key correlation term in \cref{eq:p2law}, both variants apply exact causal key centering ($\mu_k = \frac{1}{P}\sum_{t<P} k_t$, $P{=}\min(T, 2048)$, or offline $\mu_k$) inside packing at zero extra HBM cost:
(1)~\emph{\oursfwht{} (recommended default):} applies a fixed orthogonal 128-point randomized Walsh--Hadamard transform $H_{128} D_{\mathrm{sign}}/\sqrt{128}$ to $Q$ and centered $K$ in registers during packing---preserving inner products exactly, requiring zero calibration, adding $\le 0.01$\,ms overhead across 8K--128K, and remaining invariant under out-of-distribution RoPE length extrapolation (\cref{sec:accuracy}).
(2)~\emph{\oursband{} (spectral tier):} calibrates per-head key-PCA rotations $R_h$ and Q/K balancing scales $s_{h,c} = (\mathrm{rms}(k'_{h,c})/\mathrm{rms}(q'_{h,c}))^{1/2}$ once offline on 1--6 disjoint prompts ($5.5$--$7.4$\,s; zero-shot cross-domain transfer, \cref{sec:ablation}), placing the top 16 principal directions in the final 16-channel block and adding a lightweight residual correction only when stages fall within $\Delta_{\mathrm{band}}\!\in\!\{2,4\}$ logits of $m_r^{\mathrm{ref}}$.

\textbf{Causal consistency scope.}
Exact attention-graph causality (masking $j > i$) is distinct from token-level quantization-parameter causality: because NVFP4 quantization is nonlinear, $\mathcal{Q}(k_j - \mu_k) \neq \mathcal{Q}(k_j) - \mu_k$, so prefix-2048 $\mu_k$, prompt-level global scales $g_X = \max|X|/(6{\cdot}448)$, and hardware 16-key $V$ microscale tiles couple tokens within the prefill prompt (note that in standard prompt-prefill-then-decode serving and all RULER/LongBench evaluations, the full prompt is observed upfront and zero generated tokens ever feed back into prefix statistics).
For workloads requiring strict token-by-token $Q/K$ causality, bit-exact multi-turn prefix-cache invariance, or bounded-tail long-output generation, NVFP4's E4M3 microscale spans nearly 18 bits of dynamic range ($[2^{-9}, 448]$), enabling offline static per-head $(\mu_k^{\mathrm{off}}, g_Q^{\mathrm{off}}, g_K^{\mathrm{off}}, g_V^{\mathrm{off}})$ calibrated on 4 disjoint prompts ($2.8$--$6.2$\,s with $2\times$ safety headroom).
In \cref{tab:app_causal_audit}, this strictly token-causal static regime incurs $0.000\%$ E4M3 clipping on both held-out LongBench prompts (matching prompt-level error and likelihood) and 3K--4K generated Long-CoT math trajectories ($0/32.2\text{M}$ $K$ blocks and $1/32.1\text{M}$ $V$ blocks on DeepSeek-R1-Distill-Llama-8B calibrated zero-shot on 4 \texttt{gov\_report} documents), allowing \ourscache{} to flush every 64 generated tokens directly into preallocated $(\texttt{rec1}, \texttt{rec2})$ slots without repacking any historical KV stages.

\subsection{Unified Prefill and FlashDecoding System Co-Design}
\label{sec:kernel}
\label{sec:decode}

Realizing NVFP4's throughput requires aligning the algorithm with Blackwell's memory and register hierarchy across prefill and decoding (full SASS/TMA profiling, optimization trajectory, and kernel resource tables appear in \cref{app:kernel_details}).

\textbf{Single-sweep asynchronous prefill.}
An initial 4-warp \oursband{} prototype trailed 8-bit attention ($20.78$ vs.\ $16.43$\,ms on 31K Nanbeige) due to load pressure, two-pass \texttt{rec1} re-reads, $V$-tile bank conflicts, and spills.
Our final SM120 kernel eliminates all bottlenecks with zero hot-loop spills:
(i)~12 balanced warps ($192$ query rows/CTA, one warp self-producing \texttt{rec1} into a paired ring buffer) cut global traffic $3\times$ within the 168-register ceiling;
(ii)~\emph{First-Flagger-Fetches} reserves shared-memory space for \texttt{rec2} beside \texttt{rec1}, letting the first flagging warp fetch \texttt{rec2} asynchronously while flagging warps overlap $Q^{(2)}K^{(1)\top}$ on resident \texttt{rec1};
(iii)~key centering, packing, and 128-point FWHT fuse in registers ($\le 0.01$\,ms overhead; $1.8\%$ of total time at 128K).
As an isomorphic 8-bit yardstick, our 12-warp INT8-$QK^{\top}$ / NVFP4-$\widetilde{P}V$ kernel (\kwide{}, $204$\,B/key) reaches $746$--$783$\,TFLOPS ($1.3\times$ \sagetwo{}).

\textbf{Unified Two-Limb FlashDecoding and \ourscache{}.}
Our split-KV FlashDecoding kernel operates directly on prefill's $(\texttt{rec1}, \texttt{rec2})$ records with zero re-quantization, parallelizing across $(B{\cdot}H)$ KV heads (4 warps/CTA, barrier-free loop), sharing each decoded \texttt{rec1} key across $G$ queries, anchoring $m^{\mathrm{ref}}$ from stage~$0$ (BOS) and the final stage, and fetching \texttt{rec2} only on the $0.3$--$2.0\%$ flagged stages---reaching $1{,}131$--$1{,}560$\,GB/s effective bandwidth ($1.31$--$1.48\times$ faster than production FlashInfer XQA FP8 at 64K and $2.0$--$2.3\times$ our isomorphic FP8 baseline; \cref{app:decode_micro}).
Wrapped as a drop-in HuggingFace \texttt{Cache} (\ourscache{}), prefill immediately frees BF16 $K,V$ per layer and retains $(\texttt{rec1}, \texttt{rec2}, g_K, g_V, \mu_k)$ ($1.78\times$ smaller than BF16) plus a bounded $<64$-token BF16 tail buffer ($8$\,MB/seq across 32 layers) merged via exact logsumexp in the $-\mu_k$ gauge.
For long-output generation, pairing \ourscache{} with offline static scales (\cref{sec:conditioning}) flushes every 64 generated tokens in place into preallocated $(\texttt{rec1}, \texttt{rec2})$ slabs via a fused CUDA chunk packer (bit-identical to full-sequence static packing; $0.011$\,ms/layer at $B{=}16$, i.e., $0.003$--$0.005$\,ms/step amortized across 32 layers) with zero historical KV repacking.

\section{Experiments}
\label{sec:exp}

\subsection{Experimental Setup}
\label{sec:setup}

\textbf{Models, benchmarks, and baselines.}
We evaluate six open LLMs ($d{=}128$, $G\!\in\!\{3,4,6,7,8\}$):
Llama-3.1-8B-Instruct ($32$L, $H{=}8, G{=}4$) \citep{grattafiori2024llama3},
DeepSeek-R1-Distill-Llama-8B ($32$L, $H{=}8, G{=}4$; Long-CoT incremental-flush serving and clipping audit),
Qwen2.5-7B-Instruct ($28$L, $H{=}4, G{=}7$) and Qwen2.5-3B-Instruct ($36$L, $H{=}2, G{=}8$) \citep{qwen2024qwen25},
Phi-4-mini-instruct ($32$L, $H{=}8, G{=}3$) \citep{abouelenin2025phi4mini},
and Nanbeige4.2-3B ($44$ slots, $H{=}8, G{=}6$) \citep{yang2026nanbeige42}.
Standardized autoregressive evaluation covers \emph{RULER} \citep{hsieh2024ruler} (six core tasks across all models plus the \emph{full 13-task suite}---adding \texttt{niah\_s3/mk1/mk3}, diffuse aggregation \texttt{cwe/fwe}, and multi-doc \texttt{qa\_1/qa\_2}---on Llama-3.1-8B at 16K/32K) and \emph{official LongBench} \citep{bai2024longbench} (10 English tasks, $300$ prompts; 6 QA/retrieval tasks = 180 on Phi-4-mini), plus diffuse/code kernel stress profiling and teacher-forced diagnostics (\cref{app:multislot,app:tf_eval}).
All experiments run on RTX~5090 GPUs (SM120, 32\,GB, 575\,W) against PyTorch SDPA (FA2, cuDNN), official \sagetwo{} (v2.2.0, GQA), official \thrift{} CUDA kernels (raw and +centering/FWHT), production FlashInfer 0.7.6 SM120 kernels (\texttt{nvfp4\_attention\_sm120} causal GQA prefill and SM120 XQA BF16/FP8/NVFP4 paged decode), emulated \deco{} and ResQ, and isomorphic 12-warp \kwide{} (\cref{app:kernel_details}); all \ours{} and \kwide{} timings are E2E (centering, FWHT/rot, pack, attend).

\textbf{Forensic note on \sagethree{} causal mode.}
Official \sagethree{} (\texttt{sageattn3\_blackwell} v1.0.0) works non-causally on unit Gaussians $\mathcal{N}(0,0.5)$ ($0.144$ rel.\ err) but incurs $0.61$--$0.68$ error with \texttt{is\_causal=True} regardless of \texttt{per\_block\_mean}, centering, or FWHT (\cref{tab:app_sage3}), yielding $0$--$3.7\%$ RULER.
We therefore use our verified causal-exact 1-limb NVFP4 kernels (centered-only and +FWHT) as the true uniform FP4 accuracy baselines.

\subsection{Real-Activation Speed, Flag Distribution, and TTFT}
\label{sec:speed}

\begin{table*}[t]
\caption{\textbf{Same-GPU attention benchmark on \emph{real} LLM activations} (RTX~5090, SM120): Qwen2.5-7B slot~12 ($31.2\text{K}$, $6.96$\,TF), Nanbeige slot~16 ($31.3\text{K}$, $12.04$\,TF), and Llama-3.1-8B 64K ($65.5\text{K}$, $35.18$\,TF, 8-layer mean across L0,4,\dots,28 on real \texttt{gov\_report} prose). All \ours{} and \kwide{} rows are true E2E; \sagetwo{} uses native GQA.}
\label{tab:sota_speed}
\centering
\small
\setlength{\tabcolsep}{3.4pt}
\resizebox{\textwidth}{!}{%
\begin{tabular}{@{}lrrr@{\hspace{8pt}}rrr@{\hspace{8pt}}rrrrr@{}}
\toprule
 & \multicolumn{3}{c}{Qwen2.5-7B (31.2K real)} & \multicolumn{3}{c}{Nanbeige-3B (31.3K real)} & \multicolumn{5}{c}{Llama-3.1-8B (64K real, 8-layer mean)} \\
\cmidrule(lr){2-4}\cmidrule(lr){5-7}\cmidrule(l){8-12}
Method & ms & $\times$FA2 & Err & ms & $\times$FA2 & Err & ms & TF/s & $\times$FA2 & $\times$SA2++ & Err \\
\midrule
SDPA FlashAttn-2 (BF16) & 32.98 & $1.00\times$ & 0.000 & 55.73 & $1.00\times$ & 0.000 & 162.92 & 216 & $1.00\times$ & $0.37\times$ & 0.000 \\
SDPA cuDNN (BF16) & 30.71 & $1.07\times$ & 0.000 & 52.21 & $1.07\times$ & 0.000 & 152.21 & 231 & $1.07\times$ & $0.40\times$ & 0.000 \\
\thrift{} (10\% / 25\% FP16) & --- & --- & --- & 24.05\,/\,--- & $2.32\times$ & 0.164 & 72.37\,/\,90.21 & 486\,/\,390 & $2.25$/$1.81\times$ & $0.84$/$0.67\times$ & .073/.048 \\
\thrift{} (5\% FP16) & --- & --- & --- & 20.48 & $2.72\times$ & 0.182 & 66.44 & 530 & $2.45\times$ & $0.91\times$ & 0.089 \\
\sagetwo{} (native GQA) & 12.49 & $2.64\times$ & 0.032 & 20.93 & $2.66\times$ & 0.025 & 60.67 & 580 & $2.69\times$ & $1.00\times$ & 0.034 \\
\sagethree{} FP4 (\texttt{pbm=1/0})$^\dagger$ & 12.82\,/\,11.41 & $2.57$/$2.89\times$ & \textcolor{tierwarn}{0.605} & 21.71\,/\,19.25 & $2.57$/$2.89\times$ & \textcolor{tierwarn}{0.473} & 58.19\,/\,49.90 & 605\,/\,705 & $2.80$/$3.27\times$ & $1.04$/$1.22\times$ & \textcolor{tierwarn}{0.555} \\
FlashInfer NVFP4 (\texttt{pbm=1/0}, E2E)$^\S$ & --- & --- & --- & --- & --- & --- & 52.32\,/\,42.05 & 673\,/\,837 & $3.02$/$3.76\times$ & $1.16$/$1.44\times$ & .140/.185 \\
\kwide{} (12w, E2E) & 9.34 & $3.53\times$ & 0.080 & 16.04 & $3.47\times$ & 0.064 & 46.55 & 756 & $3.50\times$ & $1.30\times$ & 0.093 \\
\midrule
\ours{} (1-limb+FWHT, E2E) & 6.99 & $4.72\times$ & 0.207 & 11.75 & $4.74\times$ & 0.245 & 33.83 & 1040 & $4.82\times$ & $1.79\times$ & 0.224 \\
\rowcolor{tierfour!10}
\textbf{\oursfwht{}} ($\Delta{=}1$, E2E) & \textbf{7.90} & $\mathbf{4.17\times}$ & \textbf{0.045} & \textbf{13.89} & $\mathbf{4.01\times}$ & \textbf{0.056} & \textbf{37.16} & \textbf{947} & $\mathbf{4.38\times}$ & $\mathbf{1.63\times}$ & \textbf{0.063} \\
\rowcolor{tierfour!10}
\textbf{\oursfwht{}} ($\Delta{=}2$, E2E) & 8.17 & $4.04\times$ & \textbf{0.042} & 14.81 & $3.76\times$ & \textbf{0.046} & 37.86 & 929 & $4.30\times$ & $1.60\times$ & \textbf{0.057} \\
\rowcolor{tierband!10}
\textbf{\oursband{}} ($\Delta{=}1$, $\Delta_b{=}4$, E2E) & 9.14 & $3.61\times$ & \textbf{0.042} & 16.64 & $3.35\times$ & \textbf{0.045} & 40.42$^\ddagger$ & 870 & $4.04\times$ & $1.49\times$ & --- \\
\bottomrule
\end{tabular}}
\\[1pt]{\footnotesize $^\dagger$\sagethree{} KV expansion untimed; high error reflects its causal-mode bug (\cref{sec:setup}). $^\S$Production FlashInfer 0.7.6 \texttt{nvfp4\_attention\_sm120} (causal, native GQA, E2E incl.\ \texttt{quantize\_qkv}; kernel-only $40.82$/$40.13$\,ms, $\times$FA2 vs.\ same-venv FA2 $158.01$\,ms). $^\ddagger$Real Llama-3.1-8B L16 ($64\text{K}$).}
\end{table*}

\Cref{tab:sota_speed} and \cref{fig:hero}a--b report same-GPU E2E speed and error on \emph{real} activations.
On real Llama-3.1-8B 64K (8-layer mean), official \thrift{}~5\% takes $66.44$\,ms ($2.45\times$ FA2)---\emph{slower} than dense 8-bit \sagetwo{} ($60.67$\,ms, $2.69\times$)---while production FlashInfer 0.7.6 SM120 NVFP4 (\texttt{nvfp4\_attention\_sm120}, causal, native GQA) takes $52.32$/$42.05$\,ms E2E ($673$/$837$\,TFLOPS, err $0.140$/$0.185$).
By contrast, \oursfwht{} ($\Delta{=}1$/$\Delta{=}2$) takes $37.16$/$37.86$\,ms E2E ($947$/$929$\,TFLOPS): $4.38\times$/$4.30\times$ FA2, $1.63\times$/$1.60\times$ \sagetwo{}, $1.79\times$ \thrift{}~5\%, $1.25\times$ \kwide{}, and $\mathbf{1.13}$--$\mathbf{1.41\times}$ faster E2E than production FlashInfer SM120 NVFP4 with $\mathbf{2.2}$--$\mathbf{2.9\times}$ lower output error ($0.063$ vs.\ $0.140$/$0.185$); for reference, QAT-trained Attn-QAT \citep{zhang2026attnqat} reports $760$\,TFLOPS E2E at non-causal 32K on RTX~5090 (5090 kernel unreleased), whereas training-free causal \ours{} achieves $947$--$967$\,TFLOPS E2E ($1{,}040$\,TFLOPS 1-limb+FWHT).
Across all 32 layers of Llama-3.1-8B at 64K (\cref{app:multislot}), warp flag rates average $1.28\%$ at $\Delta{=}1$ (median $1.25\%$, p90 $1.61\%$, max $2.50\%$ on L0) and $2.16\%$ at $\Delta{=}2$ (max $9.29\%$ on L0); even worst-case L0 ($38.29$/$41.52$\,ms) beats \kwide{} ($47.07$\,ms) and \sagetwo{} ($62.10$\,ms).
\textbf{No degradation cliff on diffuse or code workloads:} profiling six real workloads on Llama-3.1-8B (\cref{tab:app_diffuse_profile})---peaked retrieval (\texttt{niah\_mk2}), diffuse common/Zipf word aggregation (\texttt{cwe}, \texttt{fwe}), multi-doc QA (\texttt{qa\_2}), dense prose (\texttt{gov\_report}), and repetitive long-context code (\texttt{lcc})---shows E2E latency at 64K stays flat within $0.9\%$ across every workload ($36.99$--$37.32$\,ms, $\mathbf{943}$--$\mathbf{951}$\,TFLOPS, $4.33$--$4.40\times$ FA2, $1.62$--$1.64\times$ \sagetwo{}, warp flag $1.06$--$1.74\%$, CTA limb-2 fetch $2.76$--$4.28\%$, rel.\ err $0.057$--$0.064$; all-32-layer 32K means $9.33$--$9.57$\,ms, $919$--$943$\,TFLOPS, warp $1.79$--$3.45\%$, CTA $4.10$--$8.08\%$).

\textbf{Real 4K--128K scaling and full-model TTFT.}
On real Llama-3.1-8B L16 from $4\text{K}$ to $128\text{K}$ (\cref{fig:hero}b; \cref{tab:scaling,tab:pack_scale}), flag rates drop from $18.3\%$ (4K) to $1.2\%$ (64K) and $0.7\%$ (128K); in-register FWHT adds $\le 0.01$\,ms, while total packing shrinks from $19.3\%$ (8K; $17.2\%$ at 4K) to $3.69\%$ (64K) and $1.78\%$ (128K, $967.1$\,TFLOPS, $4.44\times$ FA2, $1.62\times$ \sagetwo{}).
Full-model 64K prefill TTFT (\cref{fig:serving}a; \cref{tab:app_ttft}) speeds up $1.67\times$ on Llama-3.1-8B ($9.95 \to 5.97$\,s vs.\ $6.90$\,s \thrift{}~5\%, $6.78$\,s \sagetwo{}, $6.31$\,s \kwide{}), $1.57\times$ on Qwen2.5-7B, $1.82\times$ on Qwen2.5-3B, and $1.91\times$ on Phi-4-mini ($5.4$--$6.0\%$ E2E over \kwide{}).

\subsection{Standardized Autoregressive Generation Accuracy}
\label{sec:accuracy}

\begin{table*}[t]
\caption{\textbf{Standardized autoregressive RULER and official LongBench across four LLMs} ($G\!\in\!\{3,4,7,8\}$; RULER $n{=}20$/task at 8K--32K, $30$/task = 180 on Llama-8B R-48K, $15$/task on Phi-4/Qwen-3B; LB-10 $300$ prompts; LB-6 $180$). Uniform 1-limb FP4 (even +FWHT) and raw \thrift{} degrade on hard retrieval, whereas \oursfwht{} recovers BF16 accuracy within paired bootstrap $95\%$ CIs (\cref{tab:app_bootstrap}).}
\label{tab:gen_main}
\centering
\small
\setlength{\tabcolsep}{2.7pt}
\resizebox{\textwidth}{!}{%
\begin{tabular}{@{}l|cccc|c||ccc|c||ccc||cc@{}}
\toprule
 & \multicolumn{5}{c||}{\textbf{Llama-3.1-8B} ($G{=}4$)} & \multicolumn{4}{c||}{\textbf{Qwen2.5-7B} ($G{=}7$)} & \multicolumn{3}{c||}{\textbf{Phi-4-mini} ($G{=}3$)} & \multicolumn{2}{c}{\textbf{Qwen-3B} ($G{=}8$)} \\
\cmidrule(lr){2-6}\cmidrule(lr){7-10}\cmidrule(lr){11-13}\cmidrule(l){14-15}
Method & R-8K & R-16K & R-32K & R-48K$^\ast$ & LB-10 & R-8K & R-16K & R-32K$^\dagger$ & LB-10 & R-16K & R-32K & LB-6 & R-16K & R-32K \\
\midrule
BF16 SDPA & 93.50 & 94.38 & 99.79 & 99.86 & 49.14 & 90.17 & 89.62 & 82.21 & 45.37 & 98.06 & 98.33 & 33.86 & 73.94 & 70.39 \\
\sagetwo{} (GQA) & 94.54 & 95.00 & 99.79 & 100.00 & 49.22 & 90.33 & 89.96 & 80.62 & 45.19 & 97.50 & 97.83 & 34.37 & 77.78 & 70.33 \\
\kwide{} (12w 8-bit) & 96.04 & 95.83 & 99.79 & 99.72 & 49.05 & 90.33 & 89.33 & 82.46 & 45.64 & 96.11 & 95.00 & 33.17 & 75.83 & 66.67 \\
\thrift{} (5\% FP16, raw) & --- & 96.67 & 98.54 & 98.36 & 49.50 & \textcolor{tierwarn}{0.62} & \textcolor{tierwarn}{0.00} & \textcolor{tierwarn}{1.67} & \textcolor{tierwarn}{3.78} & 96.11 & 95.61 & 32.55 & \textcolor{tierwarn}{43.11} & \textcolor{tierwarn}{11.39} \\
\thrift{} (5\% + ctr + FWHT) & --- & --- & --- & 99.44 & --- & 89.00 & 88.17 & 72.67 & 45.17 & 96.67 & 96.94 & 31.20 & 70.00 & 67.33 \\
\midrule
\ours{} (1-limb, ctr only) & 94.50 & 92.12 & 93.92 & 90.83 & 47.53 & 89.58 & 82.04 & 46.79 & 44.18 & 87.78 & 84.72 & 24.68 & 57.17 & 60.56 \\
\ours{} (1-limb + ctr + FWHT) & --- & --- & --- & 95.50 & 48.10 & 89.96 & 85.62 & 67.12 & 45.60 & 91.06 & 89.22 & 29.68 & 69.11 & 64.72 \\
\rowcolor{tierfour!10}
\textbf{\oursfwht{}} ($\Delta{=}1$) & 94.21 & 95.83 & 99.38 & \textbf{100.00} & 49.80 & 90.00 & 88.79 & \textbf{81.71} & 45.36 & 96.94 & 96.72 & 33.42 & 76.61 & \textbf{70.44} \\
\rowcolor{tierfour!10}
\textbf{\oursfwht{}} ($\Delta{=}2$) & 94.50 & 95.83 & 99.38 & \textbf{100.00} & \textbf{49.82} & 89.83 & \textbf{90.00} & 80.08 & \textbf{45.39} & 97.22 & \textbf{97.28} & \textbf{35.30} & \textbf{77.28} & 68.72 \\
\rowcolor{tierband!10}
\textbf{\oursband{}} ($\Delta{=}1$, b4) & 94.67 & 95.62 & 99.38 & \textbf{100.00} & 48.95 & 89.96 & 88.25 & 64.12$^\dagger$ & 44.91 & \textbf{97.78} & 96.17 & 31.81 & 75.22 & 69.39 \\
\bottomrule
\end{tabular}}
\\[1pt]{\footnotesize $^\ast$Llama-8B R-48K high-power paired run ($n{=}30$/task = 180 prompts) for BF16, \sagetwo{}, \thrift{}~5\% (raw/+FWHT), 1-limb+FWHT, and \oursfwht{} ($\Delta{=}1$); \kwide{}, 1-limb ctr, $\Delta{=}2$, and \oursband{} use $n{=}15$/task. $^\dagger$Qwen-7B R-32K averages $38.7\text{K}$ tokens (OOD RoPE).}
\end{table*}

\Cref{tab:gen_main} summarizes autoregressive fidelity across four LLMs (teacher-forced diagnostics in \cref{app:tf_eval}).
(1)~\emph{Conditioning alone is statistically insufficient:} on high-power Llama-8B R-48K ($n{=}180$ prompts; \cref{tab:app_bootstrap}), unrefined 1-limb+FWHT drops to $95.50\%$ (\texttt{mk2} $83.3\%$, paired difference $-4.36\%$ vs.\ BF16 $99.86\%$, $95\%$ CI $[-7.33, -1.83]$, $p{=}0.0002$); similarly, 1-limb+FWHT drops $-7.0$ to $-9.1$ pts on Phi-4-mini RULER ($91.06$/$89.22\%$ vs.\ $98.06$/$98.33\%$; $-4.18$ LB-6), $-15.1$ pts on Qwen-7B R-32K ($67.12\%$ vs.\ $82.21\%$), and $-4.8$ to $-5.7$ pts on Qwen-3B.
(2)~\emph{Unconditioned vs.\ conditioned \thrift{}:} raw \thrift{}~5\% collapses on Qwen-7B ($0.0$--$1.7\%$ RULER, $3.78$ LB-10) and Qwen-3B ($43.1$/$11.4\%$), and significantly trails BF16 on Llama-8B R-48K ($98.36\%$, diff $-1.50\%$, $95\%$ CI $[-3.33, -0.14]$, $p{=}0.0342$).
Adding our centering+FWHT rescues Qwen models, yet its power-of-two bulk softmax still drops to $72.67\%$ on Qwen-7B R-32K (\texttt{mk2} $20\%$ vs.\ $81.71\%$ \oursfwht{}), trails Phi-4 LB-6 ($31.20$ vs.\ $33.86$), and runs slower than both \sagetwo{} and \oursfwht{} ($1{,}161$\,s vs.\ $950$\,s and $875$\,s on Llama-8B R-48K) while keeping an FP16 backup.
(3)~\emph{\ours{} recovers BF16 accuracy with paired bootstrap parity:} \oursfwht{} ($\Delta{=}1$) scores $100.00\%$ on Llama-8B R-48K ($n{=}180$; diff $+0.14\%$, $95\%$ CI $[0.00, +0.42]$, $p{=}0.726$, in $875$\,s vs.\ $1{,}290$\,s BF16---just $+1.7\%$ over 1-limb+FWHT's $860$\,s!) and $49.80$ on LB-10 ($n{=}300$; diff $+0.66$, $95\%$ CI $[-1.40, +2.74]$, $p{=}0.536$; unified \ourscache{} $49.74$, diff $+0.60$, $95\%$ CI $[-1.18, +2.45]$, $p{=}0.497$).
(4)~\emph{OOD RoPE invariance:} on Qwen-7B R-32K ($38.7\text{K}$), 6K-calibrated \oursband{} drops to $64.12\%$, whereas calibration-free \oursfwht{} stays at $81.71\%$ (vs.\ $82.21\%$ BF16), establishing \oursfwht{} as default.
(5)~\emph{Full 13-task RULER (incl.\ diffuse \texttt{cwe/fwe} and multi-doc \texttt{qa\_1/qa\_2}):} on Llama-3.1-8B across all 13 RULER tasks at 16K ($n{=}20$/task = 260 prompts) and 32K ($n{=}15$/task = 195 prompts; \cref{tab:app_ruler13}), \oursfwht{} ($\Delta{=}1$) matches BF16 within paired bootstrap $95\%$ CIs at both lengths ($83.63\%$ vs.\ $83.23\%$ at 16K, diff $+0.40\%$, $p{=}0.677$, $433$\,s vs.\ $504$\,s BF16 / $447$\,s \sagetwo{}; $86.50\%$ vs.\ $86.85\%$ at 32K, diff $-0.36\%$, $p{=}0.514$, $634$\,s vs.\ $850$\,s BF16 [$1.34\times$] / $670$\,s \sagetwo{}), matching BF16 verbatim on 32K \texttt{NIAH-8} ($97.92\%$), \texttt{fwe} ($84.4\%$), \texttt{qa\_1} ($93.3\%$), and \texttt{qa\_2} ($66.7\%$) while diffuse and QA tasks trigger \emph{lower} warp/CTA flag rates than peaked \texttt{niah\_mk2} ($2.67$--$4.76\%$ vs.\ $6.41\%$ warp at 32K).

\subsection{Unified Serving: \ourscache{} vs.\ Production FlashInfer XQA}
\label{sec:serving}

\begin{figure*}[t]
  \centering
  \safefig{0.92\textwidth}{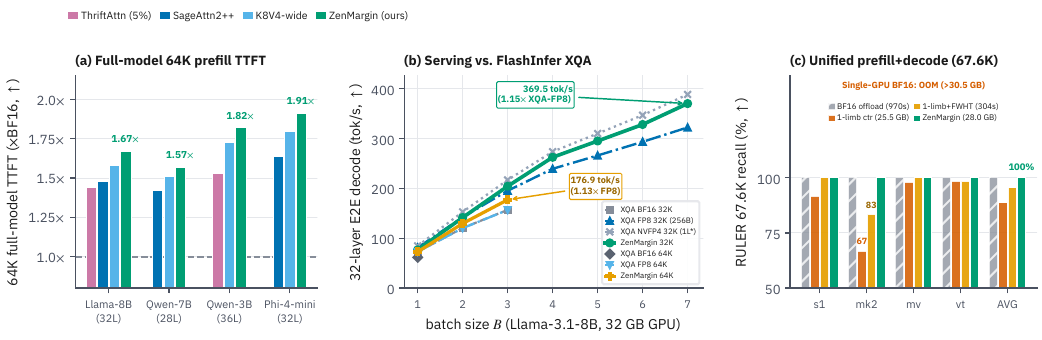}
  \vspace{-4pt}
  \caption{\textbf{Full-model prefill TTFT, 32-layer serving vs.\ production FlashInfer 0.7.6 SM120 XQA, and 67.6K unified RULER} (32\,GB RTX~5090). \emph{(a)}~64K full-model TTFT speed-up ($1.57$--$1.91\times$). \emph{(b)}~Real 32-layer Llama-3.1-8B batched serving in an identical zero-HF-overhead decode loop (real 32-layer \texttt{gov\_report} KV, flag rate $0.54$--$1.04\%$): \ourscache{} ($\Delta{=}1$) outperforms production FlashInfer XQA BF16 ($2.36\times$/$2.87\times$ peak) and XQA FP8 ($1.15\times$ at 32K $B{=}7$, $1.13\times$ at 64K $B{=}3$), reaching $94$--$95\%$ of 1-limb decode-only XQA NVFP4 throughput ($^*$1-limb XQA NVFP4 incurs $9$--$13\times$ higher decode error and requires KV repacking after prefill). \emph{(c)}~67.6K unified RULER ($48$ prompts): single-GPU BF16 OOMs, exact CPU-offloaded BF16 scores $100\%$, unified 1-limb+FWHT drops to $95.4\%$ (\texttt{mk2} $83.3\%$), and \ourscache{} fits in $28.0$\,GB at $100.0\%$ ($3.1\times$ faster than offload).}
  \label{fig:serving}
\end{figure*}

While \ourscache{} stores $12.5\%$ more KV bytes than an FP8 cache ($288$ vs.\ $256$\,B/key; both well below BF16's $512$\,B/key), it reads only $\approx 145$--$146$\,B/key per decode step ($1.75\times$ less HBM traffic than FP8) with zero prefill-to-decode repacking (\cref{fig:serving}b--c; full numbers in \cref{tab:app_serving}).
\textbf{32-layer serving vs.\ production FlashInfer 0.7.6 SM120 XQA:} to eliminate HuggingFace wrapper overhead and compare directly against production SM120 decode kernels, we benchmark all 32 Llama-3.1-8B layers in an identical zero-HF-overhead decode harness on one 32\,GB RTX~5090 using real 32-layer \texttt{gov\_report} KV caches (\ourscache{} real flag rate $1.04\%$ at 32K, $0.54\%$ at 64K):
(i)~At 32K, \ourscache{} ($\Delta{=}1$) beats production FlashInfer XQA BF16 ($156.6$\,tok/s peak, $B{=}3$) by $1.30\times$ at matched $B{=}3$ ($204.0$\,tok/s) and $\mathbf{2.36\times}$ at peak ($B{=}7$, $369.5$\,tok/s), while beating production FlashInfer XQA FP8 ($256$\,B/key) at every batch size $B{=}3$--$7$ ($1.05$--$1.15\times$; even $B{=}6$ at $327.6$\,tok/s beats XQA FP8's $B{=}7$ peak $321.7$\,tok/s).
(ii)~At 64K, both \ourscache{} and XQA FP8 cap at $B{=}3$, where \ourscache{} reaches $176.9$\,tok/s vs.\ $156.5$\,tok/s for XQA FP8 ($\mathbf{1.13\times}$ XQA FP8, $\mathbf{2.87\times}$ XQA BF16 $61.6$\,tok/s).
(iii)~Compared to production SM120-exclusive 1-limb XQA NVFP4 ($144$\,B/key, decode-only), \ourscache{} achieves $94$--$95\%$ of XQA NVFP4's throughput at matched batch sizes ($369.5$ vs.\ $387.8$\,tok/s at 32K $B{=}7$; $176.9$ vs.\ $187.3$\,tok/s at 64K $B{=}3$) while cutting single-layer decode error $\mathbf{9.2}$--$\mathbf{13.2\times}$ ($0.019$--$0.022$ vs.\ $0.176$--$0.289$; also $2.6$--$3.0\times$ lower than XQA FP8's $0.050$--$0.065$), avoiding XQA NVFP4's post-prefill KV repacking ($15.1$\,ms/32L at 64K), and accelerating prefill $4.38\times$ in the same layout (while XQA NVFP4 trades accuracy for a higher batch ceiling, up to $B{=}10$ at 32K / $B{=}5$ at 64K).
In standard HuggingFace \texttt{DynamicCache} end-to-end serving (\cref{tab:app_serving}), \ourscache{} reaches $298.5$/$161.0$\,tok/s ($3.55\times$/$4.38\times$ HF BF16 SDPA, $1.08\times$/$1.26\times$ our isomorphic \texttt{Fp8Cache}).
\textbf{Unified prefill+decode accuracy:} attending strictly to \ourscache{}, Llama-3.1-8B scores $100.0\%$ on RULER 16K/32K, matches exact CPU-offloaded BF16 ($100.0\%$, $970$\,s) at $67.6\text{K}$ in $27.97$\,GB ($317$\,s, $3.1\times$ faster; GPU BF16 OOMs, unified 1-limb+FWHT $95.42\%$), and scores $49.74$ on LB-10 (vs.\ $49.14$ BF16, paired $p{=}0.497$; 1-limb FP4 $47.28$).
\textbf{Bounded-tail incremental flush for Long-CoT Test-Time Scaling:} coupling \ourscache{} with zero-shot static scales ($0.0000\%$ $K$ clipping on real 3K--4K AIME trajectories; \cref{tab:app_causal_audit}) flushes every 64 generated tokens in place at $0.003$--$0.005$\,ms/step amortized overhead while keeping the BF16 tail strictly $<64$ tokens ($8$\,MB/seq across 32 layers).
On real 32-layer DeepSeek-R1-Distill-Llama-8B Long-CoT KV on a single 32\,GB RTX~5090 (\cref{tab:app_tts_serving}), \ourscache{} ($\Delta{=}1$, incl.\ amortized flush) scales to $B{=}26$ at 8K ($\mathbf{1{,}384.6}$\,tok/s) and $B{=}13$ at 16K ($\mathbf{684.8}$\,tok/s)---$\mathbf{2.15\times}$/$\mathbf{2.10\times}$ production FlashInfer XQA BF16 peak ($645.1$/$326.0$\,tok/s at $B{=}14$/$7$), $\mathbf{1.05\times}$/$\mathbf{1.06\times}$ XQA FP8 peak ($1{,}317.6$/$645.8$\,tok/s even though FP8 fits $B{=}28$/$14$; $\mathbf{1.09}$--$\mathbf{1.14\times}$ faster at matched $B$), and $95$--$100\%$ of 1-limb XQA NVFP4 throughput at matched $B{=}12$--$26$ ($1{,}303.5$ vs.\ $1{,}300.8$\,tok/s at 8K $B{=}24$).

\subsection{Factorial Controls and Systems Analysis}
\label{sec:ablation}

\begin{table*}[t]
\caption{\textbf{2$\times$2$\times$2 factorial control isolation on Qwen2.5-3B RULER 19.6K} ($n{=}15$/task, $60$ prompts) and \textbf{\sagethree{} causal-mode forensic diagnosis}. Left: cleanly isolates conditioning, selection rule, and refinement precision. Right: official \texttt{sageattn3\_blackwell} v1.0.0 works non-causally ($0.145$) but incurs $0.61$--$0.63$ causal error even on clean unit Gaussians regardless of \texttt{per\_block\_mean}, centering, or FWHT.}
\label{tab:app_factorial}
\label{tab:app_sage3}
\centering
\small
\setlength{\tabcolsep}{3.0pt}
\resizebox{\textwidth}{!}{%
\begin{tabular}{@{}llllrrrrr||lrrr@{}}
\toprule
\multicolumn{9}{c||}{\textbf{(a) 2$\times$2$\times$2 Factorial Isolation (Qwen2.5-3B, 19.6K)}} & \multicolumn{4}{c}{\textbf{(b) \sagethree{} v1.0.0 Causal Forensics}} \\
Mode & Cond. & Select & Refine & \texttt{s1} & \texttt{mk2} & \texttt{mv} & \texttt{vt} & \textbf{AVG} & Input ($T{=}8192$) & \texttt{pbm=1} & \texttt{pbm=0} & R-8K \\
\midrule
BF16 SDPA & --- & --- & Exact & 100 & 66.7 & 66.7 & 65.3 & \textbf{74.67\%} & Unit Gaussian (non-causal) & 0.145 & 0.146 & --- \\
1-limb (ctr) & Ctr & None & --- & 100 & 20.0 & 35.0 & 54.7 & 52.42\% & Unit Gaussian (causal) & \textcolor{tierwarn}{0.609} & \textcolor{tierwarn}{0.609} & --- \\
1-limb+FWHT & Ctr+H & None & --- & 100 & 46.7 & 71.7 & 48.0 & 66.58\% & Llama L12 real (raw) & \textcolor{tierwarn}{0.630} & \textcolor{tierwarn}{0.629} & \textcolor{tierwarn}{2.0\%} \\
\thrift{} 5\% (raw) & None & $\bar q\!\cdot\!\bar k$ & FP16$^\dagger$ & 80 & 0.0 & 48.3 & 32.0 & 40.08\% & Llama L12 real (ctr) & --- & \textcolor{tierwarn}{0.628} & \textcolor{tierwarn}{2.0\%} \\
\thrift{} 5\%+FWHT & Ctr+H & $\bar q\!\cdot\!\bar k$ & FP16$^\dagger$ & 100 & 73.3 & 61.7 & 40.0 & 68.75\% & Llama L12 real (ctr+FWHT) & --- & \textcolor{tierwarn}{0.634} & \textcolor{tierwarn}{2.0\%} \\
Ctrl A: Proxy+2L & Ctr+H & $\bar q\!\cdot\!\bar k$ & 2-limb & 100 & 80.0 & 63.3 & 65.3 & 77.17\% & \ours{} 1-limb (raw) & --- & 0.222 & 100\% \\
Ctrl B: Margin+FP16 & Ctr+H & Margin & FP16 & 100 & 66.7 & 76.7 & 62.7 & 76.50\% & \ours{} 1-limb (ctr+FWHT) & --- & 0.178 & 99.0\% \\
\rowcolor{tierfour!10}
\textbf{\oursfwht{}} ($\Delta{=}1$) & Ctr+H & Margin & 2-limb & \textbf{100} & \textbf{73.3} & \textbf{66.7} & \textbf{74.7} & \textbf{78.67\%} & \textbf{\oursfwht{}} ($\Delta{=}1$) & --- & \textbf{0.047} & \textbf{98.5\%} \\
\bottomrule
\end{tabular}}
\\[1pt]{\footnotesize $^\dagger$Official \thrift{} CUDA bulk pass uses power-of-two rounded attention weights.}
\end{table*}

\begin{table}[t]
\caption{\textbf{Conditioning ablation} (compiled SM120 kernels, LongBench $n{=}32$ $\Delta\text{lp}$, $\Delta{=}1$). Offline cal.: 6 disjoint prompts ($5.5$--$7.4$\,s).}
\label{tab:ablation}
\centering
\small
\setlength{\tabcolsep}{2.8pt}
\resizebox{\columnwidth}{!}{%
\begin{tabular}{@{}lrrr@{\hspace{8pt}}rrr@{}}
\toprule
 & \multicolumn{3}{c}{Nanbeige4.2-3B} & \multicolumn{3}{c}{Qwen2.5-7B} \\
\cmidrule(lr){2-4}\cmidrule(l){5-7}
Conditioning mode & $\Delta\text{lp}$ & ok & bad & $\Delta\text{lp}$ & ok & bad \\
\midrule
No centering + FWHT & $-0.015$ & 20 & 0 & $-7.975$ & 0 & 32 \\
Centering only (no rot.) & $-0.034$ & 20 & 1 & $-0.090$ & 21 & 4 \\
\rowcolor{tierfour!10}
\textbf{Fused FWHT (\oursfwht{})} & $\mathbf{-0.013}$ & \textbf{25} & \textbf{0} & $\mathbf{-0.018}$ & 21 & \textbf{0} \\
Online full-seq PCA & $-0.013$ & 18 & 0 & $-0.044$ & 24 & 2 \\
Causal prefix-512 PCA & $\mathbf{-0.007}$ & 21 & \textbf{0} & $\mathbf{-0.016}$ & 23 & 1 \\
Offline cal.\ $(R,s,\mu_k)$ & $\mathbf{-0.009}$ & 21 & \textbf{0} & $-0.040$ & \textbf{25} & 3 \\
Offline + band gate $\Delta_b{=}4$ & $\mathbf{-0.011}$ & 24 & \textbf{0} & $-0.042$ & 22 & 1 \\
\bottomrule
\end{tabular}}
\end{table}

\textbf{2$\times$2$\times$2 factorial isolation and ablations.}
On Qwen2.5-3B RULER 19.6K (\cref{tab:app_factorial}), isolating conditioning, selection, and refinement format confirms: (i)~conditioning alone (\texttt{z1+FWHT}, $66.6\%$) lifts unrotated 1-limb ($52.4\%$) yet still trails BF16 ($74.7\%$); (ii)~two-limb NVFP4 (\oursfwht{} $78.7\%$) matches FP16 backup under the same margin gate (\texttt{Margin+FP16} $76.5\%$); and (iii)~\thrift{}~5\% + FWHT ($68.8\%$, \texttt{vt} $40.0\%$) lags exact-softmax two-limb refinement ($77.2\%$) due to power-of-two bulk softmax (while \cref{tab:app_theory} isolates $3.9$--$4.2\times$ lower MSE for margin vs.\ block-mean under identical \oursfwht{} two-limb refinement on Llama-8B, Qwen-3B, and Phi-4).
In \cref{tab:ablation}, removing $\mu_k$ collapses Qwen2.5-7B ($-7.975$\,nats), while fused FWHT cuts centered-only loss $2.6$--$5.0\times$ with $0/32$ tail failures.
Appendix ablations (\cref{tab:app_cal}) confirm stability across $\Delta\!\in\!\{0.5..4\}$, zero-shot 1--4 prompt transfer, necessity of 2-limb $V$ and decoupled $m_r^{\mathrm{ref}}$, and $1.11$--$1.22\times$ steady-state speedup over \kwide{} under RTX~5090's 575\,W TDP wall (\cref{app:kernel_details}).

\section{Related Work}
\label{sec:related}

\textbf{Low-bit tensor-core attention.}
FlashAttention \citep{dao2022flashattention,dao2024flashattention2,shah2024flashattention3} established tiled online-softmax execution as the foundation for modern attention kernels.
SageAttention \citep{zhang2025sageattention,zhang2025sageattention2,zhang2025sageattention2pp} quantized $QK^{\top}$ to INT8 and $\widetilde{P}V$ to FP8 with per-head key smoothing, while \sagethree{} \citep{zhang2025sageattention3}, Attn-QAT \citep{zhang2026attnqat}, MpFA \citep{deng2026mpfa} (QK4V8), FP4 FlashAttention-4 \citep{hu2026fp4fa4}, and NVFP4 FlashInfer \citep{flashinfer2026nvfp4msa} extend attention to Blackwell FP4 tensor cores.
Unlike Attn-QAT---which relies on quantization-aware training to remove outlier handling and reports $760$\,TFLOPS E2E at non-causal 32K on RTX~5090 (5090 kernel unreleased)---\ours{} is strictly training-free and causal, reaching $947$--$967$\,TFLOPS E2E ($1{,}040$\,TFLOPS 1-limb+FWHT) on real 64K--128K LLM activations.
As shown in \cref{sec:exp}, unrefined 1-limb NVFP4 still degrades long-context LLM retrieval even with exact causal masking and FWHT conditioning (dropping $-4.0$ to $-15.1$ points on RULER), while official \sagethree{} additionally suffers a causal-mode bug (\cref{tab:app_sage3}).
VC-Attention \citep{li2026vcattention} and DraftAttention2 \citep{ding2026draftattention2} focus on 4-bit attention for video diffusion rather than long-context LLM retrieval.

\textbf{Selective mixed-precision and sparse attention.}
Query-aware sparse attention methods such as Quest \citep{tang2024quest} and SpargeAttn \citep{zhang2025spargeattn} drop low-importance key pages entirely, whereas DMA \citep{ding2026dma} partitions attention along diagonal tiles with mixed MXFP4/MXFP8 precision and \thrift{} \citep{sharratt2026thriftattention} runs a 4-bit bulk pass with selective FP16 refinement on a fixed top-$f$ fraction of blocks chosen via block-mean inner products $\bar q\cdot\bar k$.
Concurrently, TriAttention \citep{zhang2026triattention} combines block skipping with 4-bit/8-bit mixed precision across prefill and compressed KV decoding.
\ours{} differs in three respects: (i)~rather than diagonal heuristics, fixed top-$f$ block-mean proxies, or token dropping, we use an online per-stage logit-margin gate backed by \cref{thm:exact_retention}, adapting refinement per 16-row warp tile; (ii)~we eliminate separate BF16/FP16 or 8-bit fallback copies via shared-scale two-limb NVFP4, reducing KV storage $44\%$ below BF16 while keeping refinement on native FP4 tensor cores; and (iii)~our single-sweep SM120 kernel beats both \thrift{} and dense 8-bit \sagetwo{} while directly feeding unified two-limb FlashDecoding (\ourscache{}).

\textbf{Rotation and low-rank quantization compensation.}
Incoherent Walsh--Hadamard rotations (QuaRot~\citep{ashkboos2024quarot}; SpinQuant~\citep{liu2025spinquant}) suppress activation outliers, while ResQ \citep{saxena2024resq} and \deco{} \citep{he2026decoattention} allocate extra precision or FP4 compensation channels to high-variance principal directions.
Our analysis confirms that $89$--$91\%$ of 1-limb FP4 error sits in 16 principal directions, yet uniform channel compensation alone leaves $>0.2$\,logits of noise that still harms thin-margin retrieval (\cref{tab:longbench,tab:chain}).
\ours{} couples causal key centering with either a zero-HBM in-register 128-point FWHT or a margin-gated offline spectral top-band alongside two-limb refinement.

\textbf{Low-bit KV caches for decoding.}
KV-cache quantizers such as KIVI \citep{liu2024kivi}, QJL \citep{zandieh2025qjl}, TurboQuant \citep{zandieh2026turboquant}, BitDecoding-FP4 \citep{du2026bitdecodingfp4}, NVFP4-MLA \citep{ni2026nvfp4mla}, and ScaleSearchAttention \citep{gupta2026scalesearch} target memory-bandwidth-bound autoregressive decoding where single-token queries read stored keys and values, leaving compute-bound prefill unaccelerated or requiring post-prefill re-quantization.
\ours{} unifies both phases: prefill writes shared-scale two-limb NVFP4 records (\texttt{rec1} always read, \texttt{rec2} fetched on demand) that \ourscache{} directly consumes during batched decoding with zero re-packing and zero BF16 prefix KV---whereas FlashInfer 0.7.6's SM120 NVFP4 prefill (\texttt{nvfp4\_attention\_sm120}) and SM120 XQA NVFP4 decode use separate incompatible layouts and suffer 1-limb accuracy loss ($0.14$--$0.29$ rel.\ err)---boosting 32-layer serving by $2.36$--$2.87\times$ over FlashInfer XQA BF16 and $1.13$--$1.15\times$ over XQA FP8.

\section{Conclusion and Limitations}
\label{sec:conclusion}

\ours{} unifies fast long-context prefill and memory-efficient decoding in one shared-scale two-limb NVFP4 layout with zero FP16 backup.
Guided by our raw-error first-order covariance decomposition and Exact-Score Retention Theorem, \ours{} pairs key centering and orthogonal FWHT conditioning with online logit-margin gating and reuses the exact packed KV records in \ourscache{} for FlashDecoding---reaching $867$--$967$\,TFLOPS E2E on real 31K--128K activations ($4.01$--$4.44\times$ FlashAttention-2, $1.5$--$1.6\times$ \sagetwo{}, and $1.13$--$1.41\times$ faster with $2.2$--$2.9\times$ lower error than FlashInfer SM120 NVFP4), holding $943$--$951$\,TFLOPS across diffuse, QA, prose, and code workloads with zero degradation cliff, achieving BF16-level accuracy across full 13-task RULER and LongBench within paired bootstrap $95\%$ CIs, and lifting 32-layer serving by $2.36$--$2.87\times$ over production FlashInfer XQA BF16 ($1.13$--$1.15\times$ over XQA FP8) at 32K--64K and $2.10$--$2.15\times$ XQA BF16 ($1.05$--$1.06\times$ XQA FP8 peak) in bounded-tail incremental-flush Long-CoT serving at 8K--16K.

\textbf{Limitations.}
(1)~At $T \le 8\text{K}$, the 16-stage local window dominates, so \kwide{} stays faster and smaller ($204$ vs.\ $288$\,B/key); \ourscache{} also stores $12.5\%$ more KV bytes than FP8 ($288$ vs.\ $256$\,B/key) and $2\times$ more than 1-limb XQA NVFP4 ($144$\,B/key, which reaches higher max batch sizes at the cost of $9$--$13\times$ higher decode error and no unified prefill layout).
(2)~Dense NVFP4 hits RTX~5090's 575\,W TDP wall relative to INT8, and our kernels currently target $d{=}128$ on one consumer SM120 GPU ($d\!\in\!\{64,256\}$ and SM100 \texttt{tcgen05} remain future work).
(3)~Teacher-forced greedy exact match retains a measurable gap to BF16 ($25/32$ vs.\ $31/32$ on Nanbeige; $17$--$21/32$ vs.\ $25/32$ on Phi-4-mini), alongside the $0.044$--$0.047$ all-refined FP4 $P$/$V$ floor; while we validate bit-exact 64-token incremental flushing, $0.0000\%$ Long-CoT E4M3 clipping, and 8K--16K batched serving throughput on DeepSeek-R1-Distill-Llama-8B, large-scale end-to-end Long-CoT reasoning accuracy evaluation (e.g., full AIME/MATH-500) remains future work.

\textbf{Code availability.}
All SM120 CUDA kernels, \ourscache{} integration, and benchmark scripts are publicly available at \url{https://github.com/wrencanfly/ZenMargin}.

\section*{Use of Generative AI}
We used generative AI tools (large-language-model coding assistants) to assist with experimental design, writing and debugging experiment and analysis code, inspecting profiling logs, searching related literature, and drafting and polishing the manuscript. All AI-assisted outputs, including code, numerical results, citations, and text, were reviewed and verified by the authors, who take full responsibility for the content of the paper.

\bibliographystyle{abbrvnat}
\bibliography{refs}

\clearpage
\onecolumn
\appendix
\section{Proofs of Proposition 1, Lemma 1, and Theorem 1}
\label{app:proofs}

\begin{proof}[Proof of \cref{prop:firstorder}]
Let $p(s) = \mathrm{softmax}(s)$ with Jacobian $J_{jk} = \partial p_j / \partial s_k = p_j(\delta_{jk} - p_k)$.
For a raw score perturbation $e = \hat s - s$, first-order Taylor expansion gives
\[
  \delta p_j = \sum_{k=1}^N p_j(\delta_{jk} - p_k) e_k = p_j\!\left(e_j - \sum_{k=1}^N p_k e_k\right).
\]
Since $\sum_{j=1}^N \delta p_j = 0$ and $\sum_{j=1}^N p_j (v_j - o) = \sum_j p_j v_j - o = 0$, the centering scalar $\sum_k p_k e_k$ cancels identically when mapped to the output:
\[
  \delta o_{\mathrm{lin}} = \sum_{j=1}^N \delta p_j v_j = \sum_{j=1}^N \delta p_j (v_j - o) = \sum_{j=1}^N p_j e_j (v_j - o) - \left(\sum_{k=1}^N p_k e_k\right)\underbrace{\sum_{j=1}^N p_j(v_j - o)}_{=0} = A e,
\]
where column $j$ of $A \in \mathbb{R}^{d \times N}$ is $a_j = p_j(v_j - o)$.
Writing $e = \mu_e + (e - \mu_e)$ with $\mu_e = \mathbb{E}[e]$ and $\Sigma_e = \mathrm{Cov}(e)$ yields
\[
  \mathbb{E}\|\delta o_{\mathrm{lin}}\|^2
  = \mathbb{E}\!\left[\operatorname{tr}\!\big(A (e - \mu_e)(e - \mu_e)^\top A^\top\big)\right] + \|A\mu_e\|^2
  = \operatorname{tr}(A \Sigma_e A^\top) + \|A\mu_e\|^2.
\]
Expanding $\operatorname{tr}(A\Sigma_e A^\top) = \sum_{j,k} (\Sigma_e)_{jk}\langle a_j, a_k\rangle$ separates the diagonal independent-variance term $D_{\mathrm{raw}} = \sum_{j=1}^N p_j^2 \mathrm{Var}(e_j)\|v_j - o\|^2$ from the raw cross-key covariance term $\sum_{j\neq k} p_j p_k \mathrm{Cov}(e_j, e_k)\langle v_j - o, v_k - o\rangle$.
For any realized error vector $e$ and independent Rademacher signs $\xi_j \stackrel{\mathrm{i.i.d.}}{\sim} \mathrm{Unif}\{\pm 1\}$, $\mathbb{E}[\xi_j \xi_k] = \delta_{jk}$, hence $\mathbb{E}_\xi \|A(\xi \odot e)\|^2 = \sum_{j=1}^N p_j^2 e_j^2 \|v_j - o\|^2 = D_{\mathrm{raw}}(e)$.
Finally, when $\mu_e = 0$ and $\Sigma_e = \mathrm{diag}(\sigma_1^2, \dots, \sigma_N^2)$ with $\sigma_j = \sigma_c$ on $\mathcal{U}$ and $\sigma_j = \sigma_r$ on $\mathcal{R}$, the cross and bias terms vanish and the unrefined contribution satisfies $\sigma_c^2 \sum_{j\in\mathcal{U}} p_j^2 \|v_j - o\|^2 \le \sigma_c^2 V_{\max}^2 \sum_{j\in\mathcal{U}} p_j^2$.
\end{proof}

\begin{proof}[Proof of \cref{lem:online_superset}]
For query row $r$ and any stage traversal permutation $\pi : \{1,\dots,S\} \to \{1,\dots,S\}$, let $t_s = \pi^{-1}(s)$ be the step at which stage $s$ (containing keys $B_s$) is visited. First, if key $j \in B_s$ belongs to $\tilde{\mathcal{R}}_{\Delta,r}^{\mathrm{key\text{-}global}}$, then $\tilde m_{r,s} = \max_{j'\in B_s}\tilde s_{r,j'} \ge \tilde s_{r,j} \ge \tilde s_{r,\max} - \Delta$, so every key in $B_s$ belongs to $\tilde{\mathcal{R}}_{\Delta,r}^{\mathrm{stage\text{-}global}}$. Second, since $\{\pi(1),\dots,\pi(t_s)\} \subseteq \{1,\dots,S\}$, the running cheap-only maximum satisfies
\[
  \tilde m_r^{\mathrm{ref}}(t_s) = \max_{1 \le u \le t_s} \tilde m_{r,\pi(u)} \le \max_{1 \le u \le S} \tilde m_{r,u} = \tilde s_{r,\max}.
\]
Thus any stage $s \subseteq \tilde{\mathcal{R}}_{\Delta,r}^{\mathrm{stage\text{-}global}}$ obeys $\tilde m_{r,s} \ge \tilde s_{r,\max} - \Delta \ge \tilde m_r^{\mathrm{ref}}(t_s) - \Delta$, placing $B_s \subseteq \tilde{\mathcal{R}}_{\Delta,r}^{\mathrm{online}}$. Finally, taking the warp disjunction over all 16 query rows $r'$ in a warp tile yields $\tilde{\mathcal{R}}_{\Delta,r}^{\mathrm{key\text{-}global}} \subseteq \tilde{\mathcal{R}}_{\Delta,r}^{\mathrm{stage\text{-}global}} \subseteq \tilde{\mathcal{R}}_{\Delta,r}^{\mathrm{online}} \subseteq \bigcup_{r'=0}^{15}\tilde{\mathcal{R}}_{\Delta,r'}^{\mathrm{online}} = \tilde{\mathcal{R}}_\Delta^{\mathrm{warp\text{-}online}}$.
\end{proof}

\begin{proof}[Proof of \cref{thm:exact_retention}]
Fix row $r$ and drop the subscript $r$ for readability. Let $Z = \sum_{k=1}^N e^{s_k}$ so that $p_k = e^{s_k}/Z$, $p_{\max} = e^{s_{\max}}/Z$, and $e^{s_k - s_j} = p_k / p_j$.

\emph{Part (a).}
By \cref{lem:online_superset}, if token $j$ in stage $s(j)$ is omitted by $\tilde{\mathcal{R}}_\Delta^{\mathrm{warp\text{-}online}}$, then $j \notin \tilde{\mathcal{R}}_\Delta^{\mathrm{online}}$, hence
\[
  s_j + \eta_j = \tilde s_j \le \tilde m_{s(j)} < \tilde m^{\mathrm{ref}}(t_{s(j)}) - \Delta = s_{\max} + \varepsilon_+(t_{s(j)}) - \Delta,
\]
where $\varepsilon_+(t) = \max_{k \in \mathrm{prefix}(t)}[\,\eta_k - (s_{\max} - s_k)\,] \le \max_{1 \le k \le N}[\,\eta_k - (s_{\max} - s_k)\,] = \varepsilon_+$. Rearranging gives $\varepsilon_+(t_{s(j)}) - \eta_j > \Delta - (s_{\max} - s_j)$. Taking the contrapositive: for any $j \in \mathcal{R}_\gamma^{\mathrm{exact}}$ (where $s_{\max} - s_j \le \gamma$), $\varepsilon_+(t_{s(j)}) - \eta_j \le \Delta - \gamma$ (or the stronger condition $\varepsilon_+ - \eta_j \le \Delta - \gamma$) guarantees $j \in \tilde{\mathcal{R}}_\Delta^{\mathrm{warp\text{-}online}}$. For the exact argmax $j^*$ ($s_{j^*} = s_{\max}$, $\gamma = 0$), omission requires $\tilde m^{\mathrm{ref}}(t_{s(j^*)}) - \tilde s_{j^*} > \Delta$, i.e., there exists some key $k$ in a stage visited at or before $t_{s(j^*)}$ with $\tilde s_k - \tilde s_{j^*} > \Delta \iff \eta_k - \eta_{j^*} > \Delta + (s_{j^*} - s_k)$.

\emph{Master indicator inequality for Parts (b)--(c).}
By \cref{lem:online_superset}, $j \notin \tilde{\mathcal{R}}_\Delta^{\mathrm{warp\text{-}online}}$ implies $j \notin \tilde{\mathcal{R}}_\Delta^{\mathrm{key\text{-}global}}$, i.e., $\tilde s_j < \tilde s_{\max} - \Delta$. Since $\Delta > 0$, the maximizer $k^* = \arg\max_k \tilde s_k$ cannot equal $j$, so $\max_{k \ne j}(\tilde s_k - \tilde s_j) > \Delta$. Because $\mathbf{1}\{x > \Delta\} \le e^{\lambda(x - \Delta)}$ for all $x \in \mathbb{R}$ and every $\lambda > 0$,
\begin{equation}
  \mathbf{1}\!\left\{j \notin \tilde{\mathcal{R}}_\Delta^{\mathrm{warp\text{-}online}}\right\}
  \;\le\; \mathbf{1}\!\left\{\max_{k \ne j}(\tilde s_k - \tilde s_j) > \Delta\right\}
  \;\le\; e^{-\lambda\Delta}\max_{k \ne j} e^{\lambda(\tilde s_k - \tilde s_j)}
  \;\le\; e^{-\lambda\Delta}\sum_{k \ne j}\!\left(\frac{p_k}{p_j}\right)^{\!\lambda} e^{\lambda(\eta_k - \eta_j)}.
\label{eq:master_ind}
\end{equation}
Notice that for zero-mean $\sigma^2$-sub-Gaussian errors $\eta_1,\dots,\eta_N$, independence gives $\mathbb{E}[e^{\lambda(\eta_k - \eta_j)}] = \mathbb{E}[e^{\lambda\eta_k}]\mathbb{E}[e^{-\lambda\eta_j}] \le e^{\lambda^2\sigma^2}$ ($\tau^2 = 2\sigma^2$), while under \emph{arbitrary} cross-key dependence Cauchy--Schwarz yields $\mathbb{E}[e^{\lambda\eta_k}e^{-\lambda\eta_j}] \le (\mathbb{E}[e^{2\lambda\eta_k}]\mathbb{E}[e^{-2\lambda\eta_j}])^{1/2} \le e^{2\lambda^2\sigma^2}$ ($\tau^2 = 4\sigma^2$).

\emph{Part (b).}
For any $j$ with $\gamma_j = s_{\max} - s_j = \ln(p_{\max}/p_j)$ and any $\lambda \ge 1$, writing $p_k/p_j = (p_k/p_{\max}) e^{\gamma_j}$ in \cref{eq:master_ind} and taking expectations yields
\[
  \Pr\!\big[j \notin \tilde{\mathcal{R}}_\Delta^{\mathrm{warp\text{-}online}}\big]
  \;\le\; \left(\sum_{k \ne j}\left(\frac{p_k}{p_{\max}}\right)^{\!\lambda}\right) \exp\!\Big(-\lambda(\Delta - \gamma_j) + \tfrac{1}{2}\lambda^2\tau^2\Big).
\]
Because $p_k / p_{\max} \le 1$ and $\lambda \ge 1$, we have $\sum_{k \ne j}(p_k/p_{\max})^\lambda \le N_{\mathrm{eff}}(\lambda) := \sum_{k=1}^N (p_k/p_{\max})^\lambda \le \sum_{k=1}^N p_k/p_{\max} = 1/p_{\max}$ (and $\sum_{k\ne j}(p_k/p_{\max})^\lambda \le (1 - p_j)/p_{\max}$), establishing \cref{eq:exact_miss_prob} independently of $N$. Minimizing $-\lambda(\Delta - \gamma_j) + \frac{1}{2}\lambda^2\tau^2$ over $\lambda \ge 1$ achieves $\lambda^* = (\Delta - \gamma_j)/\tau^2 \ge 1$ whenever $\Delta - \gamma_j \ge \tau^2$, yielding exponent $-(\Delta - \gamma_j)^2/(2\tau^2)$. For the exact top-1 key $j^*$ ($\gamma_{j^*}=0$) with margin $g = s_{(1)} - s_{(2)} \ge 0$, every $k \ne j^*$ obeys $p_k/p_{\max} \le e^{-g}$, so
\[
  \sum_{k \ne j^*}\left(\frac{p_k}{p_{\max}}\right)^{\!\lambda}
  \;\le\; e^{-(\lambda - 1)g}\sum_{k \ne j^*}\frac{p_k}{p_{\max}}
  \;=\; \frac{1 - p_{\max}}{p_{\max}} e^{-(\lambda - 1)g},
\]
giving $\Pr[j^* \notin \tilde{\mathcal{R}}_\Delta^{\mathrm{warp\text{-}online}}] \le \frac{1-p_{\max}}{p_{\max}}\exp(g - \lambda(\Delta+g) + \frac{1}{2}\lambda^2\tau^2)$, which attains $\exp(g - (\Delta+g)^2/(2\tau^2))$ at $\lambda^* = (\Delta+g)/\tau^2 \ge 1$. Finally, for all $j \in \mathcal{R}_\gamma^{\mathrm{exact}}$ we have $\gamma_j \le \gamma$, so multiplying \cref{eq:exact_miss_prob} by $p_j$ and summing over $j \in \mathcal{R}_\gamma^{\mathrm{exact}}$ gives \cref{eq:exact_missed_mass}.

\emph{Part (c).}
Setting $\lambda = 1$ in \cref{eq:master_ind}, multiplying both sides by $p_j^2$, summing over $j=1,\dots,N$, and taking expectations gives
\[
  \mathbb{E}\!\left[\sum_{j \notin \tilde{\mathcal{R}}_\Delta^{\mathrm{warp\text{-}online}}} p_j^2\right]
  \;\le\; e^{-\Delta}\sum_{j=1}^N \sum_{k \ne j} p_j p_k\,\mathbb{E}\!\left[e^{\eta_k - \eta_j}\right]
  \;\le\; e^{-\Delta + \frac{1}{2}\tau^2}\sum_{j \ne k} p_j p_k
  \;=\; e^{-\Delta + \frac{1}{2}\tau^2}\!\left(1 - \sum_{j=1}^N p_j^2\right)
  \;\le\; e^{-\Delta + \frac{1}{2}\tau^2}.
\]
In the exact noiseless case ($\eta \equiv 0$, $\mathcal{R}_\Delta = \{j : s_j \ge s_{\max} - \Delta\}$), every $j \notin \mathcal{R}_\Delta$ satisfies $p_j = p_{\max} e^{s_j - s_{\max}} < p_{\max} e^{-\Delta} \le e^{-\Delta}$, so $\sum_{j \notin \mathcal{R}_\Delta} p_j^2 \le (\max_{j \notin \mathcal{R}_\Delta} p_j)\sum_{j \notin \mathcal{R}_\Delta} p_j < e^{-\Delta}(1 - \sum_{j \in \mathcal{R}_\Delta} p_j) \le e^{-\Delta}$.
\end{proof}

\section{Algorithm and SM120 Kernel Details}
\label{app:kernel_details}

\begin{figure}[ht]
  \centering
  \safefig{\textwidth}{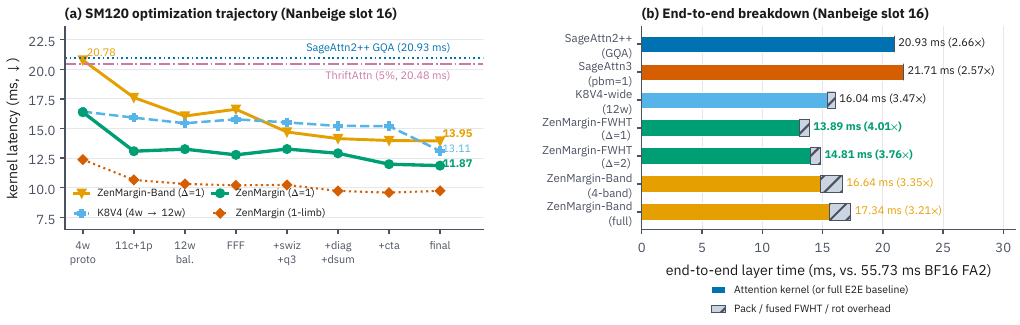}
  \caption{\textbf{SM120 systems analysis on Nanbeige4.2-3B slot~16} ($T{=}31{,}296, 12.04$\,TFLOPs, RTX~5090). \emph{(a)}~Optimization trajectory from the initial 4-warp prototype to the final 12-warp First-Flagger-Fetches kernel: $\Delta{=}1$ drops $16.38 \to 11.87$\,ms ($-27.5\%$) and $\Delta{=}1$+band drops $20.78 \to 13.95$\,ms ($-32.9\%$), surpassing official \sagetwo{}, \thrift{}, and the upgraded 12-warp \kwide{} ($13.11$\,ms). \emph{(b)}~End-to-end breakdown showing fused FWHT + CUDA packing adds $0.78$\,ms ($5.6\%$ of $13.89$\,ms E2E).}
  \label{fig:kernelopt}
\end{figure}

\Cref{alg:zenmargin} summarizes one warp's single-sweep execution in \ours{}.
Below we detail the SM120 Blackwell PTX, TMA, shared-memory, and FlashDecoding implementation behind \cref{sec:kernel}, followed by compiled register/spill profiles (\cref{tab:resources}), optimization trajectory (\cref{fig:kernelopt}), and steady-state thermal power measurements.

\textbf{Blackwell PTX arithmetic and 12-warp self-producing ring buffer.}
SM120 executes NVFP4 matrix products via \texttt{mma.sync.aligned.m16n8k64} (\texttt{mxf4nvf4}, \texttt{scale\_vec::4X}, $8{,}192$ multiply-adds/warp, $\approx 2\times$ INT8 \texttt{m16n8k32.s8}).
SASS profiling (\texttt{cuobjdump}) of an initial 4-warp ($64$\,queries/CTA) prototype ($20.78$\,ms on 31K Nanbeige slot~16) exposed five bottlenecks: TMA pressure at 4 warps/CTA, a two-pass refinement sweep re-reading \texttt{rec1}, 2-way $V$ \texttt{ldmatrix.x4} bank conflicts, cluster-copy ABI calls from \texttt{shared::cluster} TMA, and SM120's 168-register/thread ceiling at 3 warps/sub-core (spills $+7.5$\,ms).
Widening to $W_C{=}12$ consumer warps ($384$\,threads, $192$\,rows/CTA; 3 warps/sub-core) cuts TMA traffic $3\times$ without adding a 13th producer warp (which drops registers to $128$/thread): warp~11 lane~0 self-produces \texttt{rec1} loads into a $5$--$10$ slot \texttt{mbarrier} ring buffer via CTA-scoped \texttt{cp.async.bulk.shared::cta.global} (\texttt{CALL=0}, zero in-loop \texttt{\_\_syncthreads()}), using reverse-causal GQA scheduling for L2 reuse.
Each slot reserves space for both \texttt{rec1} and \texttt{rec2}: the first flagging warp wins an \texttt{atomicCAS} on the slot flag to launch \texttt{rec2} TMA while flagging warps overlap $Q^{(2)}K^{(1)\top}$ on resident \texttt{rec1}.
In-place score updates (IEEE $-\infty + \text{finite} = -\infty$ preserves diagonal masks without recomputing $Q^{(1)}K^{(1)\top}$), conflict-free $V$-swizzling (\texttt{row >> 2}, $-0.39$\,ms), and hiding initial $V$ loads behind $\ell_r$ cut kernel latency $27.5$--$32.9\%$ (\cref{fig:kernelopt}a).
In \oursband{}, the top 16 PCA coordinates occupy NVFP4 block~7, adding one $16\times 8\times 64$ MMA when within $\Delta_{\mathrm{band}}$ and masking block~7's scale byte during full refinement to prevent double-counting.

\textbf{In-register 128-pt FWHT packer and FlashDecoding kernels.}
Centering, NVFP4 packing, and 128-point randomized FWHT run via native FP8/FP4 PTX conversions where 8 lanes hold one 128-element vector via a 4-stage register + 3-stage 8-lane warp-shuffle butterfly (safe for arbitrary sequence lengths $T$, zero shared memory, zero extra HBM passes, $1.8\times 10^{-4}$ relative diff vs.\ FP32).
For unified decoding, our split-KV FlashDecoding kernel launches 4 warps/CTA (\texttt{\_\_launch\_bounds\_\_(128,4)}, zero in-loop \texttt{\_\_syncthreads()}), issuing 128-bit \texttt{uint4} coalesced loads of \texttt{rec1} with single E2M1$\to$FP16x2 key decode reused across all $G$ grouped queries and CTA-shared BOS+tail anchoring of $m^{\mathrm{ref}}$.
Our isomorphic SM120 FP8 E4M3 FlashDecoding baseline (\texttt{Fp8Cache}, $256$\,B/key) mirrors this exact 4-warp split-KV architecture using 128-bit loads and native \texttt{cvt.rn.f16x2.e4m3x2} conversions.

\begin{algorithm}[ht]
\caption{\ours{} single-sweep attention (one warp, 16 query rows). [TC]: SM120 NVFP4 \texttt{mma.m16n8k64}.}
\label{alg:zenmargin}
\small
\begin{algorithmic}[1]
\STATE \textbf{Input:} packed two-limb $Q^{(1)}, Q^{(2)}$, stage records \texttt{rec1} ($K^{(1)},V^{(1)}$), \texttt{rec2} ($K^{(2)},V^{(2)}$); thresholds $\Delta, \Delta_{\mathrm{band}}$; \texttt{recent} $=16$.
\STATE Load $Q^{(1)}_w$ (and top-band $Q^x_w$ if \oursband{}) into registers; init $m_r = m_r^{\mathrm{ref}} = -\infty$, $\ell_r = 0$, $O_r = 0$.
\FOR{$i = 0$ \textbf{to} $n_{\mathrm{vis}}-1$}
  \STATE $s \leftarrow (i < n_{\mathrm{rec}}) \;?\; (S - i) : (i - n_{\mathrm{rec}})$ \COMMENT{recent first, then BOS+}
  \STATE Wait on \texttt{full1[$b$]} for resident \texttt{rec1[$s$]}
  \STATE $S \leftarrow Q^{(1)}_w K^{(1)\top}_s$ [TC] \COMMENT{2 MMAs / $16\times 8$ tile}
  \IF{\oursband{} \textbf{and} ($\Delta_{\mathrm{band}}{=}\infty$ \textbf{or} $\mathrm{near}(S, m^{\mathrm{ref}}, \Delta_{\mathrm{band}})$)}
    \STATE $S \leftarrow S + \mathrm{band\_mma}(Q^x_w, K^x_s)$ [TC] \COMMENT{1 extra MMA, block 7}
  \ENDIF
  \IF{$s == S_w$}
    \STATE Apply causal mask ($S_{r,j} \leftarrow -\infty$ for $j > \mathrm{pos}(r)$)
  \ENDIF
  \STATE Update cheap max $m_r^{\mathrm{ref}} \leftarrow \max(m_r^{\mathrm{ref}}, \max_j S_{r,j})$
  \STATE $\mathrm{flagged} \leftarrow \bigvee_{r=0}^{15}\big(\max_j S_{r,j} \ge m_r^{\mathrm{ref}} - \Delta \log_2 e\big)$
  \IF{$\mathrm{flagged}$}
    \STATE First warp on slot $b$ issues TMA for \texttt{rec2[$s$]} via \texttt{atomicCAS}
    \STATE Load $Q^{(2)}_w$; mask blk-7 scale if band ran
    \STATE $S \leftarrow S + Q^{(2)}_w K^{(1)\top}_s$ [TC] \COMMENT{overlaps \texttt{rec2} TMA}
    \STATE Wait on \texttt{full2[$b$]}; mask blk-7 scale on $Q^{(1)}_w$ if band ran
    \STATE $S \leftarrow S + Q^{(1)}_w K^{(2)\top}_s$ [TC] \COMMENT{zero $Q^{(1)}K^{(1)\top}$ recompute}
  \ENDIF
  \STATE Update softmax $(m_r, \ell_r)$, rescale $O_r$, quantize $\hat P = \mathcal{Q}_{\mathrm{FP4}}(\exp_2(S - m_r))$
  \STATE $O_r \leftarrow O_r + \hat P V^{(1)}_s$ [TC] \COMMENT{hides $V^{(1)}$ load behind $\ell_r$ MMA}
  \IF{$\mathrm{flagged}$}
    \STATE $O_r \leftarrow O_r + \hat P V^{(2)}_s$ [TC]
  \ENDIF
  \STATE Signal \texttt{empty1[$b$]}
\ENDFOR
\STATE \textbf{return} $O_r \cdot (g_V / \ell_r)$
\end{algorithmic}
\end{algorithm}

\begin{table}[ht]
\caption{\textbf{Compiled SM120 resource profile} (\texttt{cuobjdump}, CUDA 13.2, RTX~5090, $W_C{=}12$, 384 threads/CTA). All specializations remain within the 168-register/thread sub-core ceiling with zero hot-loop spills and zero cluster-ABI calls.}
\label{tab:resources}
\centering
\small
\setlength{\tabcolsep}{5.0pt}
\begin{tabular}{@{}lrrrrrl@{}}
\toprule
Kernel specialization & Reg & Stack & STL & LDL & Call & Notes \\
\midrule
\ours{} (1-limb unref.) & 166 & 0 & 0 & 0 & 0 & Zero spill \\
\ours{} (1-limb + top-band) & 168 & 8 & 1 & 1 & 0 & Prologue only \\
\oursfwht{} ($\Delta{=}1, 2$) & 168 & 16 & 2 & 3 & 0 & Prologue only \\
\oursband{} ($\Delta{=}1, 2$) & 168 & 8 & 1 & 1 & 0 & Prologue only \\
\kwide{} (12-warp 8-bit baseline) & 168 & 0 & 0 & 0 & 0 & Zero spill \\
\bottomrule
\end{tabular}
\end{table}

\textbf{Steady-state 575\,W TDP measurements.}
After full thermal warmup on Nanbeige slot~16 (median of 10 runs), steady-state latency / throughput / clock / board power are:
\kwide{}: $15.68$\,ms / $768$\,TOPS / $2{,}790$\,MHz / $337$\,W;
1-limb FP4: $11.27$\,ms / $1{,}068$\,TOPS / $2{,}137$\,MHz / $506$\,W ($1.39\times$ \kwide{});
$\Delta{=}1$: $12.90$\,ms / $933$\,TOPS / $2{,}137$\,MHz / $506$\,W ($1.22\times$);
$\Delta{=}2$: $14.09$\,ms / $854$\,TOPS / $2{,}362$\,MHz / $559$\,W ($1.11\times$);
$\Delta{=}1, \Delta_b{=}2$: $13.97$\,ms / $862$\,TOPS / $2{,}452$\,MHz / $578$\,W ($1.12\times$);
$\Delta{=}1, \Delta_b{=}4$: $14.76$\,ms / $815$\,TOPS / $2{,}452$\,MHz / $578$\,W ($1.06\times$);
$\Delta{=}1$+full band: $15.43$\,ms / $780$\,TOPS / $2{,}460$\,MHz / $577$\,W ($1.02\times$).

\section{Extended Theoretical and Diagnostic Measurements}
\label{app:theory}

\begin{table}[ht]
\caption{\textbf{Raw-error covariance decomposition and matched-variance controls (\cref{prop:firstorder}) across four LLM families} (\oursfwht{} \texttt{m\_had}, plus unrotated \texttt{ctr-only} baseline): Llama-3.1-8B ($23.4\text{K}$), Nanbeige4.2-3B ($31.4\text{K}$), Qwen2.5-3B ($31.2\text{K}$), and Phi-4-mini ($17.2\text{K}$). \texttt{meas}: true relative MSE $\|\hat o - o\|^2/\|o\|^2$; \texttt{lin}: exact raw-error linearization $\|Ae\|^2/\|o\|^2$; $D_{\mathrm{raw}}$: exact raw independent diagonal $\sum_j p_j^2 e_j^2 \|v_j - o\|^2/\|o\|^2$ (matched to 3 digits by 8-draw Rademacher sign-flips); \texttt{shuf}: 8-draw within-set permutation of raw $e_j$ inside $\mathcal{U}, \mathcal{R}$ separately; \texttt{bias}: $\|A\mu_{\text{set}}\|^2/\|o\|^2$; \texttt{p1un}: unrefined linear attention mass $\sum_{j\in\mathcal{U}} p_j$. Unrotated 1-limb FP4 shows $3.53$--$4.10\times$ amplification over \texttt{shuf} (cut to $1.12$--$2.18\times$ by FWHT); under margin gating ($\Delta\!\in\![0.5,4]$), raw cross-key covariance + set bias still amplify \texttt{lin} by $1.79$--$3.07\times$ over $D_{\mathrm{raw}}$ ($1.77$--$2.36\times$ over \texttt{shuf}).}
\label{tab:app_theory}
\centering
\small
\setlength{\tabcolsep}{2.2pt}
\resizebox{\textwidth}{!}{%
\begin{tabular}{@{}lrrrrrrrr@{\hspace{6pt}}rrrrrrrr@{}}
\toprule
 & \multicolumn{8}{c}{Llama-3.1-8B (23.4K)} & \multicolumn{8}{c}{Nanbeige4.2-3B (31.4K)} \\
\cmidrule(lr){2-9}\cmidrule(l){10-17}
Rule (\oursfwht{}) & budg & meas & lin & $D_{\mathrm{raw}}$ & shuf & bias & $\text{lin}/D_{\mathrm{raw}}$ & p1un & budg & meas & lin & $D_{\mathrm{raw}}$ & shuf & bias & $\text{lin}/D_{\mathrm{raw}}$ & p1un \\
\midrule
Unref.\ (ctr-only, no FWHT) & 0.000 & 2.45e-1 & 2.20e-1 & 2.30e-1 & 6.22e-2 & 0 & $0.95\times$ ($3.53^\dagger$) & 1.000 & 0.000 & 1.46e-1 & 1.15e-1 & 1.17e-1 & 3.25e-2 & 0 & $0.99\times$ ($3.54^\dagger$) & 1.000 \\
Unref.\ (+FWHT) & 0.000 & 1.02e-1 & 7.80e-2 & 9.63e-2 & 5.05e-2 & 0 & $0.81\times$ ($1.54^\dagger$) & 1.000 & 0.000 & 7.24e-2 & 6.32e-2 & 6.26e-2 & 2.90e-2 & 0 & $1.01\times$ ($2.18^\dagger$) & 1.000 \\
Margin ($\Delta{=}0.5$) & 0.010 & 1.84e-2 & 2.04e-2 & 1.14e-2 & 9.28e-3 & 2.23e-3 & \textbf{1.79$\times$} & 0.227 & 0.006 & 8.13e-3 & 9.19e-3 & 4.78e-3 & 4.26e-3 & 1.24e-3 & \textbf{1.92$\times$} & 0.186 \\
Margin ($\Delta{=}1$) & 0.011 & 1.55e-2 & 1.72e-2 & 9.07e-3 & 7.71e-3 & 1.99e-3 & \textbf{1.90$\times$} & 0.205 & 0.008 & 6.40e-3 & 7.20e-3 & 3.44e-3 & 3.22e-3 & 1.09e-3 & \textbf{2.10$\times$} & 0.169 \\
Margin ($\Delta{=}2$) & 0.015 & 1.03e-2 & 1.15e-2 & 5.41e-3 & 5.05e-3 & 1.54e-3 & \textbf{2.13$\times$} & 0.169 & 0.016 & 3.97e-3 & 4.47e-3 & 1.77e-3 & 1.90e-3 & 8.40e-4 & \textbf{2.53$\times$} & 0.139 \\
Margin ($\Delta{=}4$) & 0.048 & 3.49e-3 & 3.87e-3 & 1.36e-3 & 1.64e-3 & 7.34e-4 & \textbf{2.85$\times$} & 0.107 & 0.076 & 1.66e-3 & 1.74e-3 & 6.00e-4 & 7.92e-4 & 4.54e-4 & \textbf{2.90$\times$} & 0.081 \\
\thrift{} (5\%) & 0.051 & 6.11e-2 & 4.77e-2 & 4.86e-2 & 2.58e-2 & 1.34e-3 & $0.98\times$ & 0.520 & 0.051 & 4.05e-3 & 3.87e-3 & 2.50e-3 & 1.78e-3 & 4.37e-4 & $1.55\times$ & 0.131 \\
\thrift{} (25\%) & 0.251 & 2.40e-2 & 2.13e-2 & 2.13e-2 & 1.16e-2 & 6.89e-4 & $1.00\times$ & 0.239 & 0.251 & 1.28e-3 & 1.33e-3 & 1.00e-3 & 6.55e-4 & 1.37e-4 & $1.33\times$ & 0.061 \\
All refined & 1.000 & 5.99e-4 & 5.99e-4 & 6.58e-4 & 4.38e-4 & 0 & $0.91\times$ & 0.000 & 1.000 & 4.86e-4 & 4.85e-4 & 4.63e-4 & 2.48e-4 & 0 & $1.05\times$ & 0.000 \\
\midrule
 & \multicolumn{8}{c}{Qwen2.5-3B (31.2K)} & \multicolumn{8}{c}{Phi-4-mini (17.2K)} \\
\cmidrule(lr){2-9}\cmidrule(l){10-17}
Unref.\ (ctr-only, no FWHT) & 0.000 & 1.80e-1 & 1.76e-1 & 1.87e-1 & 4.30e-2 & 0 & $0.94\times$ ($4.10^\dagger$) & 1.000 & 0.000 & 1.90e-1 & 1.55e-1 & 1.60e-1 & 4.13e-2 & 0 & $0.97\times$ ($3.76^\dagger$) & 1.000 \\
Unref.\ (+FWHT) & 0.000 & 5.68e-2 & 5.08e-2 & 6.79e-2 & 4.54e-2 & 0 & $0.75\times$ ($1.12^\dagger$) & 1.000 & 0.000 & 9.62e-2 & 7.54e-2 & 9.03e-2 & 4.46e-2 & 0 & $0.84\times$ ($1.69^\dagger$) & 1.000 \\
Margin ($\Delta{=}0.5$) & 0.010 & 8.35e-3 & 9.11e-3 & 4.42e-3 & 4.55e-3 & 1.47e-3 & \textbf{2.06$\times$} & 0.305 & 0.009 & 1.34e-2 & 1.48e-2 & 8.25e-3 & 6.91e-3 & 1.89e-3 & \textbf{1.79$\times$} & 0.224 \\
Margin ($\Delta{=}1$) & 0.016 & 6.50e-3 & 7.04e-3 & 3.08e-3 & 3.54e-3 & 1.32e-3 & \textbf{2.29$\times$} & 0.265 & 0.014 & 1.09e-2 & 1.19e-2 & 6.11e-3 & 5.43e-3 & 1.69e-3 & \textbf{1.94$\times$} & 0.193 \\
Margin ($\Delta{=}2$) & 0.041 & 3.89e-3 & 4.20e-3 & 1.47e-3 & 2.04e-3 & 1.00e-3 & \textbf{2.86$\times$} & 0.201 & 0.031 & 6.13e-3 & 6.78e-3 & 2.77e-3 & 2.94e-3 & 1.27e-3 & \textbf{2.45$\times$} & 0.146 \\
Margin ($\Delta{=}4$) & 0.128 & 1.61e-3 & 1.65e-3 & 5.99e-4 & 9.33e-4 & 5.01e-4 & \textbf{2.76$\times$} & 0.112 & 0.086 & 2.68e-3 & 2.85e-3 & 9.30e-4 & 1.27e-3 & 7.31e-4 & \textbf{3.07$\times$} & 0.088 \\
\thrift{} (5\%) & 0.052 & 2.51e-2 & 2.33e-2 & 2.25e-2 & 1.70e-2 & 1.38e-3 & $1.03\times$ & 0.361 & 0.053 & 4.53e-2 & 3.80e-2 & 3.78e-2 & 1.86e-2 & 1.23e-3 & $1.00\times$ & 0.402 \\
\thrift{} (25\%) & 0.251 & 1.14e-2 & 1.11e-2 & 1.05e-2 & 7.79e-3 & 6.64e-4 & $1.05\times$ & 0.170 & 0.253 & 1.96e-2 & 1.65e-2 & 1.63e-2 & 7.68e-3 & 4.95e-4 & $1.01\times$ & 0.152 \\
All refined & 1.000 & 4.56e-4 & 4.55e-4 & 5.26e-4 & 3.93e-4 & 0 & $0.86\times$ & 0.000 & 1.000 & 5.75e-4 & 5.77e-4 & 6.25e-4 & 3.84e-4 & 0 & $0.92\times$ & 0.000 \\
\bottomrule
\end{tabular}}
\\[1pt]{\footnotesize $^\dagger$Ratio $\text{lin}/\text{shuf}$ on unrefined rows, showing FWHT slashes raw cross-key correlation from $3.53$--$4.10\times$ down to $1.12$--$2.18\times$.}
\end{table}

\begin{table}[ht]
\caption{\textbf{Online vs.\ offline-oracle gate forensics and worst-case score error tails} (\oursfwht{} across all four LLMs). \texttt{ot16}: exact offline oracle (16-row tile); \texttt{cglob}: global cheap-score gate; \texttt{onl}: actual recent-16 + BOS-first online gate; \texttt{ltr}: naive left-to-right online scan. \texttt{FN\_cg}: false negatives vs.\ \texttt{cglob} ($0.0000$ everywhere, verifying \cref{lem:online_superset}). \texttt{FN\_or}, \texttt{mis\_p2}, \texttt{mis\_p1}: stage FN rate and missed $\sum p_j^2, \sum p_j$ vs.\ exact oracle \texttt{ot16}. Bottom: per-row cheap-score error tails across all keys vs.\ high-weight keys ($p_j \ge 0.01$).}
\label{tab:app_online_forensics}
\centering
\small
\setlength{\tabcolsep}{3.0pt}
\resizebox{\textwidth}{!}{%
\begin{tabular}{@{}lcrrrrrrrcrrr@{}}
\toprule
Model ($T$) & $\Delta$ & \texttt{ot16} & \texttt{cglob} & \texttt{onl} & \texttt{ltr} & \texttt{FN\_cg} & \texttt{FN\_or} & \texttt{mis\_p2} & \texttt{mis\_p1} & \texttt{p1un} & \texttt{mse\_or} & \texttt{mse\_onl} \\
\midrule
Llama-3.1-8B (23.4K) & 1.0 & 1.0\% & 1.1\% & \textbf{1.9\%} & 1.6\% & \textbf{0.0000} & 2.16\% & 1e-5 & 0.0005 & 0.138 & 1.70e-2 & \textbf{3.39e-3} \\
Llama-3.1-8B (23.4K) & 2.0 & 1.4\% & 1.5\% & \textbf{2.8\%} & 2.6\% & \textbf{0.0000} & 3.31\% & 1e-5 & 0.0008 & 0.124 & 1.17e-2 & \textbf{2.95e-3} \\
Nanbeige-3B (31.4K) & 1.0 & 0.7\% & 0.7\% & \textbf{1.5\%} & 2.1\% & \textbf{0.0000} & 4.45\% & 5e-5 & 0.0015 & 0.140 & 6.81e-3 & \textbf{2.27e-3} \\
Nanbeige-3B (31.4K) & 2.0 & 1.4\% & 1.5\% & \textbf{2.7\%} & 4.5\% & \textbf{0.0000} & 6.91\% & 2e-5 & 0.0021 & 0.120 & 4.15e-3 & \textbf{1.86e-3} \\
Qwen2.5-3B (31.2K) & 1.0 & 1.3\% & 1.4\% & \textbf{2.4\%} & 5.4\% & \textbf{0.0000} & 5.75\% & 1.6e-4 & 0.0055 & 0.219 & 7.37e-3 & \textbf{2.49e-3} \\
Qwen2.5-3B (31.2K) & 2.0 & 3.6\% & 3.7\% & \textbf{5.0\%} & 10.3\% & \textbf{0.0000} & 6.16\% & 2e-5 & 0.0039 & 0.169 & 4.49e-3 & \textbf{1.95e-3} \\
Phi-4-mini (17.2K) & 1.0 & 1.3\% & 1.4\% & \textbf{3.0\%} & 5.1\% & \textbf{0.0000} & 2.93\% & 2e-5 & 0.0019 & 0.136 & 1.22e-2 & \textbf{2.39e-3} \\
Phi-4-mini (17.2K) & 2.0 & 2.8\% & 3.0\% & \textbf{5.4\%} & 7.4\% & \textbf{0.0000} & 3.11\% & 1e-5 & 0.0018 & 0.107 & 7.03e-3 & \textbf{1.97e-3} \\
\bottomrule
\end{tabular}}
\\[1pt]{\footnotesize \textbf{Score error tails (\oursfwht{}) and \cref{thm:exact_retention}(c) bounds:} RMS mean (med) / $\max_{\text{all}}$ med (p99) / $\max_{p\ge 0.01}$ med (p90) in logits --- Llama-8B: $0.316$ ($0.309$) / $1.34$ ($2.38$) / $0.50$ ($0.92$); Nanbeige: $0.330$ ($0.318$) / $1.40$ ($3.08$) / $0.58$ ($1.02$); Qwen-3B: $0.401$ ($0.385$) / $1.66$ ($3.42$) / $0.69$ ($1.21$); Phi-4: $0.362$ ($0.344$) / $1.45$ ($3.17$) / $0.57$ ($1.06$). At $\Delta{=}1$, naive uniform $e^{-(\Delta-2\varepsilon_{\max})} = 5.35$--$10.20$ (vacuous), whereas $N$-free \cref{thm:exact_retention}(c) gives $0.407$--$0.432$ (indep.) and $0.449$--$0.507$ (arbitrary correlation; $0.020$--$0.025$ at $\Delta{=}4$).}
\end{table}

\paragraph{Retrieval-head margin diagnostics.}\label{app:retrieval}
Across 4 prompts at $8\text{K}$, $16\text{K}$, and $31\text{K}$ tokens, Nanbeige4.2-3B activates $55.5$--$64.8$ retrieval heads per prompt (exact gold mass $0.364$--$0.378$), of which $80$--$83\%$ have gold-vs-distractor logit margin $< 1$\,logit (medians $-0.41, -0.89, -0.64$\,logits).
Plain NVFP4 injects $1.040$--$1.219$\,logits RMS noise on these rows, dropping retained gold mass to $90\%$ at 31K; two-limb NVFP4 cuts retrieval-row noise to $0.089$--$0.104$\,logits and preserves $100\%$ of gold mass.
Similarly, Qwen2.5-3B activates $18.2$--$24.2$ retrieval heads ($75$--$88\%$ with margin $<1$\,logit; medians $-0.20, -0.63, +0.16$\,logits); plain NVFP4 injects $1.090$--$1.234$\,logits noise (dropping gold mass to $85\%$ at 31K), whereas two-limb NVFP4 cuts noise to $0.097$--$0.110$\,logits ($100\%$ retained).

\paragraph{Spectral band error distribution.}\label{app:sigma}
In the per-head key-PCA eigenbasis (ascending variance, 1-limb NVFP4), coordinate bands $[0,64)$, $[64,96)$, $[96,112)$, and $[112,128)$ carry respectively $(1.3\%, 2.9\%, 4.8\%, 90.9\%)$ of total score-error energy on Nanbeige4.2-3B at 31K tokens and $(2.1\%, 3.2\%, 5.3\%, 89.4\%)$ on Qwen2.5-3B at 31K tokens (consistent at 8K: $91.3\%$ and $90.0\%$ in the top 16 coordinates).

\section{Real 4K--128K Scaling, 32-Layer 64K Distribution, TTFT, and Unified Serving}
\label{app:multislot}

\textbf{All-32-layer real 64K distribution (Llama-3.1-8B, \texttt{gov\_report} prose).}
Across all 32 layers at $T{=}65{,}536$:
$\Delta{=}1$ warp-stage flag rate averages $1.28\%$ (min $0.70\%$, median $1.25\%$, p90 $1.61\%$, max $2.50\%$ on L0) with E2E latency mean $37.12$\,ms (min $36.62$, median $37.05$, p90 $37.61$, max $38.29$\,ms on L0; mean rel.\ err $0.0631$).
$\Delta{=}2$ warp-stage flag rate averages $2.16\%$ (min $0.97\%$, median $1.92\%$, p90 $2.63\%$, max $9.29\%$ on L0) with E2E latency mean $37.59$\,ms (max $41.52$\,ms on L0; mean rel.\ err $0.0569$).
By comparison, \kwide{} averages $46.51$\,ms ($47.07$\,ms on L0), \sagetwo{} averages $60.67$\,ms ($62.10$\,ms on L0), \thrift{}~5\% averages $66.44$\,ms ($66.26$\,ms on L0), and BF16 FA2 averages $162.91$\,ms.

\begin{table}[ht]
\caption{\textbf{Unified single-run end-to-end latency (ms) and speedup over BF16 FA2 on \emph{real} Llama-3.1-8B Layer~16 activations} ($H{=}8, G{=}4, H_q{=}32$; real \texttt{gov\_report} prose, RTX~5090). Includes official \thrift{} (5\%), native-GQA \sagetwo{}, isomorphic \kwide{}, \oursfwht{} ($\Delta{=}1, 2$ with exact warp-stage flag rate \texttt{rf}), and \oursband{} ($\Delta_b{=}4$).}
\label{tab:scaling}
\centering
\small
\setlength{\tabcolsep}{3.2pt}
\resizebox{\textwidth}{!}{%
\begin{tabular}{@{}rrrrrrrr@{}}
\toprule
$T$ & FA2 & \thrift{} (5\%) & \sagetwo{} & \kwide{} & \textbf{\oursfwht{}} ($\Delta{=}1$, \texttt{rf}) & \textbf{\oursfwht{}} ($\Delta{=}2$, \texttt{rf}) & \oursband{} ($\Delta_b{=}4$) \\
\midrule
4K & 0.79 & 0.40 ($1.99\times$) & 0.34 ($2.30\times$) & \textbf{0.23 ($3.44\times$)} & 0.26 ($3.04\times$, 18.3\%) & 0.27 ($2.91\times$, 26.7\%) & 0.33 ($2.38\times$) \\
8K & 2.81 & 1.34 ($2.10\times$) & 1.11 ($2.53\times$) & \textbf{0.80 ($3.53\times$)} & 0.80 ($3.52\times$, 9.5\%) & 0.84 ($3.33\times$, 14.2\%) & 1.05 ($2.67\times$) \\
16K & 10.54 & 4.59 ($2.30\times$) & 4.10 ($2.57\times$) & 3.05 ($3.46\times$) & \textbf{2.75 ($3.84\times$, 4.9\%)} & 2.86 ($3.68\times$, 7.3\%) & 3.36 ($3.14\times$) \\
32K & 41.20 & 17.33 ($2.38\times$) & 15.78 ($2.61\times$) & 11.77 ($3.50\times$) & \textbf{9.95 ($4.14\times$, 2.5\%)} & 10.07 ($4.09\times$, 3.8\%) & 10.98 ($3.75\times$) \\
64K & 163.20 & 66.55 ($2.45\times$) & 60.21 ($2.71\times$) & 46.48 ($3.51\times$) & \textbf{37.00 ($4.41\times$, 1.2\%)} & 37.35 ($4.37\times$, 1.9\%) & 40.42 ($4.04\times$) \\
128K & 646.35 & --- & 235.48 ($2.74\times$) & 184.41 ($3.50\times$) & \textbf{145.52 ($4.44\times$, 0.7\%)} & 146.63 ($4.41\times$, 1.2\%) & 154.39 ($4.19\times$) \\
\bottomrule
\end{tabular}}
\end{table}

\begin{table}[ht]
\caption{\textbf{Fused CUDA packing + in-register 128-pt FWHT overhead scaling on \emph{real} Llama-3.1-8B Layer~16 activations} ($4\text{K}\to 128\text{K}$, same unified run as \cref{tab:scaling}). The 7-stage register/warp-shuffle FWHT adds $\le 0.01$\,ms over plain packing and shrinks monotonically from $19.33\%$ at 8K to $1.78\%$ of total E2E layer time at 128K ($967.1$\,TFLOPS E2E). Individual components and E2E are independent medians (minor $\pm 0.05$--$0.3$\,ms host/launch variance).}
\label{tab:pack_scale}
\centering
\small
\setlength{\tabcolsep}{4.8pt}
\begin{tabular}{@{}rrrrrrrr@{}}
\toprule
$T$ & Pack plain & Pack+FWHT & FWHT $\Delta$ & Attend ($\Delta{=}1$) & \textbf{E2E \oursfwht{}} & Pack share & \textbf{E2E TFLOPS} \\
\midrule
4,096 & 0.032\,ms & 0.045\,ms & $+0.012$\,ms & 0.184\,ms & 0.260\,ms & 17.17\% & 528.0 \\
8,192 & 0.153\,ms & 0.154\,ms & $+0.001$\,ms & 0.645\,ms & 0.800\,ms & 19.33\% & 690.5 \\
16,384 & 0.323\,ms & 0.325\,ms & $+0.002$\,ms & 2.430\,ms & 2.750\,ms & 11.83\% & 801.0 \\
32,768 & 0.659\,ms & 0.659\,ms & $+0.000$\,ms & 9.129\,ms & 9.950\,ms & 6.63\% & 884.4 \\
65,536 & 1.358\,ms & 1.364\,ms & $+0.005$\,ms & 35.591\,ms & 37.000\,ms & \textbf{3.69\%} & \textbf{950.9} \\
131,072 & 2.615\,ms & 2.592\,ms & $-0.024$\,ms & 143.200\,ms & 145.520\,ms & \textbf{1.78\%} & \textbf{967.1} \\
\bottomrule
\end{tabular}
\end{table}

\begin{table}[ht]
\caption{\textbf{Full-model end-to-end prefill TTFT (ms and speedup over BF16)} across all Transformer layers + final-token \texttt{lm\_head} across four open LLMs on RTX~5090.}
\label{tab:app_ttft}
\centering
\small
\setlength{\tabcolsep}{4.5pt}
\begin{tabular}{@{}llrrrr@{}}
\toprule
Model & Method & 8K & 16K & 32K & 64K \\
\midrule
\multirow{5}{*}{Llama-3.1-8B (32L)}
& BF16 SDPA & 704 (1.00$\times$) & 1552 (1.00$\times$) & 3712 (1.00$\times$) & 9954 (1.00$\times$) \\
& \thrift{} (5\%) & 667 (1.06$\times$) & 1375 (1.13$\times$) & 2965 (1.25$\times$) & 6900 (1.44$\times$) \\
& \sagetwo{} & 656 (1.08$\times$) & 1358 (1.15$\times$) & 2931 (1.28$\times$) & 6775 (1.48$\times$) \\
& \kwide{} & 648 (1.09$\times$) & 1321 (1.17$\times$) & 2783 (1.33$\times$) & 6310 (1.58$\times$) \\
& \textbf{\oursfwht{}} ($\Delta{=}1$) & \textbf{649 (1.09$\times$)} & \textbf{1311 (1.18$\times$)} & \textbf{2725 (1.36$\times$)} & \textbf{5967 (1.67$\times$)} \\
\midrule
\multirow{4}{*}{Qwen2.5-7B (28L)}
& BF16 SDPA & 630 (1.00$\times$) & 1376 (1.00$\times$) & 3254 (1.00$\times$) & 8471 (1.00$\times$) \\
& \sagetwo{} & 590 (1.07$\times$) & 1220 (1.13$\times$) & 2624 (1.24$\times$) & 5962 (1.42$\times$) \\
& \kwide{} & 585 (1.08$\times$) & 1197 (1.15$\times$) & 2532 (1.29$\times$) & 5621 (1.51$\times$) \\
& \textbf{\oursfwht{}} ($\Delta{=}1$) & \textbf{585 (1.08$\times$)} & \textbf{1189 (1.16$\times$)} & \textbf{2491 (1.31$\times$)} & \textbf{5400 (1.57$\times$)} \\
\midrule
\multirow{4}{*}{Qwen2.5-3B (36L, $G{=}8$)}
& BF16 SDPA & 318 (1.00$\times$) & 728 (1.00$\times$) & 1809 (1.00$\times$) & 5054 (1.00$\times$) \\
& \thrift{} (5\%) & 289 (1.10$\times$) & 621 (1.17$\times$) & 1376 (1.31$\times$) & 3293 (1.53$\times$) \\
& \kwide{} & 280 (1.14$\times$) & 589 (1.24$\times$) & 1270 (1.42$\times$) & 2925 (1.73$\times$) \\
& \textbf{\oursfwht{}} ($\Delta{=}1$) & \textbf{280 (1.14$\times$)} & \textbf{583 (1.25$\times$)} & \textbf{1240 (1.46$\times$)} & \textbf{2775 (1.82$\times$)} \\
\midrule
\multirow{4}{*}{Phi-4-mini (32L, $G{=}3$)}
& BF16 SDPA & 368 (1.00$\times$) & 875 (1.00$\times$) & 2236 (1.00$\times$) & 6430 (1.00$\times$) \\
& \sagetwo{} & 328 (1.12$\times$) & 718 (1.22$\times$) & 1611 (1.39$\times$) & 3929 (1.64$\times$) \\
& \kwide{} & 319 (1.15$\times$) & 691 (1.27$\times$) & 1514 (1.48$\times$) & 3572 (1.80$\times$) \\
& \textbf{\oursfwht{}} ($\Delta{=}1$) & \textbf{320 (1.15$\times$)} & \textbf{687 (1.27$\times$)} & \textbf{1473 (1.52$\times$)} & \textbf{3370 (1.91$\times$)} \\
\bottomrule
\end{tabular}
\end{table}

\begin{table}[ht]
\caption{\textbf{Six-workload diffuse, QA, prose, and code stress profile on \emph{real} Llama-3.1-8B activations} (RTX~5090, SM120; 32K = all 32 layers, 64K = 8 evenly spaced layers L0,4,\dots,28). E2E latency and throughput for \oursfwht{} ($\Delta{=}1$) remain flat across peaked (\texttt{niah\_mk2}), diffuse word aggregation (\texttt{cwe}, \texttt{fwe}), multi-doc QA (\texttt{qa\_2}), dense prose (\texttt{gov\_report}), and repetitive long-context code (\texttt{lcc}), ruling out any refinement cliff.}
\label{tab:app_diffuse_profile}
\centering
\small
\setlength{\tabcolsep}{3.0pt}
\resizebox{\textwidth}{!}{%
\begin{tabular}{@{}clrrrrrrrrr@{}}
\toprule
$T$ & Workload (type) & FA2 (ms) & \sagetwo{} & \textbf{ZM $\Delta{=}1$ E2E (max)} & \textbf{TF/s} & $\times$FA2 & $\times$SA2++ & Warp flag (max) & CTA fetch (max) & Rel.\ err \\
\midrule
\multirow{6}{*}{\shortstack{32K\\(32L)}}
 & \texttt{niah\_mk2} (peaked) & 40.44 & 14.56 & \textbf{9.33} (9.76) & \textbf{943} & $4.33\times$ & $1.56\times$ & 3.45\% (5.43\%) & 8.08\% (13.98\%) & 0.0635 \\
 & \texttt{cwe} (diffuse agg.) & 40.57 & 14.97 & \textbf{9.36} (9.70) & \textbf{940} & $4.34\times$ & $1.60\times$ & 2.95\% (4.75\%) & 6.69\% (13.91\%) & 0.0608 \\
 & \texttt{fwe} (Zipf freq.) & 40.57 & 15.23 & \textbf{9.47} (9.92) & \textbf{929} & $4.28\times$ & $1.61\times$ & 2.89\% (5.28\%) & 6.80\% (13.61\%) & 0.0541 \\
 & \texttt{qa\_2} (multi-doc QA) & 40.73 & 15.36 & \textbf{9.36} (9.62) & \textbf{939} & $4.35\times$ & $1.64\times$ & 1.79\% (3.42\%) & 4.10\% (7.05\%) & 0.0578 \\
 & \texttt{gov\_report} (prose) & 40.86 & 15.46 & \textbf{9.57} (9.82) & \textbf{919} & $4.27\times$ & $1.61\times$ & 2.52\% (3.81\%) & 5.69\% (10.41\%) & 0.0609 \\
 & \texttt{lcc\_code} (long code) & 40.88 & 15.54 & \textbf{9.57} (9.80) & \textbf{919} & $4.27\times$ & $1.62\times$ & 2.54\% (3.69\%) & 5.31\% (8.48\%) & 0.0580 \\
\midrule
\multirow{6}{*}{\shortstack{64K\\(8L)}}
 & \texttt{niah\_mk2} (peaked) & 161.63 & 60.51 & \textbf{37.32} (37.90) & \textbf{943} & $4.33\times$ & $1.62\times$ & 1.74\% (2.37\%) & 4.28\% (7.33\%) & 0.0643 \\
 & \texttt{cwe} (diffuse agg.) & 162.23 & 60.68 & \textbf{37.08} (37.75) & \textbf{949} & $4.38\times$ & $1.64\times$ & 1.39\% (1.89\%) & 3.27\% (6.27\%) & 0.0635 \\
 & \texttt{fwe} (Zipf freq.) & 162.57 & 60.66 & \textbf{37.21} (38.22) & \textbf{945} & $4.37\times$ & $1.63\times$ & 1.45\% (2.95\%) & 3.72\% (9.16\%) & 0.0572 \\
 & \texttt{qa\_2} (multi-doc QA) & 162.68 & 60.74 & \textbf{36.99} (38.62) & \textbf{951} & $4.40\times$ & $1.64\times$ & 1.06\% (2.71\%) & 2.76\% (6.72\%) & 0.0613 \\
 & \texttt{gov\_report} (prose) & 162.84 & 60.82 & \textbf{37.11} (38.33) & \textbf{948} & $4.39\times$ & $1.64\times$ & 1.33\% (2.44\%) & 3.02\% (5.16\%) & 0.0638 \\
 & \texttt{lcc\_code} (long code) & 163.27 & 60.78 & \textbf{37.09} (38.38) & \textbf{949} & $4.40\times$ & $1.64\times$ & 1.33\% (2.37\%) & 2.84\% (5.19\%) & 0.0598 \\
\bottomrule
\end{tabular}}
\end{table}

\begin{table}[ht]
\caption{\textbf{Real 32-layer Llama-3.1-8B single-GPU (32\,GB) serving vs.\ production FlashInfer 0.7.6 SM120 XQA} (left, identical zero-HF-overhead decode loop, real 32-layer \texttt{gov\_report} KV, \ourscache{} flag rate $1.04\%$ at 32K and $0.54\%$ at 64K) and \textbf{unified prefill+decode accuracy} (right). \ourscache{} beats production XQA FP8 at every $B \ge 2$--$3$ ($1.15\times$ at 32K $B{=}7$, $1.13\times$ at 64K $B{=}3$) and reaches $94$--$95\%$ of 1-limb decode-only XQA NVFP4 throughput at matched $B$ while cutting decode error $9$--$13\times$.}
\label{tab:app_serving}
\centering
\small
\setlength{\tabcolsep}{2.7pt}
\resizebox{\textwidth}{!}{%
\begin{tabular}{@{}cc|rrrrr|r||l|ccc|c@{}}
\toprule
\multicolumn{8}{c||}{\textbf{(a) 32-Layer Batched Serving: Production FlashInfer XQA vs.\ \ourscache{} (tok/s)}} & \multicolumn{5}{c}{\textbf{(b) Unified Prefill+Decode Accuracy}} \\
$T$ & $B$ & XQA BF16 & XQA FP8 & \textbf{\ourscache{} ($\Delta{=}1$)} & XQA NVFP4$^\dagger$ & ZM Peak & $\times$XQA-FP8 & Mode & R-16K & R-32K & R-67.6K & LB-10 \\
\midrule
\multirow{8}{*}{32K}
& 1 & 72.8 & 81.2 & 76.5 & 84.5 & 18.5\,GB & $0.94\times$ & BF16 Dynamic (32\,GB) & 100.0 & 100.0 & \textcolor{tierwarn}{OOM} & 49.14 \\
& 2 & 120.3 & 142.0 & 141.3 & 152.6 & 20.9\,GB & $1.00\times$ & BF16 exact (CPU offl.) & --- & --- & 100.0 & --- \\
& 3 & 156.6 & 195.0 & \textbf{204.0} & 216.5 & 23.3\,GB & $1.05\times$ & Unified 1-limb (ctr) & 93.94 & 93.44 & 88.65 & 47.28 \\
& 4 & \textcolor{tierwarn}{OOM} & 238.9 & \textbf{262.0} & 273.2 & 25.7\,GB & $1.10\times$ & Unified 1-limb+FWHT & --- & --- & 95.42 & --- \\
& 5 & \textcolor{tierwarn}{OOM} & 265.6 & \textbf{294.6} & 309.5 & 28.2\,GB & $1.11\times$ & \textbf{\ourscache{}} ($\Delta{=}1$) & \textbf{100.0} & \textbf{100.0} & \textbf{100.0} & \textbf{49.74} \\
& 6 & \textcolor{tierwarn}{OOM} & 292.9 & \textbf{327.6} & 346.3 & 30.6\,GB & $1.12\times$ & \textbf{\ourscache{}} ($\Delta{=}2$) & \textbf{100.0} & \textbf{99.69} & \textbf{100.0} & \textbf{49.74} \\
& 7 & \textcolor{tierwarn}{OOM} & 321.7 & \textbf{369.5} & 387.8 & 33.0\,GB & $\mathbf{1.15\times}$ & \multicolumn{5}{l}{\footnotesize 67.6K: BF16 offload $970$\,s vs.\ \ourscache{} $317$\,s ($28.0$\,GB).} \\
& 8--10 & \textcolor{tierwarn}{OOM} & \textcolor{tierwarn}{OOM} & \textcolor{tierwarn}{OOM} & 424.7--492.3 & $>$32\,GB & --- & \multicolumn{5}{l}{\footnotesize 67.6K \texttt{mk2}: 1-limb $66.7\%$, 1-limb+FWHT $83.3\%$, ZM $100\%$.} \\
\midrule
\multirow{4}{*}{64K}
& 1 & 61.6 & 72.9 & 72.8 & 79.4 & 20.9\,GB & $1.00\times$ & \multicolumn{5}{l}{\footnotesize $^\dagger$XQA NVFP4 ($144$\,B/key) is 1-limb decode-only ($9$--$13\times$ higher} \\
& 2 & \textcolor{tierwarn}{OOM} & 120.1 & \textbf{129.4} & 137.5 & 25.7\,GB & $1.08\times$ & \multicolumn{5}{l}{\footnotesize decode error, $15.1$\,ms/32L repack at 64K, separate prefill layout).} \\
& 3 & \textcolor{tierwarn}{OOM} & 156.5 & \textbf{176.9} & 187.3 & 30.6\,GB & $\mathbf{1.13\times}$ & \multicolumn{5}{l}{\footnotesize $\Delta{=}2$: $324.9$\,tok/s (32K $B{=}6$, \texttt{rf} $1.26\%$), $175.9$ (64K $B{=}3$, $0.85\%$).} \\
& 4--5 & \textcolor{tierwarn}{OOM} & \textcolor{tierwarn}{OOM} & \textcolor{tierwarn}{OOM} & 228.4--252.8 & $>$32\,GB & --- & \multicolumn{5}{l}{\footnotesize HF \texttt{DynamicCache} E2E: $298.5$/$161.0$\,tok/s ($3.55$/$4.38\times$ HF BF16).} \\
\bottomrule
\end{tabular}}
\end{table}

\begin{table}[ht]
\caption{\textbf{32-layer Long-CoT Test-Time Scaling serving on DeepSeek-R1-Distill-Llama-8B} (single 32\,GB RTX~5090, real 32-layer AIME Long-CoT KV, identical zero-HF decode harness). \ourscache{} ($\Delta{=}1$, incremental 64-token flush) includes amortized flush latency ($0.003$--$0.005$\,ms/step) and preallocated bounded $<64$-token BF16 tail ($8$\,MB/seq across 32L; real decode \texttt{rf} $5.8$--$5.9\%$ at 8K, $3.9\%$ at 16K; $\Delta{=}2$ achieves $1{,}367.3$/$678.3$\,tok/s at peak $B{=}26$/$13$ with \texttt{rf} $6.4\%$/$4.5\%$). Despite storing $12.5\%$ more KV bytes than FP8, \ourscache{} outperforms production FlashInfer XQA FP8 peak at both 8K ($1.05\times$; $1.14\times$ at $B{=}24$) and 16K ($1.06\times$; $1.10\times$ at $B{=}13$), beats XQA BF16 peak by $2.10$--$2.15\times$, and reaches $95$--$100\%$ of 1-limb XQA NVFP4 at matched $B{=}12$--$26$.}
\label{tab:app_tts_serving}
\centering
\small
\setlength{\tabcolsep}{3.0pt}
\resizebox{\textwidth}{!}{%
\begin{tabular}{@{}cc|rrrrr|rr||cc|rrrrr|rr@{}}
\toprule
\multicolumn{9}{c||}{\textbf{(a) 8K Long-CoT Batched Serving ($T{=}8{,}192$, tok/s)}} & \multicolumn{9}{c}{\textbf{(b) 16K Long-CoT Batched Serving ($T{=}16{,}384$, tok/s)}} \\
$B$ & Tail & XQA BF16 & XQA FP8 & \textbf{ZM ($\Delta{=}1$)} & XQA NVFP4 & Peak & $\times$FP8 & $\times$NV4 & $B$ & Tail & XQA BF16 & XQA FP8 & \textbf{ZM ($\Delta{=}1$)} & XQA NVFP4 & Peak & $\times$FP8 & $\times$NV4 \\
\midrule
1 & 8\,MB & 88.0 & 90.4 & 77.9 & 91.1 & 16.7\,GB & $0.86\times$ & $0.86\times$ & 1 & 8\,MB & 83.0 & 87.6 & 77.5 & 89.4 & 17.3\,GB & $0.88\times$ & $0.87\times$ \\
4 & 34\,MB & 287.2 & 316.0 & 293.5 & 328.3 & 18.5\,GB & $0.93\times$ & $0.89\times$ & 4 & 34\,MB & 243.7 & 287.6 & 282.6 & 309.3 & 20.9\,GB & $0.98\times$ & $0.91\times$ \\
8 & 67\,MB & 451.9 & 526.2 & 518.4 & 560.7 & 21.0\,GB & $0.99\times$ & $0.92\times$ & 6 & 50\,MB & 297.2 & 365.0 & \textbf{373.5} & 402.8 & 23.4\,GB & $1.02\times$ & $0.93\times$ \\
12 & 101\,MB & 592.2 & 725.7 & \textbf{743.7} & 781.0 & 23.4\,GB & $1.02\times$ & $0.95\times$ & 7 & 59\,MB & 326.0 & 410.1 & \textbf{430.9} & 458.5 & 24.6\,GB & $1.05\times$ & $0.94\times$ \\
14 & 117\,MB & 645.1 & 814.4 & \textbf{847.3} & 901.8 & 24.7\,GB & $1.04\times$ & $0.94\times$ & 8 & 67\,MB & \textcolor{tierwarn}{OOM} & 451.2 & \textbf{478.3} & 510.0 & 25.8\,GB & $1.06\times$ & $0.94\times$ \\
16 & 134\,MB & \textcolor{tierwarn}{OOM} & 891.2 & \textbf{944.6} & 1005.9 & 25.9\,GB & $1.06\times$ & $0.94\times$ & 10 & 84\,MB & \textcolor{tierwarn}{OOM} & 523.6 & \textbf{573.0} & 608.6 & 28.2\,GB & $1.09\times$ & $0.94\times$ \\
20 & 168\,MB & \textcolor{tierwarn}{OOM} & 1073.5 & \textbf{1160.9} & 1252.0 & 28.3\,GB & $1.08\times$ & $0.93\times$ & 12 & 101\,MB & \textcolor{tierwarn}{OOM} & 589.7 & \textbf{644.9} & 668.2 & 30.7\,GB & $1.09\times$ & $0.97\times$ \\
24 & 201\,MB & \textcolor{tierwarn}{OOM} & 1147.7 & \textbf{1303.5} & 1300.8 & 30.8\,GB & $\mathbf{1.14\times}$ & $\mathbf{1.00\times}$ & 13 & 109\,MB & \textcolor{tierwarn}{OOM} & 620.8 & \textbf{684.8} & 719.3 & 31.9\,GB & $\mathbf{1.10\times}$ & $0.95\times$ \\
26 & 218\,MB & \textcolor{tierwarn}{OOM} & 1234.2 & \textbf{1384.6} & 1405.0 & 32.0\,GB & $\mathbf{1.12\times}$ & $0.99\times$ & 14 & --- & \textcolor{tierwarn}{OOM} & 645.8 & \textcolor{tierwarn}{OOM} & 768.8 & $>$32\,GB & --- & --- \\
28/32 & --- & \textcolor{tierwarn}{OOM} & 1317.6/\textcolor{tierwarn}{OOM} & \textcolor{tierwarn}{OOM} & 1504/1748 & $>$32\,GB & --- & --- & 16--20 & --- & \textcolor{tierwarn}{OOM} & \textcolor{tierwarn}{OOM} & \textcolor{tierwarn}{OOM} & 851--1021 & $>$32\,GB & --- & --- \\
\bottomrule
\end{tabular}}
\end{table}

\paragraph{Single-layer FlashDecoding microbenchmarks vs.\ production FlashInfer XQA.}\label{app:decode_micro}
On real Llama-3.1-8B Layer~16 activations ($H{=}8, G{=}4, d{=}128$, RTX~5090), at $T{=}64\text{K}$ ($B{=}1, 2, 4$):
PyTorch BF16 SDPA takes $0.648 / 1.280 / 2.544$\,ms ($414$--$422$\,GB/s);
production FlashInfer 0.7.6 XQA BF16 ($512$\,B/key) takes $0.166 / 0.323 / 0.638$\,ms ($1{,}622$--$1{,}684$\,GB/s, rel.\ err $0.0035$);
production FlashInfer XQA FP8 ($256$\,B/key) takes $0.088 / 0.168 / 0.325$\,ms ($1{,}530$--$1{,}650$\,GB/s, rel.\ err $0.0495$ at 64K / $0.0652$ at 32K);
production FlashInfer XQA NVFP4 ($144$\,B/key, 1-limb) takes $0.041 / 0.098 / 0.188$\,ms ($1{,}542$--$1{,}859$\,GB/s, rel.\ err $0.1763$ at 64K / $0.2894$ at 32K, plus $0.471$\,ms/layer = $15.1$\,ms/32L KV repacking via \texttt{nvfp4\_kv\_quantize}).
Reading only $144.7$\,B/key (\texttt{rf} $0.49\%$), \ourscache{} ($\Delta{=}1$) takes $\mathbf{0.067 / 0.125 / 0.220}$\,ms ($1{,}131$--$1{,}379$\,GB/s effective, $\mathbf{1.31}$--$\mathbf{1.48\times}$ faster than production XQA FP8) with rel.\ err $\mathbf{0.0191}$ ($0.0169$ at $\Delta{=}2$; $0.0219$/$0.0209$ at 32K)---$2.6$--$3.0\times$ lower error than XQA FP8 and $9.2$--$13.2\times$ lower than XQA NVFP4 with $0.00$\,ms repacking.
(Our custom isomorphic SM120 \texttt{Fp8Cache} baseline runs at $1{,}011$--$1{,}222$\,GB/s, err $0.0268$, reaching $275.6$/$128.1$\,tok/s in HuggingFace E2E serving vs.\ \ourscache{}'s $298.5$/$161.0$\,tok/s.)

\begin{table}[ht]
\caption{\textbf{Multi-slot mean E2E and kernel timings on real 31K prompts} (RTX~5090). Left: Nanbeige4.2-3B ($T{=}31{,}360$, 8 evenly spaced slots). Right: Qwen2.5-7B-Instruct ($T{=}31{,}232$, 5 slots).}
\label{tab:app_multislot}
\centering
\small
\setlength{\tabcolsep}{5.0pt}
\begin{tabular}{@{}lrrrrr@{\hspace{12pt}}rrrrr@{}}
\toprule
 & \multicolumn{5}{c}{Nanbeige4.2-3B (8-slot mean)} & \multicolumn{5}{c}{Qwen2.5-7B (5-slot mean)} \\
\cmidrule(lr){2-6}\cmidrule(l){7-11}
Variant & ms & $\times$FA2 & $\times$K8 & Flag & Err & ms & $\times$FA2 & $\times$K8 & Flag & Err \\
\midrule
FlashAttn-2 (BF16) & 57.10 & $1.00\times$ & $0.27\times$ & --- & 0.0000 & 33.28 & $1.00\times$ & $0.26\times$ & --- & 0.0000 \\
\kwide{} (kern / E2E) & 15.32 / 15.91 & $3.59\times$ & $0.96\times$ & --- & 0.0889 & 8.70 / 8.92 & $3.73\times$ & $0.97\times$ & --- & 0.0977 \\
1-limb FP4 (kern / E2E) & 11.36 / 12.04 & $4.74\times$ & $1.27\times$ & 0.000 & 0.2532 & 6.41 / 6.92 & $4.81\times$ & $1.26\times$ & 0.000 & 0.2144 \\
$\Delta{=}1$ (kern / E2E) & 12.56 / 13.13 & $4.35\times$ & $1.17\times$ & 0.026 & 0.0608 & 7.17 / 7.57 & $4.40\times$ & $1.15\times$ & 0.029 & 0.0607 \\
$\Delta{=}2$ (kern / E2E) & 12.92 / 13.44 & $4.25\times$ & $1.14\times$ & 0.047 & 0.0548 & 7.52 / 7.78 & $4.28\times$ & $1.12\times$ & 0.058 & 0.0567 \\
$\Delta{=}1, \Delta_b{=}2$ (kern / E2E) & 13.32 / 14.00 & $4.08\times$ & $1.09\times$ & 0.026 & 0.0560 & 7.63 / 8.04 & $4.14\times$ & $1.08\times$ & 0.028 & 0.0504 \\
$\Delta{=}1, \Delta_b{=}4$ (kern / E2E) & 13.67 / 14.97 & $3.82\times$ & $1.02\times$ & 0.026 & 0.0496 & 7.84 / 8.22 & $4.05\times$ & $1.06\times$ & 0.028 & 0.0461 \\
$\Delta{=}1$+band (kern / E2E) & 14.99 / 15.29 & $3.74\times$ & $1.00\times$ & 0.026 & 0.0470 & 8.59 / 8.88 & $3.75\times$ & $0.98\times$ & 0.028 & 0.0431 \\
All refined (100\%, kern) & 25.83 & $2.21\times$ & $0.59\times$ & 1.000 & 0.0436 & 14.98 & $2.22\times$ & $0.58\times$ & 1.000 & 0.0406 \\
\bottomrule
\end{tabular}
\end{table}

\clearpage
\section{Teacher-Forced Diagnostics, Cross-Domain Transfer, and Extended Benchmarks}
\label{app:tf_eval}

\begin{table}[ht]
\caption{\textbf{Teacher-forced LongBench likelihood diagnostics across four LLM families} ($n{=}32$, 12K--30K tokens, mean $18.6\text{K}$): paired $\Delta\text{logp}$ (nats vs.\ BF16), greedy exact match \texttt{tf-ok} ($/32$), and tail failures \texttt{bad} ($<-0.3$\,nats). Compiled CUDA kernels: $\star$.}
\label{tab:longbench}
\centering
\small
\setlength{\tabcolsep}{3.4pt}
\resizebox{\textwidth}{!}{%
\begin{tabular}{@{}lccc@{\hspace{10pt}}ccc@{\hspace{10pt}}ccc@{\hspace{10pt}}ccc@{}}
\toprule
 & \multicolumn{3}{c}{Nanbeige4.2-3B (\texttt{-0.167})} & \multicolumn{3}{c}{Qwen2.5-3B (\texttt{-0.106})} & \multicolumn{3}{c}{Qwen2.5-7B (\texttt{-0.163})} & \multicolumn{3}{c}{Phi-4-mini (\texttt{-0.278})} \\
\cmidrule(lr){2-4}\cmidrule(lr){5-7}\cmidrule(lr){8-10}\cmidrule(l){11-13}
Method & $\Delta\text{lp}$ & ok & bad & $\Delta\text{lp}$ & ok & bad & $\Delta\text{lp}$ & ok & bad & $\Delta\text{lp}$ & ok & bad \\
\midrule
BF16 FlashAttn-2$^\star$ & 0.000 & 31 & 0 & 0.000 & 28 & 0 & 0.000 & 27 & 0 & 0.000 & 25 & 0 \\
\sagetwo{} (GQA)$^\star$ & $-0.011$ & 24 & 0 & --- & --- & --- & $-0.003$ & 23 & 0 & --- & --- & --- \\
\kwide{} (8-bit $K$, 4-bit $V$)$^\star$ & $-0.036$ & 15 & 1 & $-0.029$ & 20 & 0 & $-0.037$ & 19 & 0 & $-0.006$ & 21 & 0 \\
\midrule
Plain NVFP4 (emul.) & $-0.134$ & 17 & 5 & $-10.628$ & 0 & 32 & $-10.539$ & 0 & 32 & $-1.876$ & 5 & 17 \\
\sagethree{} FP4 (\texttt{pbm=1})$^\star$ & $-6.171$ & 1 & 31 & --- & --- & --- & $-9.804$ & 0 & 32 & --- & --- & --- \\
\ours{} (1-limb band, no ref.)$^\star$ & $-0.112$ & 14 & 5 & $-0.527$ & 12 & 14 & $-0.310$ & 9 & 11 & $-0.258$ & 8 & 10 \\
\thrift{} (5\%)$^\star$ & $-0.064$ & 15 & 1 & $-1.903$ & 7 & 25 & --- & --- & --- & --- & --- & --- \\
\thrift{} (5\% / 25\%, centered)$^\star$ & ---\,/\,$-0.014$ & ---\,/\,24 & ---\,/\,0 & $-0.366$/$-0.284$ & 11/16 & 11/6 & --- & --- & --- & --- & --- & --- \\
\deco{} (32ch, centered) & $-0.047$ & 15 & 1 & $-0.096$ & 18 & 3 & $-0.049$ & 21 & 3 & $-0.087$ & 12 & 2 \\
ResQ (16-dim, centered) & $-0.039$ & 19 & 1 & $-0.125$ & 18 & 7 & --- & --- & --- & --- & --- & --- \\
\midrule
\textbf{\oursfwht{}} ($\Delta{=}1$)$^\star$ & $\mathbf{-0.013}$ & \textbf{25} & \textbf{0} & $-0.042$ & 21 & 1 & $\mathbf{-0.018}$ & 21 & \textbf{0} & --- & --- & --- \\
\textbf{\oursband{}} (offline, $\Delta{=}1$)$^\star$ & $\mathbf{-0.011}$ & 24 & \textbf{0} & $-0.041$ & 22 & 1 & $-0.040$ & \textbf{25} & 3 & --- & --- & --- \\
\textbf{\oursband{}} (full band, $\Delta{=}1$)$^\star$ & $\mathbf{-0.002}$ & 23 & \textbf{0} & $\mathbf{-0.020}$ & \textbf{24} & 1 & $\mathbf{-0.032}$ & 22 & \textbf{0} & $\mathbf{-0.034}$ & 17 & 1 \\
\textbf{\oursband{}} (full band, $\Delta{=}2$)$^\star$ & $\mathbf{-0.010}$ & 21 & \textbf{0} & $\mathbf{-0.014}$ & \textbf{24} & \textbf{0} & $-0.038$ & \textbf{25} & 1 & $-0.038$ & \textbf{21} & 1 \\
\bottomrule
\end{tabular}}
\end{table}

\begin{table}[ht]
\caption{\textbf{32K and 64K 2-hop JSON chain lookup} (hop-2 $\Delta\text{logp}$ vs.\ BF16, \texttt{bad} $<-0.3$\,nats, refined share \texttt{rf}). Compiled SM120 kernels: $\star$.}
\label{tab:chain}
\centering
\small
\setlength{\tabcolsep}{3.0pt}
\resizebox{\textwidth}{!}{%
\begin{tabular}{@{}llrrrr@{\hspace{10pt}}llrrrr@{}}
\toprule
Model & Method & $\Delta\text{lp}$ & med & bad & rf & Model & Method & $\Delta\text{lp}$ & med & bad & rf \\
\midrule
\multirow{5}{*}{\shortstack[l]{Nanbeige\\32K ($n{=}16$)}}
 & \thrift{} 5\%/25\%$^\star$ & $-0.35$/$-0.11$ & --- & --- & .05/.25 &
\multirow{4}{*}{\shortstack[l]{Qwen-3B\\32K ($n{=}16$)}}
 & \thrift{} 5\%/25\%$^\star$ & $-1.20$/$-0.15$ & --- & 13/4 & .05/.25 \\
 & \sagetwo{}$^\star$ & $-0.041$ & --- & 0/16 & --- &
 & \deco{} (32ch) & $-0.127$ & $-0.076$ & 4/16 & 0.000 \\
 & \kwide{}$^\star$ & $-0.130$ & $-0.070$ & 3/16 & --- &
 & \kwide{}$^\star$ & $-0.081$ & --- & 2/16 & --- \\
 & \textbf{FWHT} ($\Delta{=}1$)$^\star$ & $\mathbf{+0.074}$ & $+0.001$ & 1/16 & 0.032 &
 & \textbf{Band} (off, $\Delta_b{=}2$)$^\star$ & $\mathbf{+0.002}$ & --- & \textbf{0/16} & 0.040 \\
 & \textbf{Band} (off, $\Delta{=}1$)$^\star$ & $\mathbf{+0.021}$ & $+0.000$ & \textbf{0/16} & 0.033 &
\multirow{4}{*}{\shortstack[l]{Qwen-7B\\32K ($n{=}16$)}}
 & \sagethree{}$^\star$ & $-3.372$ & --- & 16/16 & --- \\
\cmidrule(lr){1-6}
\multirow{3}{*}{\shortstack[l]{Nanbeige\\64K ($n{=}8$)}}
 & Plain NVFP4 & $-0.619$ & $-0.550$ & 5/8 & 0.000 &
 & \sagetwo{}$^\star$ & $+0.014$ & --- & 0/16 & --- \\
 & \deco{} (32ch) & $-0.213$ & $-0.091$ & 3/8 & 0.000 &
 & \kwide{}$^\star$ & $-0.059$ & $-0.005$ & 1/16 & --- \\
 & \textbf{Band} ($\Delta{=}1$) & $\mathbf{-0.005}$ & $+0.001$ & \textbf{0/8} & \textbf{0.019} &
 & \textbf{Band} (off, $\Delta_b{=}2$)$^\star$ & $\mathbf{+0.004}$ & --- & \textbf{0/16} & 0.037 \\
\bottomrule
\end{tabular}}
\end{table}

\begin{table}[ht]
\caption{\textbf{Extended ablations: (a)~$\Delta$ sensitivity and cross-domain offline transfer} (Llama-3.1-8B RULER 16K, $n{=}15$/task; offline spectral basis $(R, s, \mu_k)$ calibrated on 1--4 disjoint documents transfers zero-shot to synthetic needle retrieval); \textbf{(b)~necessity of two-limb $V$ refinement} (paired 1-limb vs.\ 2-limb $V$ at matched refined stage budget \texttt{rf}, $\Delta{=}2$) \textbf{and decoupled cheap-only $m_r^{\mathrm{ref}}$} (compiled SM120 kernel, $\Delta{=}1$, Qwen2.5-3B LongBench $n{=}32$ and Nanbeige slot~16 kernel latency).}
\label{tab:app_cal}
\centering
\small
\setlength{\tabcolsep}{6.0pt}
\begin{tabular}{@{}lrrrrrr@{}}
\toprule
\multicolumn{7}{@{}l}{\textbf{(a) $\Delta$ Sensitivity and Zero-Shot Cross-Domain Calibration Transfer (Llama-3.1-8B RULER 16K)}} \\
Setting & \texttt{niah\_mk2} & \texttt{niah\_mv} & \texttt{niah\_mq} & \texttt{vt} & \textbf{AVG} & Refined \texttt{rf} \\
\midrule
\oursfwht{} ($\Delta{=}0.5$) & \textbf{100.0} & \textbf{100.0} & \textbf{81.7} & \textbf{100.0} & \textbf{95.42\%} & 0.051 \\
\oursfwht{} ($\Delta{=}1$) & \textbf{100.0} & \textbf{100.0} & 80.0 & \textbf{100.0} & \textbf{95.00\%} & 0.069 \\
\oursfwht{} ($\Delta{=}2$) & \textbf{100.0} & \textbf{100.0} & 76.7 & \textbf{100.0} & 94.17\% & 0.118 \\
\oursfwht{} ($\Delta{=}4$) & \textbf{100.0} & \textbf{100.0} & 75.0 & \textbf{100.0} & 93.75\% & 0.270 \\
\midrule
Cal.\ \texttt{gov\_report} ($n{=}1$ summary) & \textbf{100.0} & \textbf{100.0} & 78.3 & \textbf{100.0} & \textbf{94.58\%} & 0.069 \\
Cal.\ \texttt{lcc} ($n{=}4$ Python code) & \textbf{100.0} & \textbf{100.0} & \textbf{81.7} & \textbf{100.0} & \textbf{95.42\%} & 0.068 \\
Cal.\ \texttt{qasper} ($n{=}4$ science QA) & \textbf{100.0} & \textbf{100.0} & 73.3 & \textbf{100.0} & 93.33\% & 0.068 \\
\bottomrule
\end{tabular}

\vspace{5pt}
\setlength{\tabcolsep}{4.0pt}
\resizebox{\textwidth}{!}{%
\begin{tabular}{@{}lrrrl@{}}
\toprule
\multicolumn{5}{@{}l}{\textbf{(b) Two-Limb $V$ Refinement and Decoupled $m_r^{\mathrm{ref}}$ Ablations (Paired Controls)}} \\
Benchmark \& Component Ablation & Ablated (1-limb $V$ / coupled $m_r$) & \textbf{Full (2-limb $V$ / decoupled $m_r^{\mathrm{ref}}$)} & \texttt{rf} & Effect \\
\midrule
Qwen2.5-3B LB ($n{=}32$, 1- vs.\ 2-limb $V$) & $\Delta\text{lp}\!=\!-0.039$, med $-0.018$, 23/32 & $\Delta\text{lp}\!=\!\mathbf{-0.019}$, med $\mathbf{0.000}$, \textbf{26/32} & 0.063 & $2.1\times$ lower loss, $+3$ ok \\
Phi-4-mini LB ($n{=}32$, 1- vs.\ 2-limb $V$) & $\Delta\text{lp}\!=\!-0.014$, med $-0.007$, 18/32 & $\Delta\text{lp}\!=\!\mathbf{0.000}$, med $\mathbf{+0.001}$, \textbf{20/32} & 0.060 & Loss eliminated, $+2$ ok \\
Phi-4-mini 32K chain ($n{=}16$, 1- vs.\ 2-limb $V$) & hop-2 $-0.134$, med $-0.136$, bad 4/16 & \textbf{hop-2} $\mathbf{-0.024}$, med $\mathbf{0.000}$, bad \textbf{3/16} & 0.037 & $5.6\times$ lower hop-2 loss \\
Nanbeige 32K chain ($n{=}16$, 1- vs.\ 2-limb $V$) & hop-2 $+0.042$, bad 3/16 & \textbf{hop-2} $\mathbf{+0.007}$, bad \textbf{0/16} & 0.033 & Tail failures $3 \to 0$ \\
Qwen2.5-3B LB ($n{=}32$, coupled $m_r$ vs.\ $m_r^{\mathrm{ref}}$) & $-0.024$\,nats shift ($11.99$\,ms slot~16) & $\Delta\text{lp}\!=\!\mathbf{-0.020}$, 24/32 (\textbf{11.87\,ms}) & 0.044 & Zero latency cost \\
\bottomrule
\end{tabular}}
\end{table}

\begin{table}[ht]
\caption{\textbf{Strict token-causal quantization consistency audit (\cref{sec:conditioning})} on 16 held-out LongBench prompts (\texttt{hotpotqa} + \texttt{narrativeqa}) under \oursfwht{} ($\Delta{=}1$). Static offline $(\mu_k^{\mathrm{off}}, g_Q^{\mathrm{off}}, g_K^{\mathrm{off}}, g_V^{\mathrm{off}})$ calibrated on 4 disjoint prompts (\texttt{gov\_report}, \texttt{qasper}; $5.2$--$6.2$\,s, $2\times$ headroom) removes all future-token coupling from $Q, K$ quantization with $0.000\%$ E4M3 microscale clipping and matching attention relative error.}
\label{tab:app_causal_audit}
\centering
\small
\setlength{\tabcolsep}{4.5pt}
\begin{tabular}{@{}llrrrrrr@{}}
\toprule
Model & Quantization parameter regime & mean $\Delta\text{lp}$ & med $\Delta\text{lp}$ & \texttt{tf-ok} & \texttt{bad} & rel.\ err & E4M3 clip (Q/K/V) \\
\midrule
\multirow{3}{*}{Llama-3.1-8B}
 & Prompt default (prefix-2048 $\mu_k$, prompt $g_X$) & $+0.0156$ & $+0.0221$ & 4/16 & 0 & 0.1066 & 0.000/0.000/0.000\% \\
 & Offline $\mu_k^{\mathrm{off}}$ + prompt $g_X$ & $+0.0674$ & $+0.0545$ & 4/16 & 1 & 0.1046 & 0.000/0.000/0.000\% \\
 & \textbf{Strict token-causal static} ($\mu_k^{\mathrm{off}}, g_X^{\mathrm{off}}$) & $-0.0273$ & $-0.0093$ & 4/16 & 1 & \textbf{0.1042} & \textbf{0.000/0.000/0.000\%} \\
\midrule
\multirow{3}{*}{Qwen2.5-7B}
 & Prompt default (prefix-2048 $\mu_k$, prompt $g_X$) & $-0.0507$ & $+0.0014$ & 2/16 & 3 & 0.0873 & 0.000/0.000/0.001\% \\
 & Offline $\mu_k^{\mathrm{off}}$ + prompt $g_X$ & $+0.0040$ & $-0.0085$ & 2/16 & 3 & 0.0854 & 0.000/0.000/0.001\% \\
 & \textbf{Strict token-causal static} ($\mu_k^{\mathrm{off}}, g_X^{\mathrm{off}}$) & $+0.3407$ & $+0.0026$ & 2/16 & \textbf{2} & \textbf{0.0861} & \textbf{0.000/0.000/0.000\%} \\
\midrule
\begin{tabular}{@{}l@{}}DeepSeek-R1-\\Llama-8B (CoT)\end{tabular}
 & \textbf{Zero-shot static} (4 \texttt{gov\_report}, $2.8$\,s, $2\times$)$^\dagger$ & --- & --- & --- & --- & 0.0154$^\ddagger$ & \textbf{---/0.000/0.000\%} \\
\bottomrule
\end{tabular}
\\[1pt]{\footnotesize $^\dagger$Audited across all 32 layers of 4 generated AIME Long-CoT trajectories ($2{,}933$--$4{,}096$ generated tokens): $0 / 32{,}198{,}656$ $K$ blocks and $1 / 32{,}112{,}640$ $V$ blocks clipped. $^\ddagger$Bit-exact incremental 64-token chunk pack (\texttt{pack\_kv\_chunk}) vs.\ full-sequence static pack; left-padded decode rel.\ err $0.0154$ ($\cos{=}0.99988$), $0.011$\,ms/layer at $B{=}16$ ($0.0055$\,ms/step amortized over 64 steps).}
\end{table}

\begin{table}[ht]
\caption{\textbf{Full 13-task RULER evaluation on Llama-3.1-8B-Instruct} at 16K ($n{=}20$/task = 260 prompts) and 32K ($n{=}15$/task = 195 prompts), adding \texttt{niah\_s3/mk1/mk3}, diffuse aggregation (\texttt{cwe}, \texttt{fwe}), and multi-doc QA (\texttt{qa\_1} SQuAD, \texttt{qa\_2} HotpotQA). \oursfwht{} ($\Delta{=}1$) matches BF16 within paired bootstrap $95\%$ CIs at both lengths while running $1.16$--$1.34\times$ faster E2E than BF16 (faster than \sagetwo{}); diffuse and QA tasks exhibit \emph{lower} prefill warp/CTA flag rates than peaked \texttt{niah\_mk2}.}
\label{tab:app_ruler13}
\centering
\small
\setlength{\tabcolsep}{2.2pt}
\resizebox{\textwidth}{!}{%
\begin{tabular}{@{}clrrrrrrrrrrrrr|rrr@{}}
\toprule
$T$ & Method & \texttt{s1} & \texttt{s2} & \texttt{s3} & \texttt{mk1} & \texttt{mk2} & \texttt{mk3} & \texttt{mv} & \texttt{mq} & \texttt{vt} & \texttt{cwe} & \texttt{fwe} & \texttt{qa1} & \texttt{qa2} & \textbf{AVG13} & Diff vs.\ BF16 ($p$) & Time \\
\midrule
\multirow{5}{*}{16K}
 & BF16 SDPA & 100 & 100 & 100 & 100 & 95.0 & 100 & 100 & 87.5 & 71.0 & 3.5 & 80.0 & 90.0 & 55.0 & 83.23\% & $+0.00$ (1.000) & 504\,s \\
 & \sagetwo{} (GQA) & 100 & 100 & 100 & 100 & 100 & 100 & 100 & 87.5 & 56.0 & 3.5 & 80.0 & 90.0 & 55.0 & 82.46\% & $-0.77$ (0.308) & 447\,s \\
 & 1-limb + FWHT & 100 & 95.0 & 100 & 100 & 100 & 90.0 & 100 & 80.0 & 89.0 & 13.0 & 78.3 & 90.0 & 55.0 & 83.87\% & $+0.64$ (0.613) & 425\,s \\
 & \cellcolor{tierfour!10}\textbf{\oursfwht{} ($\Delta{=}1$)} & \cellcolor{tierfour!10}95.0 & \cellcolor{tierfour!10}95.0 & \cellcolor{tierfour!10}100 & \cellcolor{tierfour!10}100 & \cellcolor{tierfour!10}100 & \cellcolor{tierfour!10}100 & \cellcolor{tierfour!10}100 & \cellcolor{tierfour!10}88.8 & \cellcolor{tierfour!10}88.0 & \cellcolor{tierfour!10}0.5 & \cellcolor{tierfour!10}80.0 & \cellcolor{tierfour!10}90.0 & \cellcolor{tierfour!10}50.0 & \cellcolor{tierfour!10}\textbf{83.63\%} & \cellcolor{tierfour!10}$+0.40$ (0.677) & \cellcolor{tierfour!10}\textbf{433\,s} \\
 & {\footnotesize \emph{ZM warp/CTA flag (\%)}} & {\footnotesize 10.1/16} & {\footnotesize 10.2/16} & {\footnotesize 10.1/16} & {\footnotesize 10.0/16} & {\footnotesize 11.6/21} & {\footnotesize 8.2/16} & {\footnotesize 10.0/16} & {\footnotesize 9.9/16} & {\footnotesize 10.1/16} & {\footnotesize \textbf{7.1/14}} & {\footnotesize \textbf{8.9/16}} & {\footnotesize \textbf{6.5/14}} & {\footnotesize \textbf{5.3/12}} & --- & --- & --- \\
\midrule
\multirow{5}{*}{32K}
 & BF16 SDPA & 100 & 100 & 100 & 100 & 100 & 100 & 100 & 83.3 & 100 & 1.3 & 84.4 & 93.3 & 66.7 & 86.85\% & $+0.00$ (1.000) & 850\,s \\
 & \sagetwo{} (GQA) & 100 & 100 & 100 & 100 & 100 & 100 & 100 & 83.3 & 100 & 0.0 & 86.7 & 93.3 & 66.7 & 86.92\% & $+0.07$ (0.949) & 670\,s \\
 & 1-limb + FWHT & 100 & 100 & 100 & 100 & 100 & 93.3 & 100 & 88.3 & 100 & 6.0 & 84.4 & 100 & 60.0 & 87.09\% & $+0.23$ (0.805) & 629\,s \\
 & \cellcolor{tierfour!10}\textbf{\oursfwht{} ($\Delta{=}1$)} & \cellcolor{tierfour!10}100 & \cellcolor{tierfour!10}100 & \cellcolor{tierfour!10}100 & \cellcolor{tierfour!10}100 & \cellcolor{tierfour!10}100 & \cellcolor{tierfour!10}100 & \cellcolor{tierfour!10}100 & \cellcolor{tierfour!10}83.3 & \cellcolor{tierfour!10}94.7 & \cellcolor{tierfour!10}2.0 & \cellcolor{tierfour!10}84.4 & \cellcolor{tierfour!10}93.3 & \cellcolor{tierfour!10}66.7 & \cellcolor{tierfour!10}\textbf{86.50\%} & \cellcolor{tierfour!10}$-0.36$ (0.514) & \cellcolor{tierfour!10}\textbf{634\,s} \\
 & {\footnotesize \emph{ZM warp/CTA flag (\%)}} & {\footnotesize 5.4/9.0} & {\footnotesize 5.5/9.1} & {\footnotesize 5.5/9.1} & {\footnotesize 5.5/9.1} & {\footnotesize 6.4/13.4} & {\footnotesize 4.2/9.2} & {\footnotesize 5.4/9.0} & {\footnotesize 5.4/9.0} & {\footnotesize 5.5/9.1} & {\footnotesize \textbf{3.7/8.8}} & {\footnotesize \textbf{4.8/9.9}} & {\footnotesize \textbf{3.6/8.9}} & {\footnotesize \textbf{2.7/6.5}} & --- & --- & --- \\
\bottomrule
\end{tabular}}
\end{table}

\begin{table}[ht]
\caption{\textbf{Paired bootstrap $95\%$ confidence intervals vs.\ BF16} ($10{,}000$ resamples, per-prompt paired differences) on Llama-3.1-8B high-power RULER 48K ($n{=}30$/task = 180 prompts) and official 10-task LongBench ($n{=}300$ prompts). Unrefined 1-limb+FWHT ($p{=}0.0002$) and raw \thrift{}~5\% ($p{=}0.0342$) significantly degrade on RULER 48K, whereas \oursfwht{} ($\Delta{=}1$) and unified \ourscache{} are statistically indistinguishable from BF16.}
\label{tab:app_bootstrap}
\centering
\small
\setlength{\tabcolsep}{3.6pt}
\resizebox{\textwidth}{!}{%
\begin{tabular}{@{}lrrrr@{\hspace{10pt}}rrr@{}}
\toprule
 & \multicolumn{4}{c}{Llama-3.1-8B RULER 48K ($n{=}180$ prompts)} & \multicolumn{3}{c}{Llama-3.1-8B Official LB-10 ($n{=}300$ prompts)} \\
\cmidrule(lr){2-5}\cmidrule(l){6-8}
Method & Mean [95\% CI] & Diff vs.\ BF16 [95\% CI] & $p$-val & Time & Mean [95\% CI] & Diff vs.\ BF16 [95\% CI] & $p$-val \\
\midrule
BF16 SDPA & 99.86 [99.58, 100.00] & $+0.00$ [0.00, 0.00] & 1.0000 & 1290\,s & 49.14 [44.41, 53.79] & $+0.00$ [0.00, 0.00] & 1.0000 \\
\sagetwo{} (GQA) & 100.00 [100.00, 100.00] & $+0.14$ [+0.00, +0.42] & 0.7260 & 950\,s & 49.22 [44.52, 53.76] & $+0.08$ [$-1.33$, $+1.62$] & 0.9192 \\
\thrift{} 5\% (raw) & 98.36 [96.53, 99.72] & \textcolor{tierwarn}{$-1.50$ [$-3.33$, $-0.14$]} & \textcolor{tierwarn}{0.0342} & 976\,s & 49.50 [44.83, 54.20] & $+0.36$ [$-1.99$, $+2.84$] & 0.7602 \\
\thrift{} 5\% (+ctr+FWHT) & 99.44 [98.89, 99.86] & $-0.42$ [$-1.11$, $+0.14$] & 0.2464 & 1161\,s & --- & --- & --- \\
\ours{} (1-limb+FWHT) & 95.50 [92.56, 98.03] & \textcolor{tierwarn}{$-4.36$ [$-7.33$, $-1.83$]} & \textcolor{tierwarn}{0.0002} & 860\,s & 48.10 [43.39, 52.75] & $-1.04$ [$-3.58$, $+1.43$] & 0.4268 \\
\rowcolor{tierfour!10}
\textbf{\oursfwht{}} ($\Delta{=}1$) & \textbf{100.00 [100.00, 100.00]} & $+0.14$ [+0.00, +0.42] & 0.7260 & \textbf{875\,s} & \textbf{49.80 [45.07, 54.44]} & $+0.66$ [$-1.40$, $+2.74$] & 0.5364 \\
\rowcolor{tierfour!10}
\textbf{\ourscache{}} (unified) & --- & --- & --- & --- & \textbf{49.74 [45.04, 54.42]} & $+0.60$ [$-1.18$, $+2.45$] & 0.4972 \\
\bottomrule
\end{tabular}}
\end{table}

\paragraph{Extended LongBench tasks and BF16 noise floor.}\label{app:extended}
On three additional LongBench tasks (\texttt{triviaqa}, \texttt{qmsum}, \texttt{passage\_count}; Qwen2.5-3B, $n{=}30$, $12\text{K}$--$30\text{K}$ tokens), centered plain FP4 loses $-1.035$\,nats ($17/30$ bad), centered \thrift{} (5\%) loses $-0.265$\,nats ($8/30$ bad), \deco{} loses $-0.092$\,nats ($3/30$ bad), and emulated \oursband{} ($\Delta{=}2$, $7.2\%$ refined) achieves median $0.000$\,nats with $22/30$ \texttt{tf-ok} (vs.\ BF16 $24/30$) and $1/30$ bad example.
Diagnostic inspection of that single \texttt{passage\_count} outlier showed that Qwen2.5-3B itself is sensitive to BF16 summation order on that prompt: switching BF16 FlashAttention-2 to exact chunked BF16 SDPA shifts mean logp by $-0.037$\,nats, and switching to exact eager FP32-softmax shifts mean logp by $-0.169$\,nats ($1/10$ bad), while refining $100\%$ of stages reproduces the same outlier ($-0.428$\,nats).
By contrast, on 32K 2-hop chain lookup ($n{=}32$), the BF16 arithmetic noise floor between FlashAttention-2, chunked SDPA, and eager FP32-softmax is within $\pm 0.007$\,nats ($0/32$ bad).

\end{document}